\documentclass{article}
\usepackage{microtype}
\usepackage{graphicx}
\usepackage{booktabs} % for professional tables
\usepackage{amsmath}
\usepackage{microtype}
\usepackage{enumitem}
\usepackage{colortbl}
\usepackage[disable]{todonotes}
\usepackage{rotating}
\usepackage{tabulary}
\usepackage{caption}
\usepackage{subcaption}
\usepackage{mathtools}
\usepackage{hyperref}
\usepackage[thmmarks, thref, amsmath]{ntheorem}
\newtheorem*{lemma-nono}{Lemma}
\newtheorem*{theorem-nono}{Theorem}
\theoremstyle{plain}
\newtheorem{theorem}{Theorem}
\newtheorem{corollary}{Corollary}
\newtheorem{lemma}{Lemma}
\theoremheaderfont{\itshape}
\theorembodyfont{\upshape}
\theoremstyle{plain}
\newtheorem{case}{Case}
\theoremstyle{nonumberplain}
\theoremheaderfont{\scshape}
\theorembodyfont{\upshape}
\theorempostwork{\setcounter{case}{0}}
\newtheorem{proof}{Proof}

\usepackage[accepted]{icml2021}

\icmltitlerunning{\model: Gradient Matching based Data Subset Selection for Efficient Training}
\usepackage{bm}
\usepackage{appendix}
\usepackage{graphicx}
\usepackage{xcolor}
\usepackage{amssymb}
\usepackage{todonotes}
\newcommand{\wb}{\mathbf{w}}
\usepackage{multirow}
\usepackage{adjustbox}
\usepackage{microtype}
\usepackage{color}

\usepackage{caption}

\usepackage{comment}
\usepackage{defs}

\begin{document}
\twocolumn[
\icmltitle{\textsc{Grad-Match}: Gradient Matching based Data Subset Selection for Efficient Deep Model Training}



\icmlsetsymbol{equal}{*}

\begin{icmlauthorlist}
\icmlauthor{Krishnateja Killamsetty}{to}
\icmlauthor{Durga Sivasubramanian}{ed}
%\icmlauthor{Baharan Mirzasoleiman}{goo}
\icmlauthor{Ganesh Ramakrishnan}{ed}
\icmlauthor{Abir De}{ed}
\icmlauthor{Rishabh Iyer}{to}
% \icmlauthor{Aaoeu Iasoh}{goo}
% \icmlauthor{Buiui Eueu}{ed}
% \icmlauthor{Aeuia Zzzz}{ed}
% \icmlauthor{Bieea C.~Yyyy}{to,goo}
% \icmlauthor{Teoau Xxxx}{ed}
% \icmlauthor{Eee Pppp}{ed}
\end{icmlauthorlist}

\icmlaffiliation{to}{Department of Computer Science, University of Texas at Dallas, Dallas, USA}
%\icmlaffiliation{goo}{Department of Computer Science, University of California, Los Angeles, USA}
\icmlaffiliation{ed}{Department of Computer Science and Engineering, Indian Institute of Technology, Bombay, India}

\icmlcorrespondingauthor{Krishnateja Killamsetty}{krishnateja.killamsetty@utdallas.edu}
% \icmlcorrespondingauthor{Eee Pppp}{ep@eden.co.uk}

\icmlkeywords{Machine Learning, Efficient Machine Learning, Data Selection}
\vskip 0.3in]



\printAffiliationsAndNotice{}  % leave blank if no need to mention equal contribution
\begin{abstract}
The great success of modern machine learning models on large datasets is contingent on extensive computational resources with high financial and environmental costs. One way to address this is by extracting subsets that generalize on par with the full data. In this work, we propose a general framework, \model{}, which finds subsets that closely match the gradient of the \emph{training or validation} set. We find such subsets effectively using an orthogonal matching pursuit algorithm. We show rigorous theoretical and convergence guarantees of the proposed algorithm and, through our extensive experiments on real-world datasets, show the effectiveness of our proposed framework. We show that \textsc{Grad-Match} significantly and consistently outperforms several recent data-selection algorithms and achieves the best accuracy-efficiency trade-off. \textsc{Grad-Match} is available as a part of the CORDS toolkit: \url{https://github.com/decile-team/cords}.
%\footnote{\scriptsize{The code of \model{} is available at \url{https://anonymous.4open.science/r/2eada7f6-5514-4409-8d90-ff4f8d7d5c03/}}}.
%For example, when training with ResNet-18 on CIFAR-10, we observe a 7x and 3x speedup (and similar energy saving) with an accuracy drop of 2.5\% and 0.8\% respectively, while for CIFAR-100, we achieve a 4.8x and 3x speedup and energy saving, while loosing around 2.1\% and 0.7\% in accuracy.\looseness-1
\end{abstract}

\section{Introduction}
Modern machine learning systems, especially deep learning frameworks, have become very computationally expensive and data-hungry. Massive training datasets have significantly increased end-to-end training times, computational and resource costs~\cite{sharir2020cost}, energy requirements~\cite{strubell2019energy}, and carbon footprint~\cite{schwartz2019green}. Moreover, most machine learning models require extensive hyper-parameter tuning, further increasing the cost and time, especially on massive datasets.
In this paper, we study efficient machine learning through the paradigm of subset selection, which seeks to answer the following question: \emph{Can we train a machine learning model on much smaller subsets of a large dataset, with negligible loss in test accuracy?} 

Data subset selection enables efficient learning at multiple levels. First, by using a subset of a large dataset, we can enable learning on relatively low resource computational environments without requiring a large number of GPU and CPU servers. Second, since we are training on a subset of the training dataset, we can significantly improve the end-to-end turnaround time, which often requires multiple training runs for hyper-parameter tuning.
Finally, this also enables significant reduction in the energy consumption and CO2 emissions of deep learning~\cite{strubell2019energy}, particularly since a large number of deep learning experiments need to be run in practice. 
%The critical requirements of the data selection techniques are speed, scalability, and ability to handle massive datasets. 
Recently, there have been several efforts to make
machine learning models more efficient via data subset selection~\cite{wei2014fast, kaushal2019learning,coleman2020selection,har2004coresets,clarkson2010coresets,mirzasoleiman2020coresets,killamsetty2021glister}. Existing approaches either use proxy functions to select data points, or are specific to particular machine learning models, or use approximations of quantities such as gradient error or generalization errors. In this work, we propose a data selection framework called \textsc{Grad-Match}, which exactly minimizes a residual error term obtained by analyzing adaptive data subset selection algorithms, therefore admitting theoretical guarantees on convergence.

\subsection{Contributions of this work}
\noindent \textbf{Analyses of convergence bounds of adaptive data subset selection algorithms. } A growing number of recent approaches~\cite{mirzasoleiman2020coresets,killamsetty2021glister} can be cast within the framework of \emph{adaptive data subset selection}, where the data subset selection algorithm (which selects the data subset depending on specific criteria) is applied in conjunction with the model training. As the model training proceeds, the subset on which the model is being trained is improved via the current model's snapshots. We analyze the convergence of this general framework and show that the convergence bound \emph{critically} depends on an additive error term depending on how well the subset's weighted gradients match either the full training gradients or the full validation gradients ({\em c.f.},  Section~\ref{adap-dss-framework}).
\\\\
\noindent \textbf{Data selection framework with convergence guarantees. } Inspired by the result above, we present \model, a gradient-matching algorithm ({\em c.f.}, Section~\ref{grad-match-algo}), which directly tries to minimize the \emph{gradient matching error}. As a result, we are able to show convergence bounds for a large class of convex loss functions. We also argue that the resulting bounds we obtain are tighter than those of recent works~\cite{mirzasoleiman2020coresets, killamsetty2021glister}, which either use upper bounds or their approximations. We then show that minimizing the gradient error can be cast as a weakly submodular maximization problem and propose an orthogonal matching pursuit based greedy algorithm. We then propose several implementation tricks ({\em c.f.}, Section~\ref{speedups-further}), which provide significant speedups for the data selection step. 

\begin{figure}[h]
\begin{center}
\includegraphics[width = 6cm]{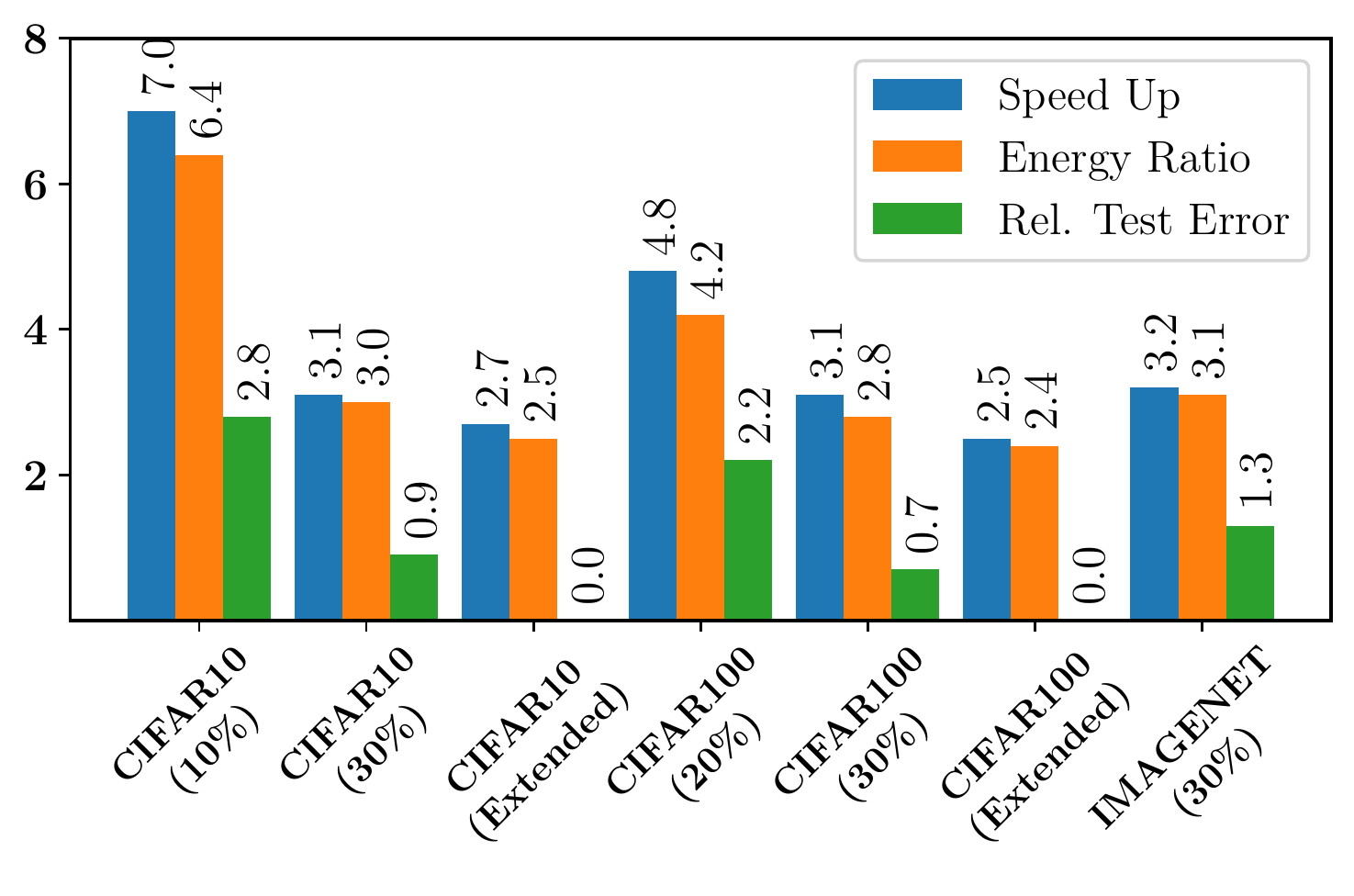}
\caption{Efficiency of \model\ relative to full training on the CIFAR-10, CIFAR-100, and Imagenet datasets.}
\label{fig:mainresults}
\end{center}
\end{figure}
% \vspace{-2ex}

\noindent \textbf{Best tradeoff between efficiency and accuracy.} In our empirical results ({\em c.f.}, Section~\ref{exp-results}), \emph{we show that \model\ achieves much a better accuracy and training-time trade-off than several state-of-the-art approaches such as  \textsc{Craig}~\cite{mirzasoleiman2020coresets}, \textsc{Glister}~\cite{killamsetty2021glister}, random subsets, and even full training with early stopping.} % (with comparable time/energy consumption)
%We see more significant improvements on very small subset sizes (between 5 - 10\%), resulting in larger speedups on a single GPU and energy savings. 
When training with ResNet-18 on Imagenet, CIFAR-10, and CIFAR-100, we observe around 3$\times$ efficiency improvement with an accuracy drop of close to 1\% for 30\% subset. With smaller subsets ({\em e.g.}, 20\% and 10\%), we sacrifice a little more in terms of accuracy ({\em e.g.}, 2\% and 2.8\%) for a larger speedup in training (4$\times$ and 7$\times$). Furthermore, we see that by extending the training beyond the specified number of epochs (300 in this case), we can match the accuracy on the full dataset using just 30\% of the data while being overall 2.5$\times$ faster. In the case of MNIST, the speedup improvements are even more drastic, ranging from 27$\times$ to 12$\times$ with 0.35\% to 0.05\% accuracy degradation.
%Similarly, on CIFAR-10, we observe 7x and 3x   efficiency improvements with accuracy drops of 2.8\% and 0.9\% for 10\% and 30\% subsets respectively. For CIFAR-100, we achieve 4.8x and 3x efficiency improvements, while losing around 2.1\% and 0.7\% in accuracy for 20\% and 30\% subsets respectively. In the case of MNIST, we achieve 27x, 16x and 12x efficiency improvements while losing 0.35\%, 0.12\% and 0.05\% in accuracy for 1\%, 3\% and 5\% subsets respectively. We observe similar results for imbalanced datasets, thus demonstrating the robustness of  \model.\looseness-1

\begin{figure*}
\begin{center}
\includegraphics[width = 0.8 \textwidth]{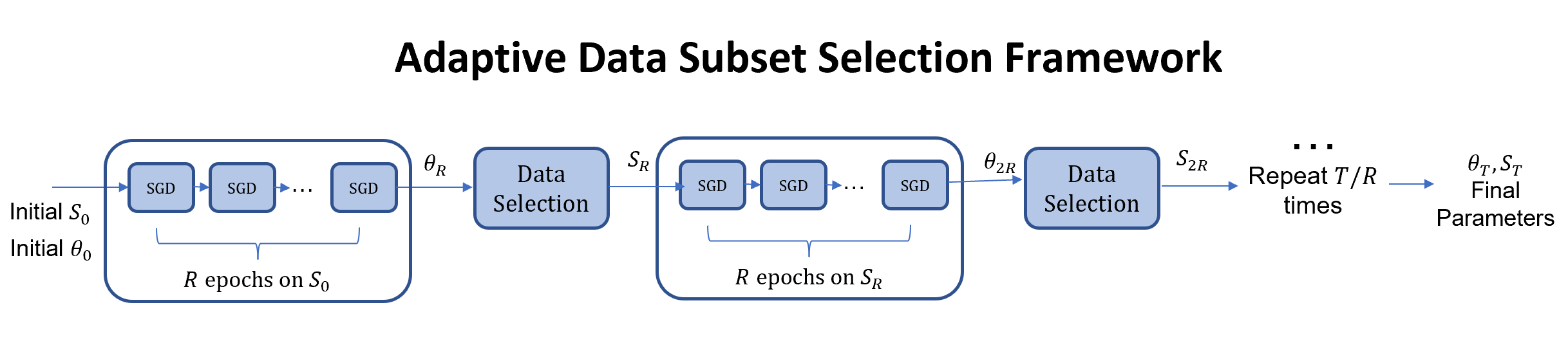}
\caption{Block Diagram of any adaptive data selection algorithm, where data selection is performed every $R$ epochs of (stochastic) gradient descent, and the gradient descent updates are performed on the subsets obtained by the data selection.}
\label{adaptivedss}
\end{center}
\end{figure*}
%\vspace{-4ex}

\subsection{Related work}
A number of recent papers have used submodular functions as \emph{proxy} functions~\cite{wei2014fast,wei2014unsupervised,kirchhoff2014submodularity,kaushal2019learning} (to actual loss). Let $n$ be the number of data points in the ground set. Then a set function $f: 2^{[n]} \rightarrow \mathbf{R}$ is submodular ~\cite{fujishige2005submodular} if it satisfies the diminishing returns property: for subsets $S, T \subseteq [n], f(j | S) \triangleq  f(S \cup j) - f(S) \geq f(j | T)$.
%These approaches were used in several domains, including speech recognition~\cite{wei2014unsupervised,wei2014submodular}, machine translation~\cite{kirchhoff2014submodularity} and computer vision~\cite{kaushal2019learning}. 
Another commonly used approach is that of coresets. Coresets are weighted subsets of the data, which approximate certain desirable characteristics of the full data ({\em, e.g.}, the loss function)~\cite{feldman2020core}. Coreset algorithms have been used for several problems including $k$-means and $k$-median clustering~\cite{har2004coresets}, SVMs~\cite{clarkson2010coresets} and Bayesian inference~\cite{campbell2018bayesian}. Coreset algorithms require algorithms that are often specialized and very specific to the model and problem at hand and have had limited success in deep learning.

A very recent coreset algorithm called \textsc{Craig}~\cite{mirzasoleiman2020coresets} has shown promise for both deep learning and classical machine learning models such as logistic regression. Unlike other coreset techniques that largely focus on approximating loss functions, \textsc{Craig} selects representative subsets of the training data that closely approximate the full gradient. Another approach poses the data selection as a discrete bi-level optimization problem and shows that, for several choices of loss functions and models~\cite{killamsetty2021glister, wei2015submodularity}, the resulting optimization problem is submodular. A few recent papers have studied data selection approaches for robust learning. \citet{mirzasoleiman2020coresetsnoise} extend \textsc{Craig} to handle noise in the data, whereas \citet{killamsetty2021glister} study class imbalance and noisy data settings by assuming access to a clean validation set. In this paper, we study data selection under class imbalance in a setting similar to~\cite{killamsetty2021glister}, where we assume access to a clean validation set. Our work is also related to highly distributed deep learning systems~\cite{jia2018highly} which make deep learning significantly faster using a cluster of hundreds of GPUs. In this work, we instead focus on \emph{single GPU} training runs, which are more practical for smaller companies and academic labs and potentially complementary to~\cite{jia2018highly}. Finally, our work is also complementary to that of~\cite{wang2019e2}, where the authors employ tricks such as selective layer updates, low-precision backpropagation, and random subsampling to achieve significant energy reductions. In this work, we demonstrate both energy and time savings \emph{solely} based on a more principled subset selection approach.

\section{\textsc{Grad-Match} through the lens of adaptive data subset selection}
\label{adap-dss-framework}
In this section, we will study the convergence of general adaptive subset selection algorithms and use the result (Theorem~\ref{thm:convergence-result}) to motivate \model.

\textbf{Notation. } Denote $\Ucal = \{(x_i, y_i)\}_{i=1}^{N}$, as the set of training examples, and let $\Vcal = \{(x_j, y_j)\}_{j=1}^{M}$ denote the validation set. Let $\theta$ be the classifier model parameters. Next, denote by $L_T^i(\theta) = L_T(x_i, y_i, \theta)$,  the training loss at the $i^{th}$ epoch of training, and let $L_T(\Xcal, \theta) = \sum_{i \in \Xcal} L_T(x_i, y_i, \theta)$ be the loss on a subset $\Xcal \subseteq \Ucal$ of the training examples. Let the validation loss be denoted by $L_V$. 
%Finally, we let $R$  denote the frequency at which data selection is performed. 
In Table 1
%~\ref{tab:main-notations} 
in Appendix~\ref{app-notation-summary}, we organize and tabulate the notations used throughout this paper. 

\noindent \textbf{Adaptive data subset selection: } (Stochastic) gradient descent algorithms proceed by computing the gradients of \emph{all} the training data points for $T$ epochs. We study the alternative approach of  adaptive data selection, in which training is performed using a weighted sum of gradients of the training data subset instead of the full training data.  In the adaptive data selection approach, the data selection is performed \emph{in conjunction}  with training such that subsets get incrementally refined as the learning algorithm proceeds. This incremental subset refinement allows the data selection to \emph{adapt} to the learning and produces increasingly effective subsets with progress of the learning algorithm. Assume that an adaptive data selection algorithm produces weights $\wb^t$ and subsets $\Xcal^t$ for $t = 1, 2, \cdots, T$ through the course of the algorithm. In other words, at iteration $t$, the parameters $\theta_t$ are updated using the weighted loss of the model on subset $\Xcal^t$ by weighing each example $i \in \Xcal^t$ with its corresponding weight $w^t_i$. For example, if we use gradient descent, we can write the update step as: $\theta_{t+1} = \theta_t - \alpha \sum_{i \in \Xcal^t} w^t_i \nabla_{\theta} L_T(x_i, y_i, \theta_t)$. 
%A very similar approach can be taken for stochastic gradient descent.\todo{Is there a need for mentioning anything about the stochastic version here?} 
Note that though this framework uses a potentially different set $\Xcal^t$ in each iteration $t$, we need not perform data selection every epoch. In fact, in practice, we run data selection only every $R$ epochs, in which case, the same subsets and weights will be used between epochs $t = R\tau$ and $t = R(\tau + 1)$. In contrast, the non-adaptive data selection settings~\cite{wei2014submodular,wei2014unsupervised,kaushal2019learning} employ the same $\Xcal^t = \Xcal$ for gradient descent in every iteration. In Figure~\ref{adaptivedss}, we present the broad adaptive data selection scheme. In this work, we focus on a setting in which the subset size is fixed, {\em i.e.}, $|\Xcal^t| = k$ (typically a small fraction of the training dataset). 

\noindent \textbf{Convergence analysis: } We now study the conditions for the convergence of either the full training loss or the validation loss achieved by any adaptive data selection strategy. Recall that we can characterize any data selection algorithm by a set of weights $\wb^t$ and subsets $\Xcal^t$ for $t = 1, \cdots, T$. We provide a convergence result which holds for any adaptive data selection algorithm, and applies to Lipschitz continuous, Lipschitz smooth, and strongly convex loss functions. Before presenting the result, we define the term:
{
$$\small \mbox{Err}(\wb^t, \Xcal^t, L, L_T, \theta_t) = {\left\Vert \sum_{i \in \Xcal^t} w^t_i \nabla_{\theta}L_T^i(\theta_t) -  \nabla_{\theta}L(\theta_t)\right\Vert}$$}

The norm considered in the above equation is the l2 norm. Next, assume that the parameters satisfy $||\theta||^2 \leq D^2$, and let $L$ denote either the training or validation loss. We next state the convergence result:
\begin{theorem}
\small
\label{thm:convergence-result}
Any adaptive data selection algorithm. run with full gradient descent (GD), defined via weights $\wb^t$ and subsets $\Xcal^t$ for $t = 1, \cdots, T$, enjoys the following guarantees:\vspace{2mm}\\
\noindent (1). If $L_T$ is Lipschitz continuous with parameter $\sigma_T$, optimal model parameters are $\theta^*$, and $\alpha = \frac{D}{\sigma_T \sqrt{T}}$, then $\min_{t = 1:T} L(\theta_t) - L(\theta^*) \leq \frac{D\sigma_T}{\sqrt{T}} + \frac{D}{T}\sum_{t=1}^{T-1} \mbox{Err}(\wb^t, \Xcal^t, L, L_T, \theta_t)$. \vspace{2mm}\\
\noindent (2) If $L_T$ is Lipschitz smooth with parameter $\mathcal{L}_T$, optimal model parameters are $\theta^*$, and $L_T^i$ satisfies $0 \leq L_T^i(\theta) \leq \beta_T$, $\forall i$, then setting $\alpha = 1/{\mathcal{L}_T}$, we have $\min_{t = 1:T} L(\theta_t) - L(\theta^*) \leq \frac{D^2 \mathcal{L}_T + 2\beta_T}{2T}  +\frac{D}{T}\sum_{t=1}^{T-1} \mbox{Err}(\wb^t, \Xcal^t, L, L_T, \theta_t)$. \vspace{2mm}\\
 \noindent (3) If $L_T$ is Lipschitz continuous with parameter $\sigma_T$, optimal model parameters are $\theta^*$, and $L$ is strongly convex with parameter $\mu$, then setting a learning rate $\alpha_t = \frac{2}{\mu(1+t)}$, we have $\min_{t = 1:T} L(\theta_t) - L(\theta^*) \leq \frac{2{\sigma_T}^2}{\mu (T+1)} + \sum_{t=1}^{t=T} \frac{2Dt}{T(T+1)} \mbox{Err}(\wb^t, \Xcal^t, L, L_T, \theta_t)$.
\end{theorem}
% \vspace{-1ex}
We present the proof in Appendix~\ref{app-conv-res-proof}. We can also extend this result to the case in which SGD is used instead of full GD to update the model. The main difference in the result is that the inequality holds with expectations over both sides. We defer the convergence result and proof to Appendix~\ref{app-conv-res-sgd}. 
%The convergence bounds for adaptive subset selection with full or stochastic gradient descent indicates that we can get better convergence by controlling the error term $\mbox{Err}(\wb^t, X_t, L, L_T, \theta_t)$. 
Theorem~\ref{thm:convergence-result} suggests that an effective data selection algorithm should try to obtain subsets which have very small error $\mbox{Err}(\wb^t, \Xcal^t, L, L_T, \theta_t)$ for $t = 1, \cdots, T$. If the goal of data selection is to select a subset $\Xcal^t$ at every epoch which \emph{approximates} the full training set, the data selection procedure should try to minimize $\mbox{Err}(\wb^t, \Xcal^t, L_T, L_T, \theta_t)$. On the other hand, to select subset $\Xcal^t$ at every epoch which \emph{approximates} the validation set, the data selection procedure should try to minimize $\mbox{Err}(\wb^t, \Xcal^t, L_V, L_T, \theta_t)$. In Appendix~\ref{app-reduce-loss-dss}, we provide conditions under which an adaptive data selection approach reduces the loss function $L$ (which can either be the training loss $L_T$ or the validation loss $L_V$). In particular, we show that any data selection approach which attempts to minimize the error term $\mbox{Err}(\wb^t, \Xcal^t, L, L_T, \theta_t)$ is also likely to reduce the loss value at every iteration. In the next section, we present \model{}, that directly minimizes this error function $\mbox{Err}(\wb^t, \Xcal^t, L, L_T, \theta_t)$.
\section{The \model\ Algorithm}
\label{grad-match-algo}
Following the result of Theorem~\ref{thm:convergence-result}, we now design an adaptive data selection algorithm that minimizes the gradient error term: $\mbox{Err} (\wb^t, \Xcal^t, L, L_T, \theta_t)$ (where $L$ is either the training loss or the validation loss) as the training proceeds. The complete algorithm is shown in Algorithm~\ref{alg:grad-match}. In Algorithm~\ref{alg:grad-match}, isValid is a boolean Validation flag indicator that indicates whether to match the subset loss gradient with validation set loss gradient like in the case of class imbalance (isValid=True) or training set loss gradient (isValid=False). As discussed in Section~\ref{adap-dss-framework}, we do the subset selection only every $R$ epochs, and during the rest of the epochs, we update the model parameters (using Batch SGD) on the previously chosen set $\Xcal^t$ with associated weights $\wb^t$. \emph{This ensures that the subset selection time in itself is negligible compared to the training time, thereby ensuring that the adaptive selection runs as fast as simple random selection.} The full algorithm is shown in Algorithm~\ref{alg:grad-match}. Line 9 of Algorithm~\ref{alg:grad-match} is the mini-batch SGD and takes as inputs the weights, subset of instances, learning rate, training loss, batch size, and the number of epochs. We randomly shuffle elements in the subset $\Xcal^t$, divide them up into mini-batches of size $B$, and run mini-batch SGD with instance weights. 

Lines 3 and 5 in Algorithm~\ref{alg:grad-match} are the data subset selection steps, run either with the full training gradients or validation gradients. We basically minimize the error term $\mbox{Err}(\wb^t, \Xcal^t, L, L_T, \theta_t)$ with minor difference. Define the regularized version of $\mbox{Err}(\wb^t, \Xcal^t, L, L_T, \theta_t)$ as:
\begin{align}
    \mbox{Err}_{\lambda}(\wb, \Xcal, L, L_T, \theta_t) = &\mbox{Err}(\wb, \Xcal, L, L_T, \theta_t) +  \lambda \Vert\wb\Vert^2
\end{align}
The data selection optimization problem then is:
\begin{equation}
\begin{aligned}
  \hspace{-2mm}\wb^t, \Xcal^t = \underset{\wb, \Xcal: |\Xcal| \leq k}\argmin \mbox{Err}_{\lambda}(\wb, \Xcal, L, L_T, \theta_t) 
  %\\
  %  &= \underset{\wb, \Xcal: |\Xcal| \leq k}\argmin \Vert \sum_{i \in \Xcal^t} w^t_i \nabla_{\theta}L_T^i(\theta_t) -  \nabla_{\theta}L(\theta_t)\Vert + \\ 
  %  & \hspace{1.7cm} \lambda \Vert\wb\Vert^2
    \label{eq1}
\end{aligned}
\end{equation}
In $\mbox{Err}_{\lambda}(\wb, \Xcal, L, L_T, \theta_t)$, the first term is the additional error term that adaptive subset algorithms have from the convergence analysis discussed in Section~\ref{adap-dss-framework}, and the second term is a squared l2 loss regularizer over the weight vector $\wb$ with a regularization coefficient $\lambda$ to prevent overfitting by discouraging the assignment of large weights to individual data instances or mini-batches. 
During data-selection, we select the weights $\wb^t$ and subset $\Xcal^t$ by optimizing equation~\eqref{eq1}. To this end, we define:
\begin{align}\label{eqn:ex}
    E_{\lambda}(\Xcal) = \min_{\wb} \mbox{Err}_{\lambda}(\wb, \Xcal, L, L_T, \theta_t)
\end{align}
Note that the optimization problem in Eq.~\eqref{eq1} is equivalent to solving the optimization problem  $\min_{\Xcal: |\Xcal| \leq k} E_{\lambda}(\Xcal)$. The detailed optimization algorithm is presented in Section~\ref{omp-algo}. 

\noindent \textbf{\textsc{Grad-Match} for mini-batch SGD: } We now discuss an alternative formulation of \model, specifically for mini-batch SGD. Recall from Theorem~\ref{thm:convergence-result} and optimization problem~\eqref{eq1}, that in the case of full gradient descent or SGD, we select a subset of data points for the data selection. However, mini-batch SGD is a combination of SGD and full gradient descent, where we randomly select a mini-batch and compute the full gradient on that mini-batch. To handle this, we consider a variant of \model, which we refer to as \modelpb. Here we select a subset of mini-batches by matching the weighted sum of mini-batch training gradients to the full training loss (or validation loss) gradients. Once we select a subset of mini-batches, we train the neural network on the mini-batches, weighing each mini-batch by its corresponding weight. Let $B$ be the batch size, $b_n = n/B$ as the total number of mini-batches, and $b_k = k/B$ as the number of batches to be selected. Let $\nabla_{\theta} L_T^{B_1}(\theta_t), \cdots, \nabla_{\theta} L_T^{B_{b_n}}(\theta_t)$ denote the mini-batch gradients. The optimization problem is then to minimize $E^B_{\lambda}(\Xcal)$, which is defined as:
\begin{align*}
    E^B_{\lambda}(\Xcal) = \min_{\wb} \lVert \sum_{i \in \Xcal} w^i_t \nabla_{\theta} L_T^{B_i}(\theta_t) - \nabla_{\theta} L(\theta_t) \rVert + \lambda \lVert \wb \rVert^2
\end{align*}
The constraint now is $|\Xcal| \leq b_k$, where $\Xcal$ is a subset of mini-batches instead of being a subset of data points. The use of mini-batches considerably reduces the number of selection rounds during the OMP algorithm by a factor of $B$, resulting in $B\times$ speed up. In our experiments, we compare the performance of \model\ and \modelpb\  and show that \modelpb\ is considerably more efficient while being comparable in performance. \modelpb\ is a simple modification to lines 3 and 5 of Algorithm~\ref{alg:grad-match}, where we send the mini-batch gradients instead of individual gradients to the orthogonal matching pursuit (OMP) algorithm (discussed in the next section). Further, we use the subset of mini-batches selected directly without any additional shuffling or sampling in our current experiments. We will consider augmenting the selected mini-batch subsets with additional shuffling or including new mini-batches with augmented images in our future work. 

\begin{algorithm}
\caption{\model\ Algorithm}
\label{alg:algorithm1}
\small{
\begin{algorithmic}[1]
\REQUIRE Train set: ${\mathcal U}$; validation set: ${\mathcal V}$; initial subset: $\Xcal^{(0)}$; subset size: $k$; TOL: $\epsilon$; initial params: $\theta_{0}$; learning rate: $\alpha$; total epochs: $T$, selection interval: $R$, Validation Flag: \mbox{isValid}, Batchsize: $B$
\FOR {epochs $t$ in $1, \cdots, T$}
        \IF{$(t \mbox{ mod } R == 0)$ and $(\mbox{isValid} == 1)$}
            \STATE $\Xcal^{t}, \wb^t = \operatorname{OMP}(L_T, L_V,\theta_{t}, k, \epsilon)$
            \ELSIF{$(t \mbox{ mod } R == 0)$ and $(\mbox{isValid} == 0)$}
            \STATE $\Xcal^{t}, \wb^t = \operatorname{OMP}(L_T, L_T, \theta_{t}, k, \epsilon)$ 
        \ELSE
            \STATE $\Xcal^{t} = \Xcal^{t-1}$
        \ENDIF
        \STATE $\theta_{t+1} = \mbox{BatchSGD}(\Xcal^{t}, \wb^t, \alpha, L_T, B, \mbox{Epochs} = 1)$
        \ENDFOR
\STATE Output final model parameters $\theta_{T}$
\end{algorithmic}}
\label{alg:grad-match}
\end{algorithm}

\subsection{Orthogonal Matching Pursuit (OMP) algorithm}\label{omp-algo}
We next study the optimization algorithm for solving equation~\eqref{eq1}. Our objective is to minimize $E_{\lambda}(\Xcal)$ subject to the constraint $\Xcal: |\Xcal| \leq k$. We can also convert this into a maximization problem. For that, define: $F_{\lambda}(\Xcal) = L_{\max} - \min_{\wb} \mbox{Err}_{\lambda}(\Xcal,\wb, L, L_T, \theta_t)$. Note that we minimize $E_{\lambda}(\Xcal)$ subject to the constraint $\Xcal: |\Xcal| \leq k$ until $E_{\lambda}(\Xcal) \le \epsilon$, where $\epsilon$ is the tolerance level.
\begin{algorithm}
\caption{OMP}
\small{
  \begin{algorithmic}
  \REQUIRE Training loss $L_T$, target loss: $L$, current parameters: $\theta$, regularization coefficient: $\lambda$, subset size: $k$, tolerance: $\epsilon$
%\STATE \textbf{Input: } $E(.), \lambda, k$ OR $\epsilon$
%\STATE \textbf{Output: } $\Xcal, {\wb}$
\STATE $\Xcal\leftarrow \emptyset$
 \STATE ${r}\leftarrow \nabla_w \mbox{Err}_{\lambda}(\Xcal,\wb,L,L_T,\theta)|_{{\wb}={0}}$
 \WHILE{$|\Xcal| \leq k$ and $E_{\lambda}(\Xcal) \geq \epsilon$}
 \STATE $e = \text{argmax}_j |r_j|$
 \STATE $\Xcal\leftarrow \Xcal \cup \{e\}$
 \STATE ${\wb}\leftarrow \text{argmin}_{{\wb}}\mbox{Err}_{\lambda}(\Xcal,\wb,L,L_T,\theta)$
 \STATE ${r}\leftarrow \nabla_{\wb} \mbox{Err}_{\lambda}(\Xcal,\wb,L,L_T,\theta)$
 \ENDWHILE
 \STATE \textbf{return} $\Xcal, \ {\wb}$
\end{algorithmic}}
\label{alg:algo1}
\end{algorithm}
Note that minimizing $E_{\lambda}$ is equivalent to maximizing $F_{\lambda}$. The following result shows that $F_{\lambda}$ is weakly submodular.
\begin{theorem}\label{thm:approx-submod-result}
\small
    If $|\Xcal|\leq k$ and $\max_{i} ||\nabla_{\theta} L^i _T(\theta_t)||_{2} < \nabla_{\max}$, then $F_{\lambda}(\Xcal)$ is $\gamma$-weakly submodular, with
    $\gamma \geq \frac{\lambda}{\lambda+k\nabla_{\max} ^2  }$
\end{theorem}
We present the proof in Appendix~\ref{app-approx-submod-result}. Recall that a set function $F: 2^{[n]} \rightarrow \mathbb{R}$ is $\gamma$-weakly submodular~\cite{gatmiry2018non,das2011submodular} if $F(j | S) \geq \gamma F(j | T), S \subseteq T \subseteq [n]$. Since $F_{\lambda}(\Xcal)$ is approximately submodular~\cite{das2011submodular}, a greedy algorithm~\cite{nemhauser1978analysis,das2011submodular,elenberg2018restricted} admits a $(1-\exp(-\gamma))$ approximation guarantee. While the greedy algorithm is very appealing, it needs to compute the gain $F_{\lambda}(j | \Xcal)$, $O(nk)$ number of times. Since  computation of each gain involves solving a least squares problem, this step will be computationally expensive, thereby defeating the purpose of data selection. To address this issue, we consider a slightly different algorithm, called the orthogonal matching
pursuit (OMP) algorithm, studied in~\cite{elenberg2018restricted}. We present OMP in Algorithm~\ref{alg:algo1}. %, and refer to equation~\eqref{eq:errlam} for the expression $\mbox{Err}_{\lambda}$. 
In the corollary below, we provide the approximation guarantee for Algorithm~\ref{alg:algo1}. 
\begin{corollary}
\small
Algorithm~\ref{alg:algo1}, when run with a cardinality constraint $|\Xcal| \leq k$ returns a $1-\exp\left(\frac{-\lambda}{\lambda + k\nabla_{\max} ^2}\right)$ approximation for maximizing $F_{\lambda}(\Xcal)$ with  $\Xcal: |\Xcal| \leq k$.
\end{corollary}
We can also find the minimum sized subset such that the resulting error $E_{\lambda}(\Xcal) \leq \epsilon$. We note that this problem is essentially: $\min_{\Xcal} |\Xcal|$ such that $F_{\lambda}(\Xcal) \geq L_{\max} - \epsilon$. This is a weakly submodular set cover problem, and the following theorem shows that a greedy algorithm~\cite{wolsey1982analysis} as well as OMP (Algorithm~\ref{alg:algo1}) with a stopping criterion $E_{\lambda}(\Xcal) \leq \epsilon$ achieve the following approximation bound:
\begin{theorem}\label{weak-submod-setcover}
\small
If the function $F_{\lambda}(\Xcal)$ is $\gamma$-weakly submodular, $\Xcal^*$ is the optimal subset and $\max_{i} ||\nabla_{\theta} L^i _T(\theta_t)||_{2} < \nabla_{\max}$, (both) the greedy algorithm and OMP (Algorithm~\ref{alg:algo1}), run with stopping criteria  $E_{\lambda}(\Xcal) \leq \epsilon$ result in sets $\Xcal$ such that $|\Xcal| \leq \frac{|\Xcal^*|}{\gamma} \log\left(\frac{L_{\max}}{\epsilon}\right)$ where $L_{\max}$ is an upper bound of $F_{\lambda}$ .
\end{theorem}
The proof of this theorem is in Appendix~\ref{app-weak-submod-setcover}. Finally, we derive the convergence result for \model\ as a corollary of Theorem~\ref{thm:convergence-result}. In particular, assume that by running OMP, we can achieve sets $\Xcal^t$ such that $E_{\lambda}(\Xcal^t) \leq \epsilon$, for all $t = 1, \cdots, T$. If $L$ is Lipschitz continuous, we obtain a convergence bound of $\min_t L(\theta_t) - L(\theta^*) \leq \frac{D\sigma_T}{\sqrt{T}} + D\epsilon$. In the case of smooth or strongly convex functions, the result can be improved to $O(1/T)$. For example, with smooth functions, we have: $\min_t L(\theta_t) - L(\theta^*) \leq \frac{D^2 \mathcal{L}_T + 2\beta_T}{2T} + D\epsilon$. More details can be found in Appendix~\ref{app-omp-convergence}.

%The following Lemma shows the convergence of \model\ (proof in Appendix XX):
%\begin{lemma}\label{omp-convergence}
%Suppose there exists an $\epsilon > 0$, such that by running \model, we obtain subsets $\Xcal_t$ such that $E_{\lambda}(\Xcal_t) \leq \epsilon$, for all $t = 1, \cdots, T$, then the following convergence result holds: $\min_t L(\theta_t) - L(\theta^*) \leq \frac{D\sigma_T}{\sqrt{T}} + D\epsilon$
%\end{lemma}
%Note that the result of Lemma~\ref{omp-convergence} is very similar to the result in Lemma~\ref{fac-loc-convergence}. However there is one key difference. OMP directly optimizes the error term, while the facility location data selection minimizes an upper bound of the error. As a result, facility location based data selection (\textsc{Craig}) could require larger subset sizes to achieve the desired $\epsilon$ error in the gradient difference, compared to what \model\ achieves. In fact, we see empirically that \model\ significantly outperforms \textsc{Craig} particularly at smaller subset sizes.

\subsection{Connections to existing work}\label{connections}
Next, we discuss connections of \model\ to existing approaches such as \textsc{Craig} and \textsc{Glister}, and contrast the resulting theoretical bounds. Let $\Xcal$ be a subset of $k$ data points from the training or validation set. Consider the expression for loss $L(\theta) = \sum_{i \in \Wcal} L(x_i, y_i, \theta)$ so that $L = L_T$ when $\Wcal = \Ucal$ and $L = L_V$ when $\Wcal = \Vcal$. Define $\hat{E}(\Xcal)$ to be:
\begin{align}
    \hat{E}(\Xcal) & =  \sum_{i\in \Wcal} \min_{j \in \Xcal} \| \nabla_{\theta} L^i(\theta_t) - \nabla_{\theta} L_T^j(\theta_t) \| 
    %\nonumber \\ & = \sum_{i\in W}  \| \nabla_{\theta} L^i(\theta_t) - \nabla_{\theta} L_T^{\pi_l^i}(\theta_t) \|
\end{align}
Note that $\hat{E}(\Xcal)$ is an upper bound for $E(\Xcal)$~\cite{mirzasoleiman2020coresets}:
\begin{align}
\label{eq:min_upper1}
E(\Xcal)& = \min_{\wb} \mbox{Err}( {\wb}, \Xcal, L, L_T, \theta_t) \leq \hat{E}(\Xcal)
\end{align}
Given the set $\Xcal^t$ obtained by optimizing $\hat{E}$, the weight $\wb^t_j$ associated with the $j^{th}$ point in the subset $\Xcal^t$ will be: $\wb^t_j=\sum_{i\in \Wcal} \mathbb{I} \big[ j = {\arg\min_{s \in \Xcal^t}} \| \nabla_{\theta} L_T^i(\theta_t) - \nabla_{\theta} L^{s}(\theta_t) \|]$. Note that we can minimize both sides with respect to a cardinality constraint $\Xcal: |\Xcal| \leq k$; the right hand side of eqn.~\eqref{eq:min_upper1} is minimized when $\Xcal$ is the set of $k$ {\em medoids} \cite{kaufman1987clustering} for all the components in the gradient space. In Appendix~\ref{app-craig-details}, we prove the inequality~\eqref{eq:min_upper1}, and also discuss the maximization version of this problem and how it relates to facility location. Similar to \model, we also use the mini-Batch version for CRAIG~\cite{mirzasoleiman2020coresets}, which we refer to as \textsc{CraigPB}. Since \textsc{CraigPB} operates on a much smaller groundset (of mini-batches), it is much more efficient than the original version of \textsc{Craig}.

Next, we point out that a  convergence bound, very similar to \model\ can be shown for \textsc{Craig} as well. This result is \emph{new and different} from the one shown in~\cite{mirzasoleiman2020coresets} since it is for full GD and SGD, and not for incremental gradient descent algorithms discussed in~\cite{mirzasoleiman2020coresets}. If the subsets $\Xcal^t$ obtained by running \textsc{Craig} satisfy $\hat{E}(\Xcal^t) \leq \epsilon, \forall t = 1, \cdots, T$, we can show $O(1/\sqrt{T})$ and $O(1/T)$ bounds (depending on the nature of the loss) with an additional $\epsilon$ term. However, the bounds obtained for \textsc{Craig} will be weaker than the one for \textsc{Grad-Match} since $\hat{E}$ is an upper bound of $E$. Hence, \emph{to achieve the same error, a potentially larger subset could be required for \textsc{Craig} in comparison to \textsc{Grad-Match}}. 

Finally, we also connect \textsc{Grad-Match} to \textsc{Glister}~\cite{killamsetty2021glister}.  While the setup of \textsc{Glister} is different from \textsc{Grad-Match} since it directly tries to optimize the generalization error via a bi-level optimization, the authors use a Taylor-series approximation to make \textsc{Glister} efficient. The Taylor-series approximation can be viewed as being similar to maximizing the dot product between $\sum_{i\in \Xcal} \nabla_{\theta} L_T ^i (\theta_t)$ and $\nabla_{\theta} L(\theta_t)$. Furthermore, \textsc{Glister} does not consider a weighted sum the way we do, and is therefore slightly sub-optimal. In our experiments, we show that \model\  outperforms \textsc{Craig} and \textsc{Glister} across a number of deep learning datasets.
% \vspace{-2ex}
\section{Speeding up \textsc{Grad-Match}}
\label{speedups-further}
In this section, we propose several implementational and practical tricks to make \model\ scalable and efficient (in addition to those discussed above). In particular, we will discuss various approximations to \model\ such as running OMP per class, using the last layer of the gradients, and warm-start to the data selection.

% \vspace{-2ex}
\paragraph{Last-layer gradients. } 
The number of parameters in modern deep models is very large, leading to very high dimensional gradients. The high dimensionality of gradients slows down OMP, thereby decreasing the efficiency of subset selection. To tackle this problem,  we adopt a last-layer gradient approximation similar to~\cite{ash2020deep,mirzasoleiman2020coresets,killamsetty2021glister} by only considering the last layer gradients for neural networks in \model. This simple trick significantly improves the speed of \model\ and other baselines.
%last-layer gradient approximation reduces the dimensionality of the model gradients and increases the OMP algorithm's speed. 

\noindent \textbf{Per-class and per-gradient approximations of \model: } 
To solve the \model\ optimization problem, we need to store the gradients of all instances in memory, leading to high memory requirements for large datasets. In order to tackle this problem, we consider  per-class and per-gradient approximation. We solve multiple gradient matching problems using per-class approximation - one for each class by only considering the data instances belonging to that class. The per-class approximation was also adopted in~\cite{mirzasoleiman2020coresets}. To further reduce the memory requirements, we additionally adopt the per-gradient approximation 
%in addition to the per-class approximation 
by considering only the corresponding last linear layer's gradients for each class. The per-gradient and per-class approximations not only reduce the memory usage, but also significantly speed up (reduce running time of) the data selection itself. By default, we use the per-class and per-gradient approximation, with the last layer gradients and we will call this algorithm \model. We do not need 
these approximations for \modelpb\ since it is on a much smaller ground-set (mini-batches instead of individual items).
%In our experiments, we run ablation studies to study the loss in approximation and efficiency gained with these approximations. 

\begin{figure*}[t]
\centering
\includegraphics[width = 16cm, height=0.8cm]{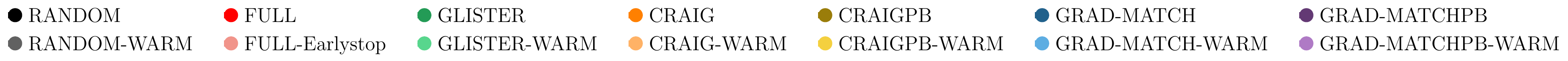}
\centering
\hspace{-0.6cm}
\begin{subfigure}[b]{0.225\textwidth}
\centering
\includegraphics[width=3.1cm, height=2.5cm]{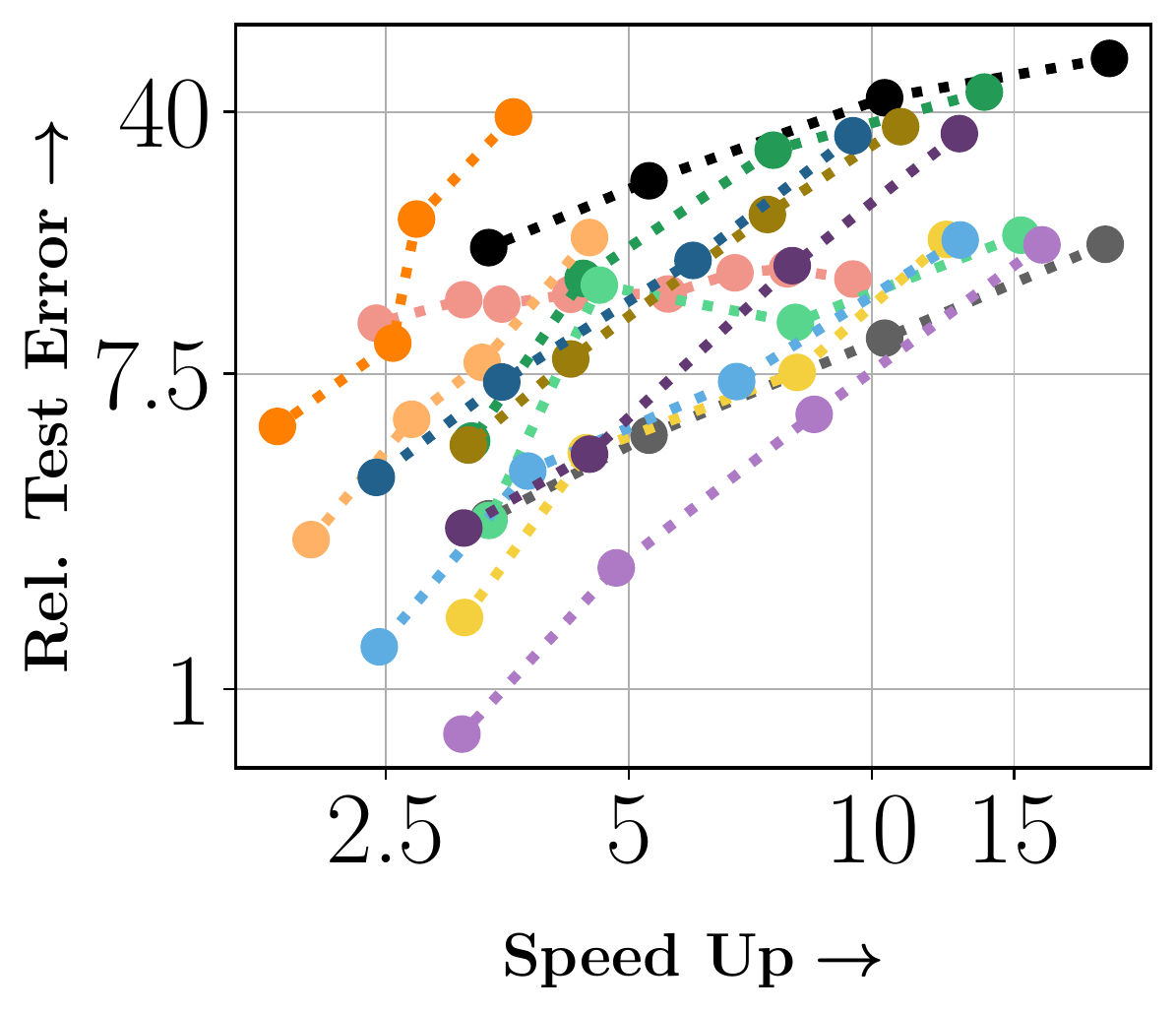}
\caption*{$\underbracket[1pt][1.0mm]{\hspace{3.2cm}}_{\substack{\vspace{-4.0mm}\\
\colorbox{white}{(a) \scriptsize CIFAR100}}}$}
\phantomcaption
\label{fig:CIFAR100}
\end{subfigure}
\begin{subfigure}[b]{0.225\textwidth}
\centering
\includegraphics[width=3.1cm, height=2.5cm]{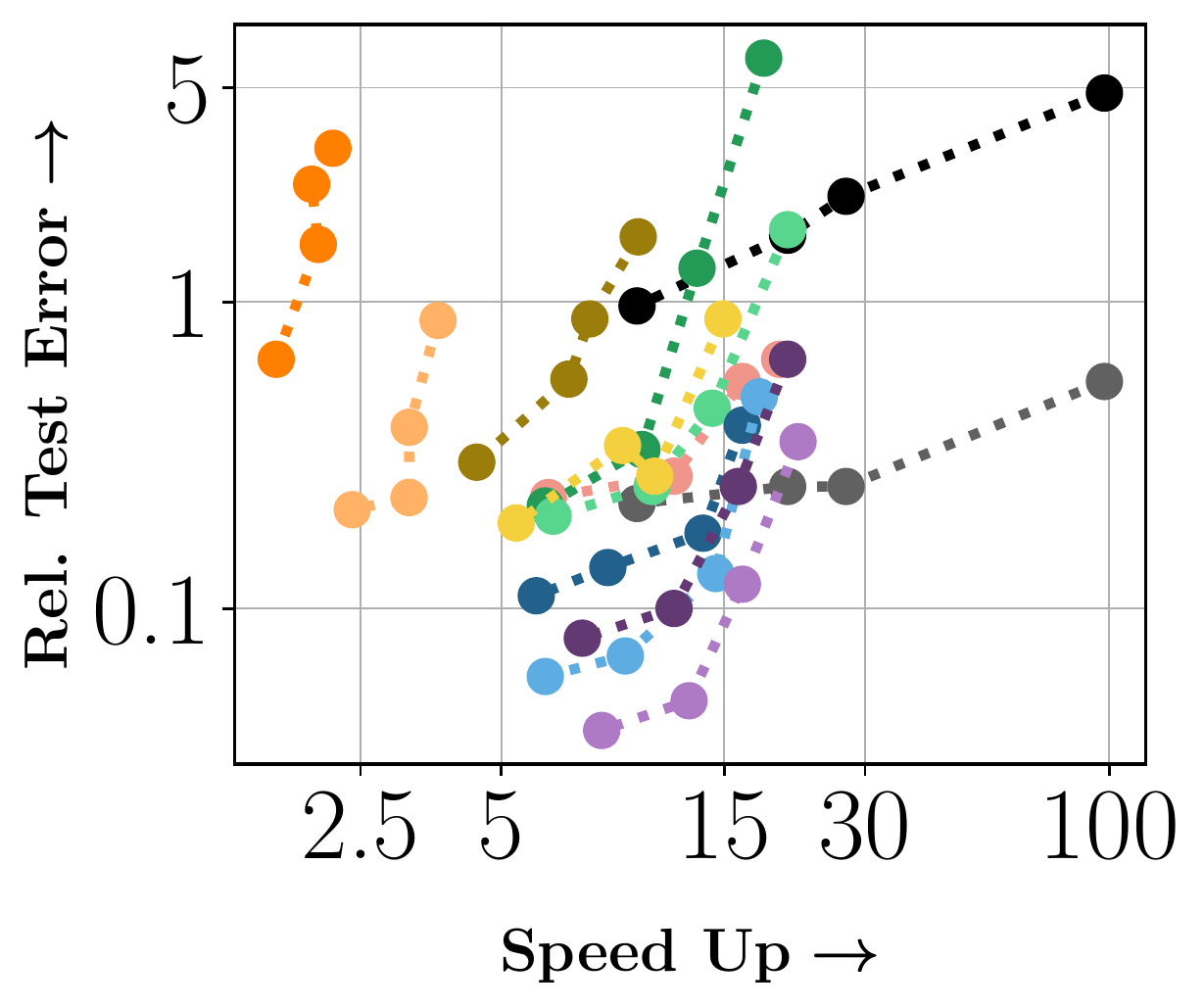}
\caption*{$\underbracket[1pt][1.0mm]{\hspace{3.2cm}}_{\substack{\vspace{-4.0mm}\\
\colorbox{white}{(b) \scriptsize MNIST}}}$}
\phantomcaption
\label{fig:MNIST}
\end{subfigure}
\begin{subfigure}[b]{0.225\textwidth}
\centering
\includegraphics[width=3.1cm, height=2.5cm]{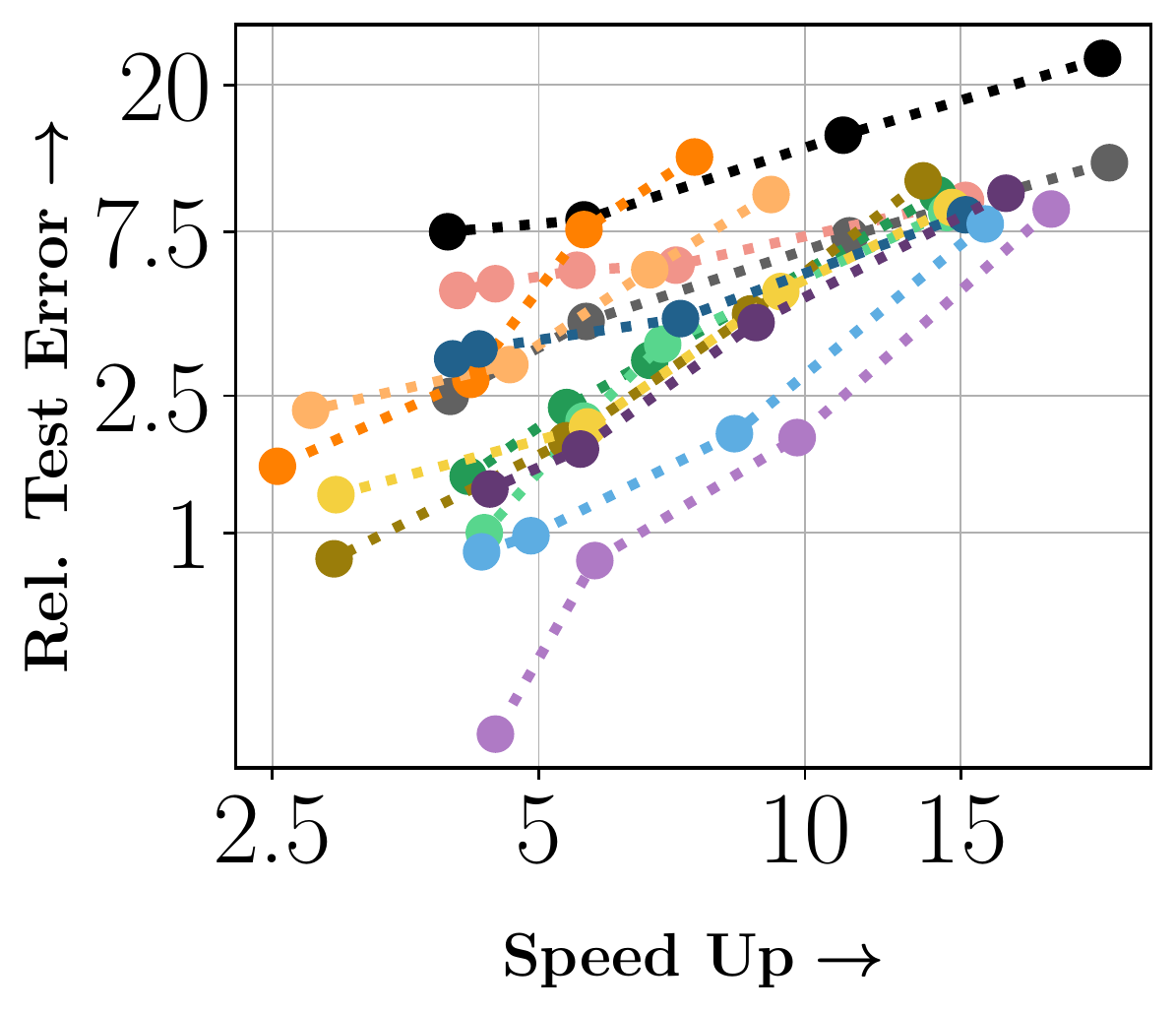}
\caption*{$\underbracket[1pt][1.0mm]{\hspace{3.2cm}}_{\substack{\vspace{-4.0mm}\\
\colorbox{white}{(c) \scriptsize CIFAR10}}}$}
\phantomcaption
\label{fig:CIFAR10}
\end{subfigure}
\begin{subfigure}[b]{0.225\textwidth}
\centering
\includegraphics[width=3.1cm, height=2.5cm]{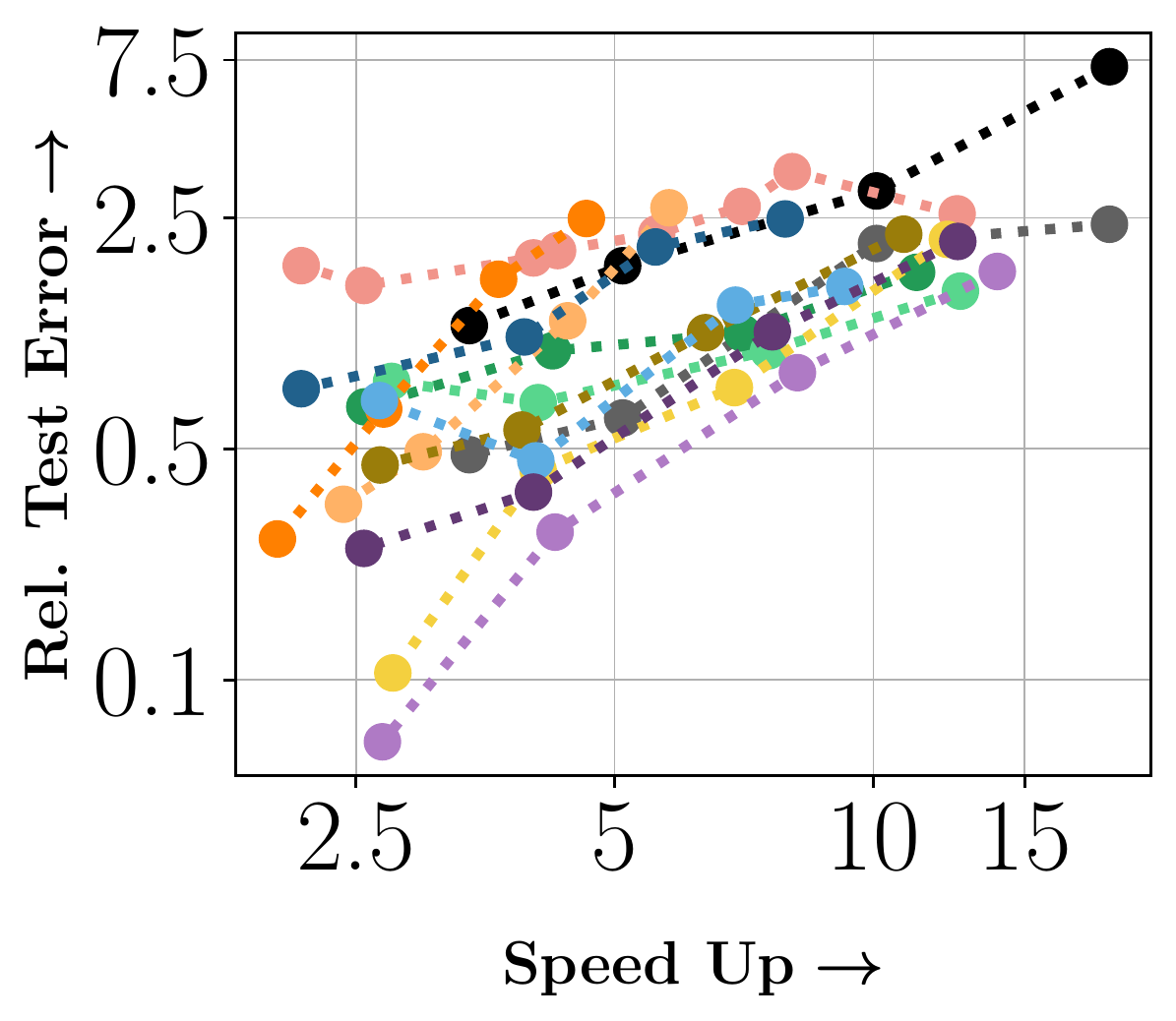}
\caption*{$\underbracket[1pt][1.0mm]{\hspace{3.2cm}}_{\substack{\vspace{-4.0mm}\\
\colorbox{white}{(d) \scriptsize SVHN}}}$}
\phantomcaption
\label{fig:SVHN}
\end{subfigure}
\begin{subfigure}[b]{0.225\textwidth}
\centering
\includegraphics[width=3.1cm, height=2.5cm]{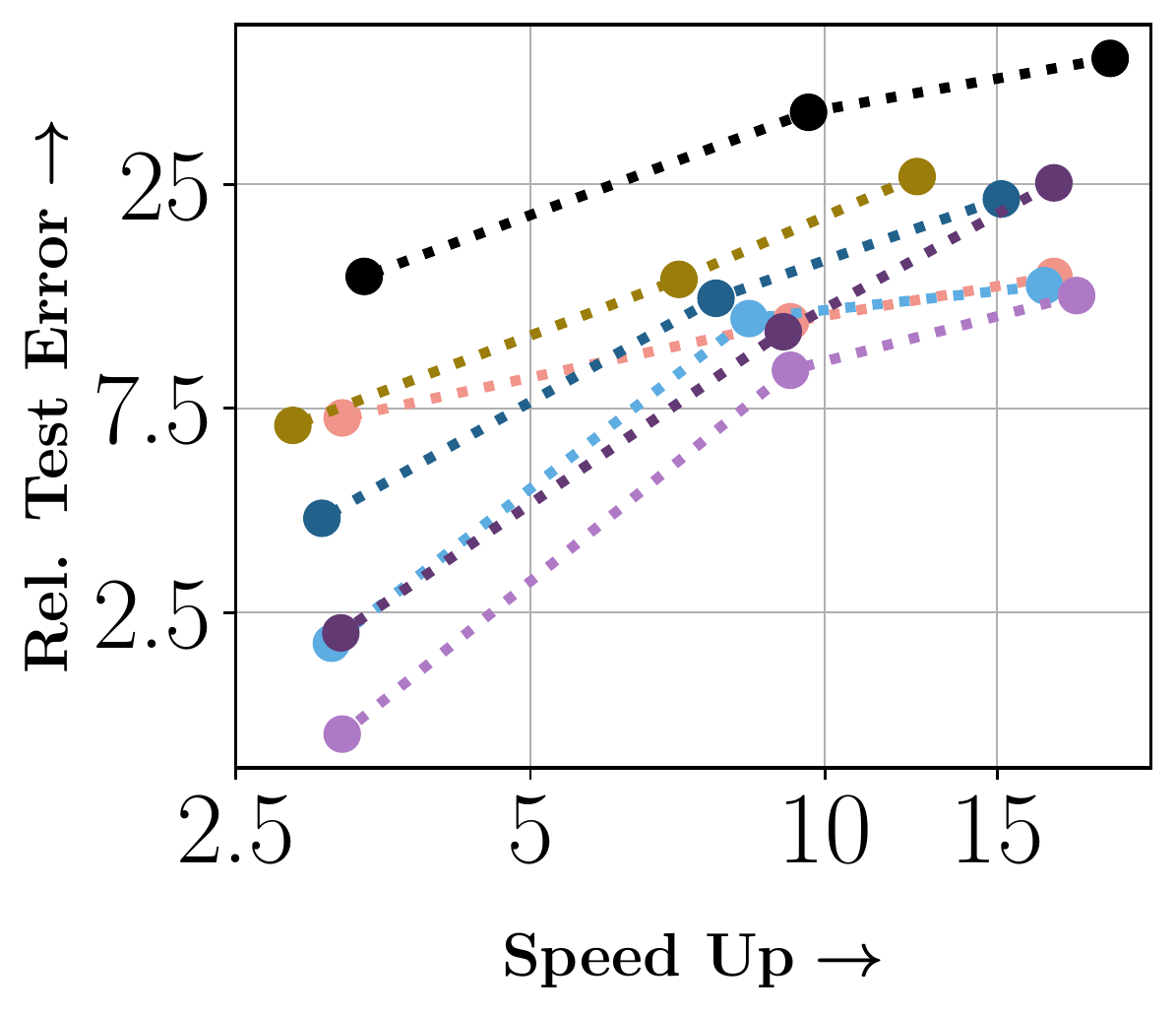}
\caption*{$\underbracket[1pt][1.0mm]{\hspace{3.2cm}}_{\substack{\vspace{-4.0mm}\\
\colorbox{white}{(e) \scriptsize ImageNet}}}$}
\phantomcaption
\label{fig:imagenet}
\end{subfigure}
\hspace{-0.6cm}
\begin{subfigure}[b]{0.225\textwidth}
\centering
\includegraphics[width=3.1cm, height=2.5cm]{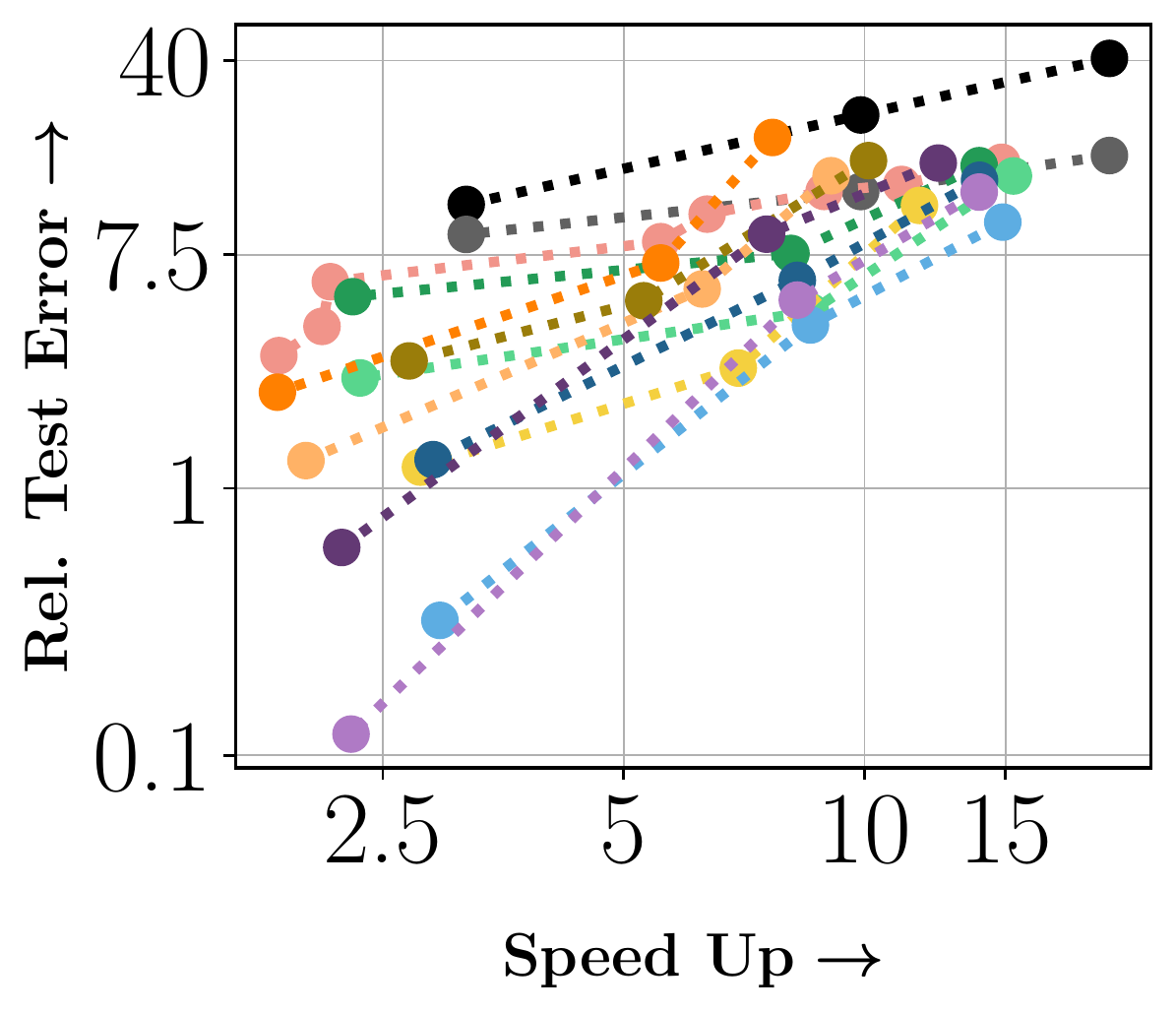}
\caption*{$\underbracket[1pt][1.0mm]{\hspace{3.5cm}}_{\substack{\vspace{-4.0mm}\\
\colorbox{white}{(f) \scriptsize CIFAR10 Imbalance}}}$}
\phantomcaption
\label{fig:CIFAR10Imb}
\end{subfigure}
\begin{subfigure}[b]{0.23\textwidth}
\centering
\includegraphics[width=3.1cm, height=2.5cm]{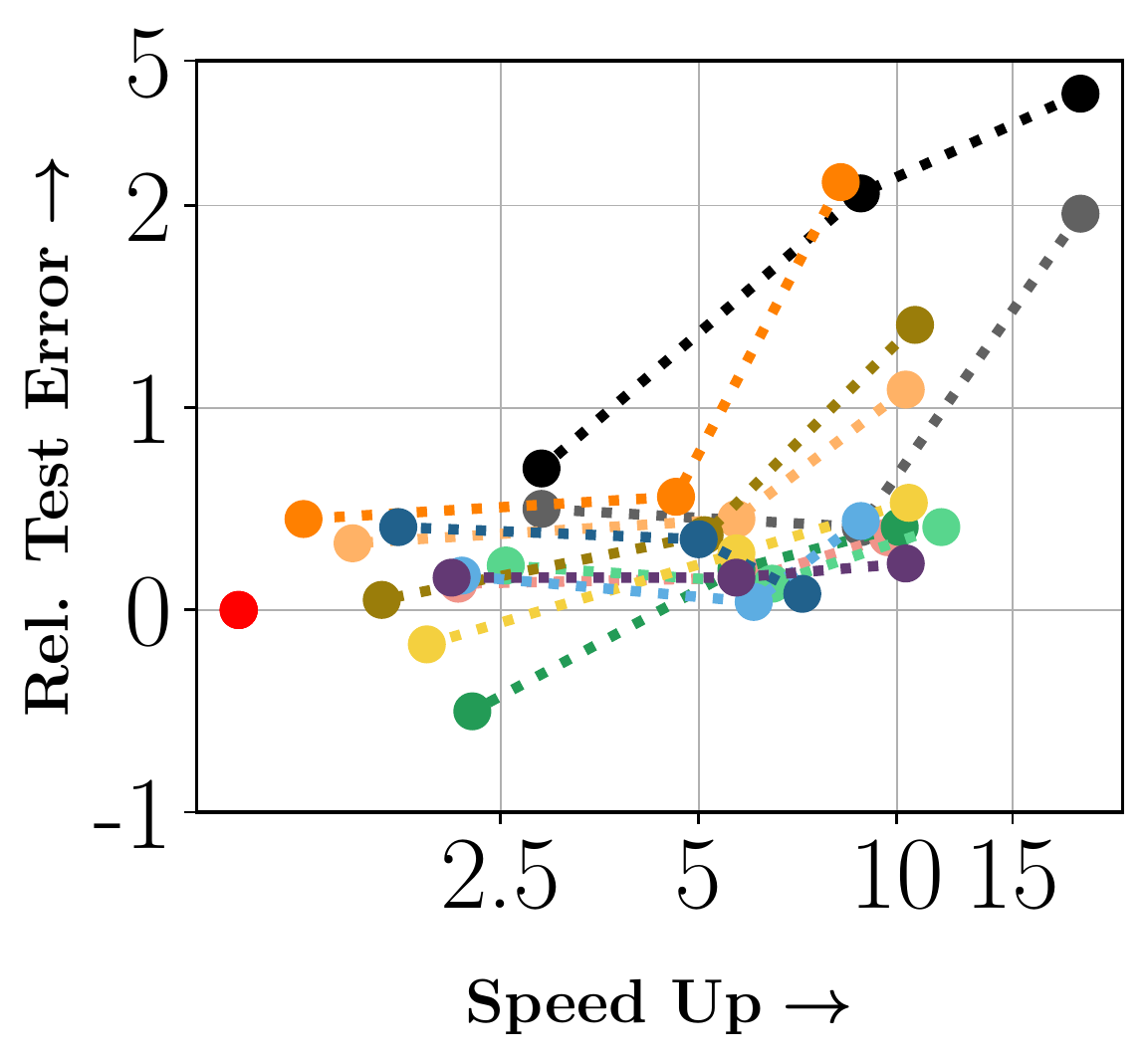}
\caption*{$\underbracket[1pt][1.0mm]{\hspace{3.5cm}}_{\substack{\vspace{-4.0mm}\\
\colorbox{white}{(g) \scriptsize MNIST Imbalance}}}$}
\phantomcaption
\label{fig:MNISTImb}
\end{subfigure}
\begin{subfigure}[b]{0.225\textwidth}
\centering
\includegraphics[width=3.1cm, height=2.5cm]{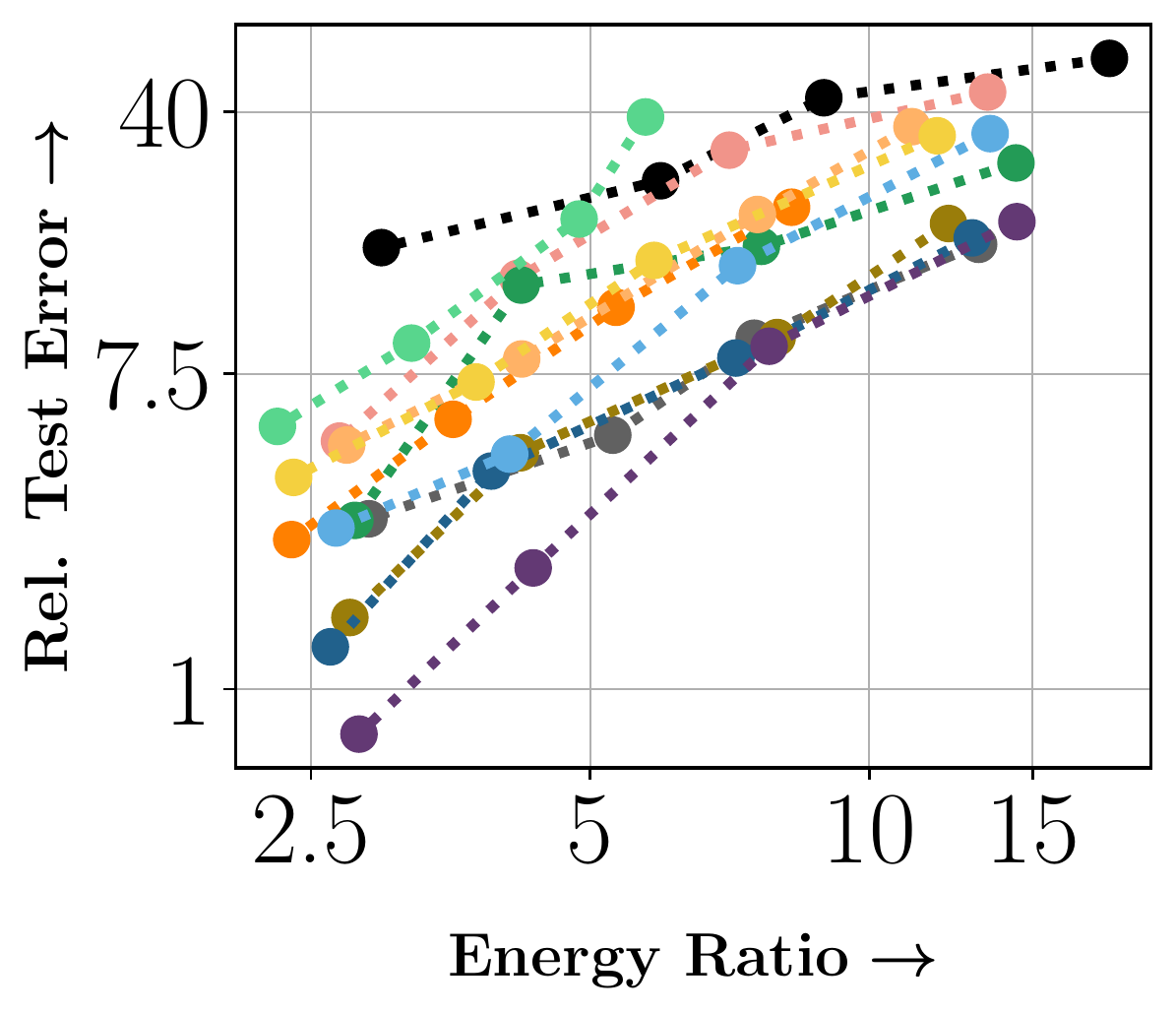}
\caption*{$\underbracket[1pt][1.0mm]{\hspace{3.5cm}}_{\substack{\vspace{-4.0mm}\\
\colorbox{white}{(h) \scriptsize CIFAR100 Energy}}}$}
\phantomcaption
\label{fig:CIFAR100Energy}
\end{subfigure}
\begin{subfigure}[b]{0.225\textwidth}
\centering
\includegraphics[width=3.1cm, height=2.5cm]{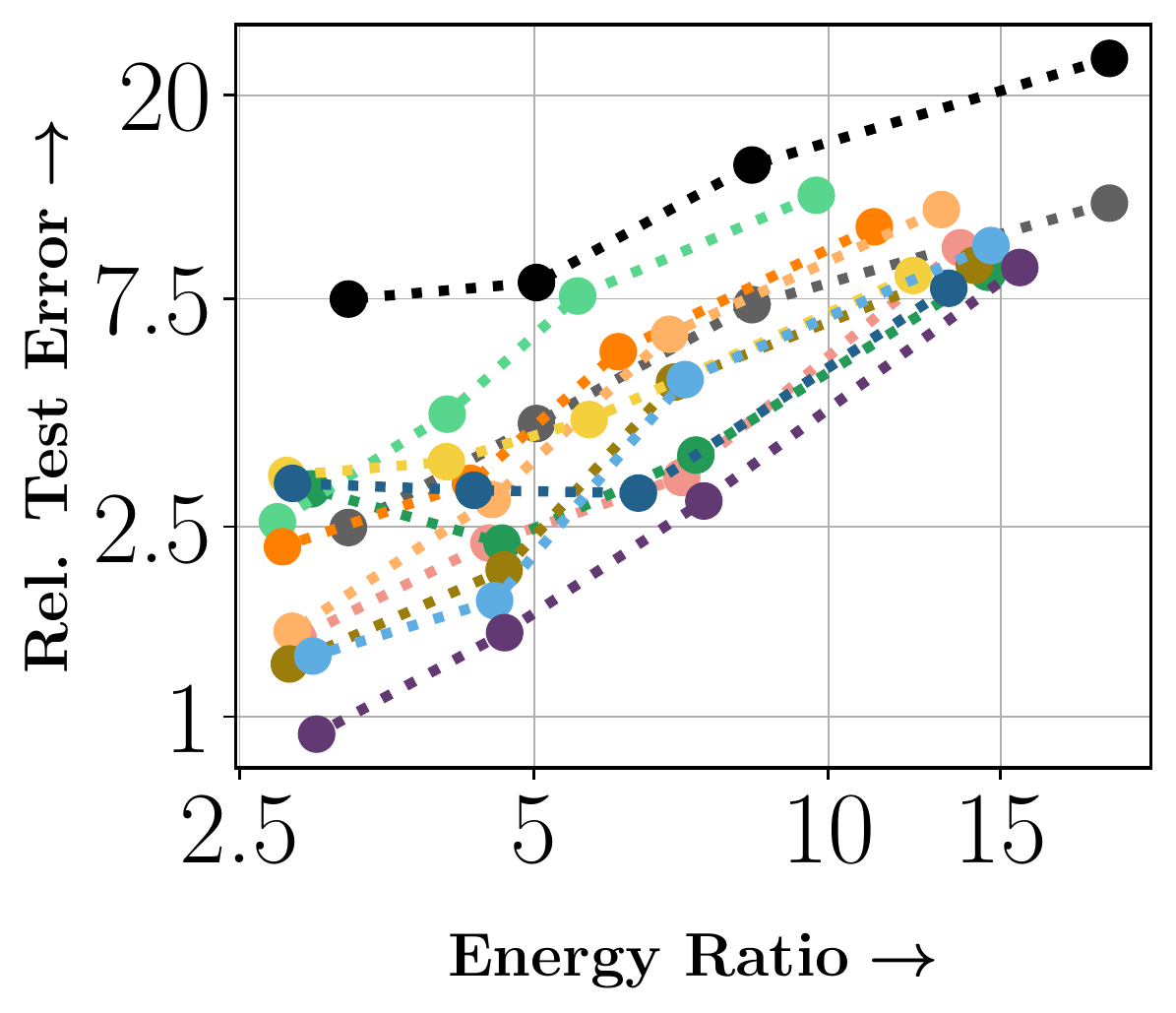}
\caption*{$\underbracket[1pt][1.0mm]{\hspace{3.5cm}}_{\substack{\vspace{-4.0mm}\\
\colorbox{white}{(i) \scriptsize CIFAR10 Energy}}}$}
\phantomcaption
\label{fig:CIFAR10Energy}
\end{subfigure}
\begin{subfigure}[b]{0.225\textwidth}
\centering
\includegraphics[width=3.1cm, height=2.5cm]{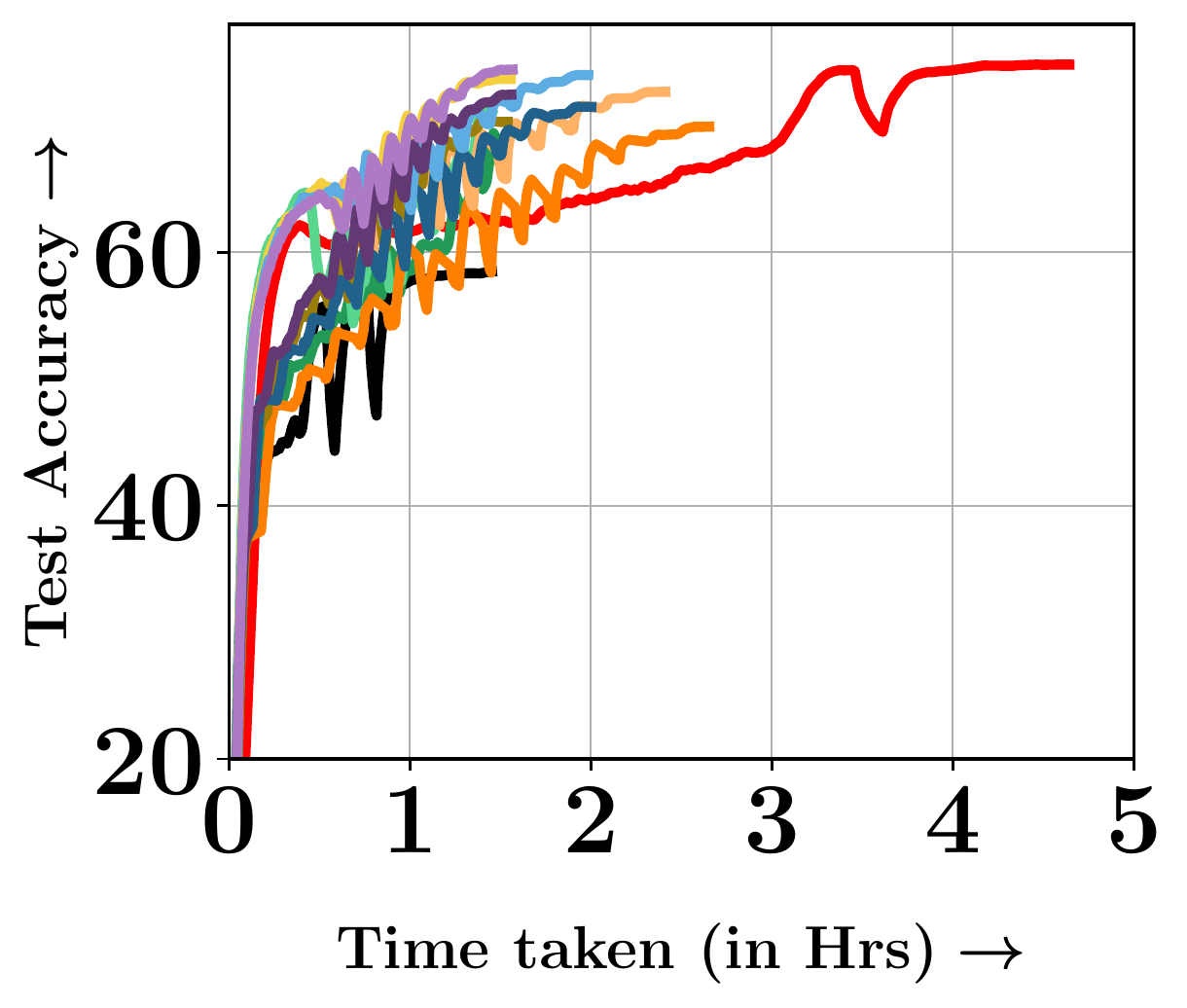}
\caption*{$\underbracket[1pt][1.0mm]{\hspace{3.5cm}}_{\substack{\vspace{-4.0mm}\\
\colorbox{white}{(j) \scriptsize CIFAR100 Convergence}}}$}
\phantomcaption
\label{fig:CIFAR100Conv}
\end{subfigure}
\begin{subfigure}[b]{0.25\textwidth}
\centering
\includegraphics[width=3.1cm, height=2.5cm]{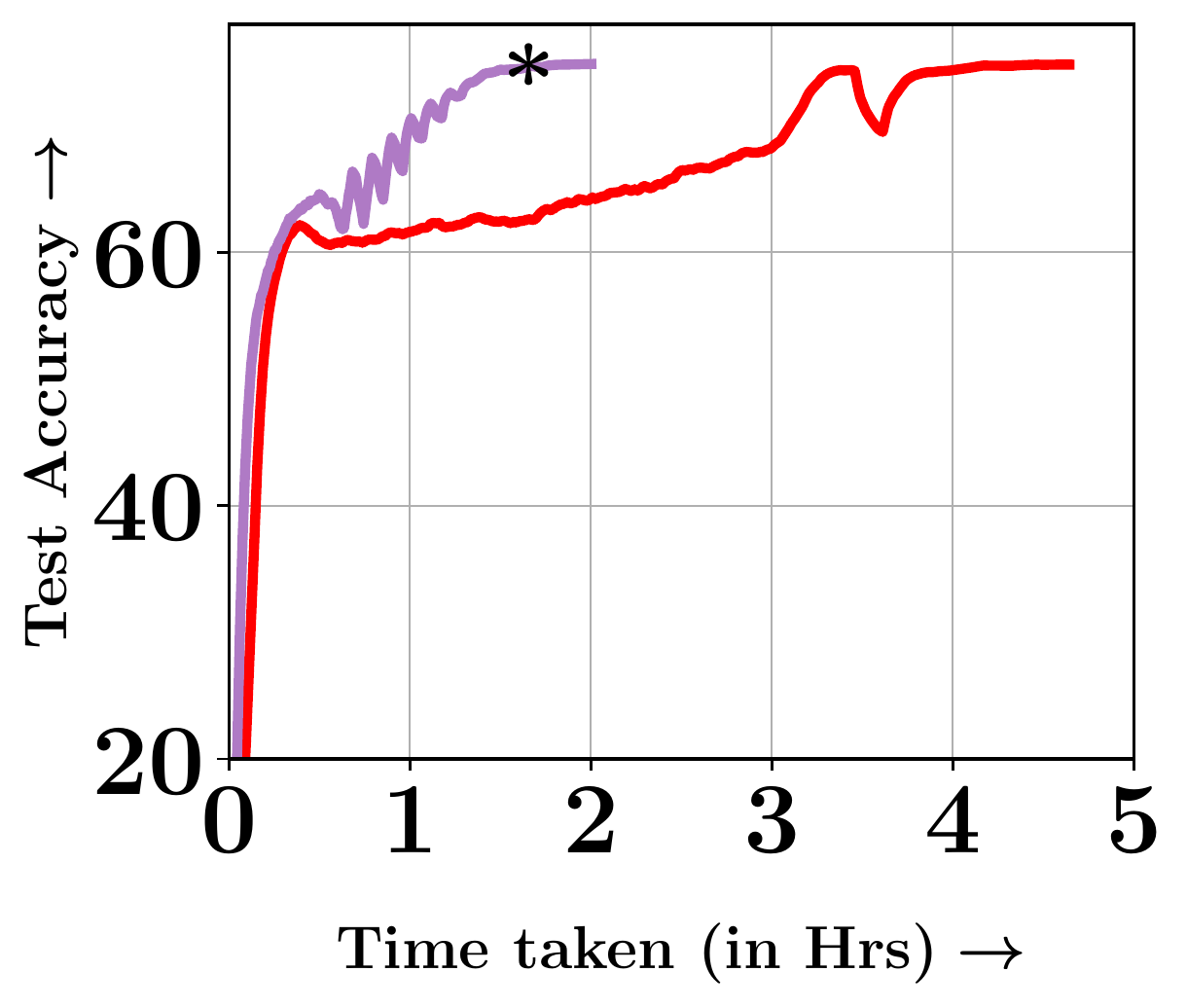}
\caption*{$\underbracket[1pt][1.0mm]{\hspace{4.2cm}}_{\substack{\vspace{-4.0mm}\\
\colorbox{white}{(k) {\scriptsize  CIFAR100 Ext. Convergence}}}}$}
\phantomcaption
\label{fig:CIFAR100Ext}
\end{subfigure}
\begin{subfigure}[b]{0.225\textwidth}
\centering
\includegraphics[width=3.5cm, height=2.5cm]{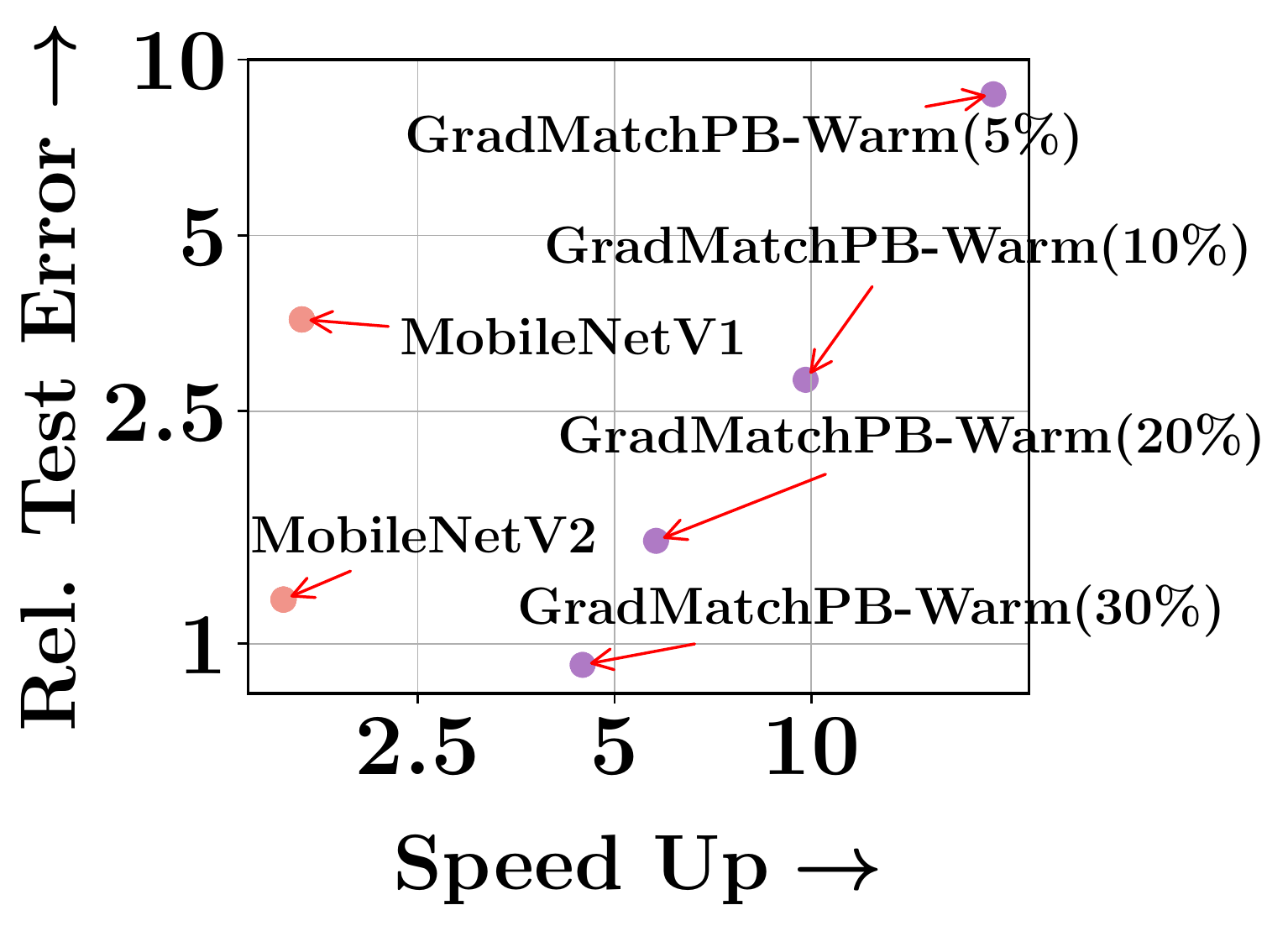}  
\caption*{$\underbracket[1pt][1.0mm]{\hspace{3.7cm}}_{\substack{\vspace{-4.0mm}\\
\colorbox{white}{(l) {\scriptsize CIFAR10 (MobileNet)}}}}$}
\phantomcaption
\label{fig:CIFAR10_MobileNet}
\end{subfigure}
\caption{{Sub-figures (a-g) show speedup vs relative error in \% tradeoff (\textbf{both log-scale}) of different algorithms. In each scatter plot, smaller subsets are on the left, and larger ones are on the right. Results are shown for (a) CIFAR-100, (b) MNIST, (c) CIFAR-10, (d) SVHN, (e) ImageNet, (f) CIFAR-10 imbalance, and (g) MNIST imbalance. Sub-figures (h, i) show energy gains vs. relative error for CIFAR-100 and CIFAR-10. Sub-figure (j) shows a convergence plot of different strategies at 30\% subset of CIFAR-100. Sub-figure (k) shows an extended convergence plot of \textsc{Grad-MatchPB-Warm} at 30\% subset of CIFAR-100 by running it for more epochs. Sub-figure (l) shows results of \textsc{Grad-MatchPB-Warm} using ResNet18 and Full training using MobileNet-V1 and MobileNet-V2 models on the CIFAR-10 dataset. In every case, the speedups \& energy ratios are computed w.r.t full training. \emph{Variants of \textsc{Grad-Match} achieve best speedup-accuracy tradeoff (bottom-right in each scatter plot represents best speedup-accuracy tradeoff region) in almost all cases}}.}
\label{fig:dss_general_experiments}
\end{figure*}

\noindent \textbf{Warm-starting data selection: } For each of the algorithms we consider in this paper ({\em i.e.}, \model, \modelpb, \textsc{Craig}, \textsc{CraigPB}, and \textsc{Glister}), we also consider a warm-start variant, where we run $T_f$ epochs of full training. We set $T_f$ in a way such that the number of epochs $T_s$ with the subset of data is a fraction $\kappa$ of the total number of epochs, {\em i.e.}, $T_s = \kappa T$ and $T_f = \frac{T_s k}{n}$, where $k$ is the subset size. We observe that doing full training for the first few epochs helps obtain good \emph{warm-start} models, resulting in much better convergence. Setting $T_f$ to a large value yields results similar to the full training with early stopping (which we use as one of our baselines) since there is not enough data-selection. 

\noindent \textbf{Other speedups: }
We end this section by reiterating two implementation tricks already discussed in Section~\ref{grad-match-algo}, namely, doing data selection every $R$ epochs (in our experiments, we set $R = 20$, but also study the effect of the choice of $R$), and the per-batch (PB) versions of \textsc{Craig} and \textsc{Grad-match}.
% \vspace{-2ex}
\section{Experiments}
\label{exp-results}
Our experiments aim to demonstrate the stability and efficiency of \textsc{Grad-Match}. While in most of our experiments, we study the tradeoffs between accuracy and efficiency (time/energy), we also study the robustness of data-selection under class imbalance. For most data selection experiments, we use the full loss gradients ({\em i.e.}, $L = L_T$). As an exception, in the case of class imbalance,  following~\cite{killamsetty2021glister}, we use $L = L_V$ ({\em i.e.}, we assume access to a clean validation set). 
%We focus on two settings for data selection: a) standard data selection, b) data selection with class imbalance. In case a) we used the full loss gradients ({\em, i.e.}, $L = L_T$), while for the case b) we use the validation loss gradients for generalization. We also assume that we have a clean validation set in the case of class imbalance. In the class imbalance setting, we make 30\% of the classes imbalanced by reducing the number of data points by 50\% generating Classimbalance versions of the datasets. 

\noindent \textbf{Baselines in each setting.} We compare the variants of our proposed algorithm ({\em i.e.,} \model, \textsc{Grad-MatchPB}, \textsc{Grad-Match-Warm}, \textsc{Grad-MatchPB-Warm}) with variants of \textsc{Craig}~\cite{mirzasoleiman2020coresets} ({\em i.e.,} \textsc{Craig}, \textsc{CraigPB}, \textsc{Craig-Warm}, \textsc{CraigPB-Warm}), and variants of \textsc{Glister}~\cite{killamsetty2021glister} ({\em i.e.,} \textsc{Glister}, \textsc{Glister-Warm}). Additionally, we compare against \textsc{Random} ({\em i.e.,} randomly select points equal to the budget), and \textsc{Full-EarlyStop}, where we do an early stop to full training to match the time taken (or energy used) by the subset selection. %We use a stronger \textsc{Random} baseline in the class imbalance setting, where we randomly sample points based on the (clean) validation set distribution. 

\noindent \textbf{Datasets, model architecture and experimental setup: }  To demonstrate the effectiveness of \model\ and its variants on real-world datasets, we performed experiments on CIFAR100 (60000 instances)~\cite{Krizhevsky09learningmultiple}, MNIST (70000 instances)~\cite{lecun2010mnist}, CIFAR10 (60000 instances)~\cite{Krizhevsky09learningmultiple}, SVHN (99,289 instances)~\cite{Netzer2011}, and
ImageNet-2012 (1.4 Million instances)~\cite{ILSVRC15} datasets. Wherever the datasets do not have a pre-specified validation set, we split the original training set into a new train (90\%) and validation sets (10\%). We ran experiments using an SGD optimizer with an initial learning rate of 0.01, a momentum of 0.9, and a weight decay of 5e-4. We decay the learning rate using cosine annealing~\cite{loshchilov2017sgdr} for each epoch. For MNIST, we use the LeNet model~\cite{lecun1989backpropagation} and train the model for 200 epochs. For all other datasets, we use the ResNet18 model~\cite{he2016deep} and train the model for 300 epochs (except for ImageNet, where we train the model for 350 epochs).  In most of our experiments, we train the data selection algorithms (and full training) using the same number of epochs; the only difference is that each epoch is much smaller with smaller subsets, thereby enabling speedups/energy savings. We consider one additional experiment where we run \textsc{Grad-MatchPB-Warm} for $50$ more epochs to see how quickly it achieves comparable accuracy to full training. All experiments were run on V100 GPUs. Furthermore, the accuracies reported in the results are mean accuracies after five runs, and the standard deviations are given in Appendix~\ref{app:std}. More details are in Appendix~\ref{app-exp}.

\begin{figure*}[!htb]
\centering
\includegraphics[width = 16cm, height=0.8cm]{icml_style/figures/legend_notbold.pdf}
\centering
\hspace{-0.6cm}
\begin{subfigure}[b]{0.21\textwidth}
\centering
\includegraphics[width=3.2cm, height=2.5cm]{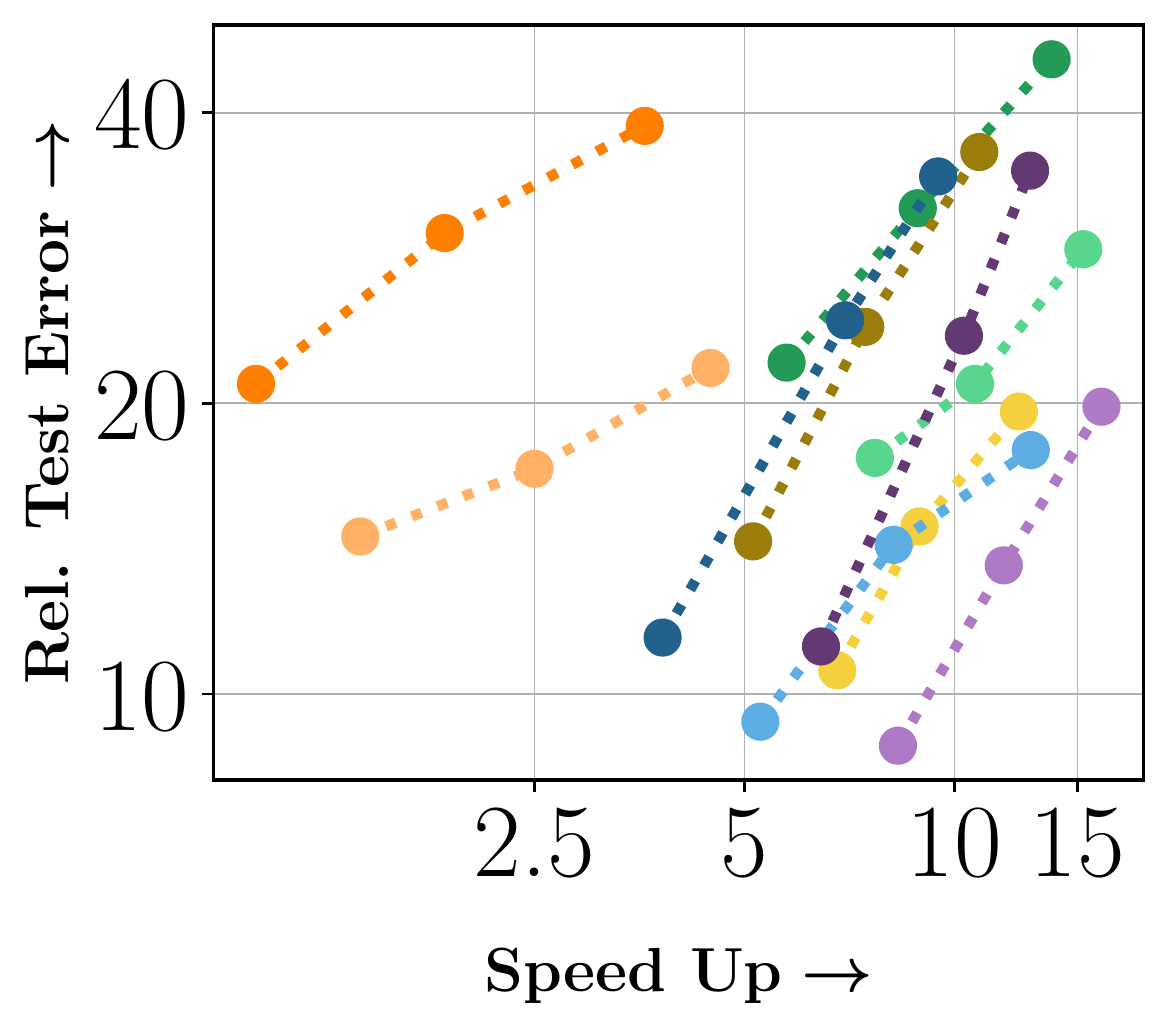}
\caption*{$\underbracket[1pt][1.0mm]{\hspace{3.2cm}}_{\substack{\vspace{-4.0mm}\\
\colorbox{white}{(a) \scriptsize R analysis}}}$}
\phantomcaption
\label{fig:Rplot-cifar100}
\end{subfigure}
\begin{subfigure}[b]{0.21\textwidth}
\centering
\includegraphics[width=3.2cm, height=2.5cm]{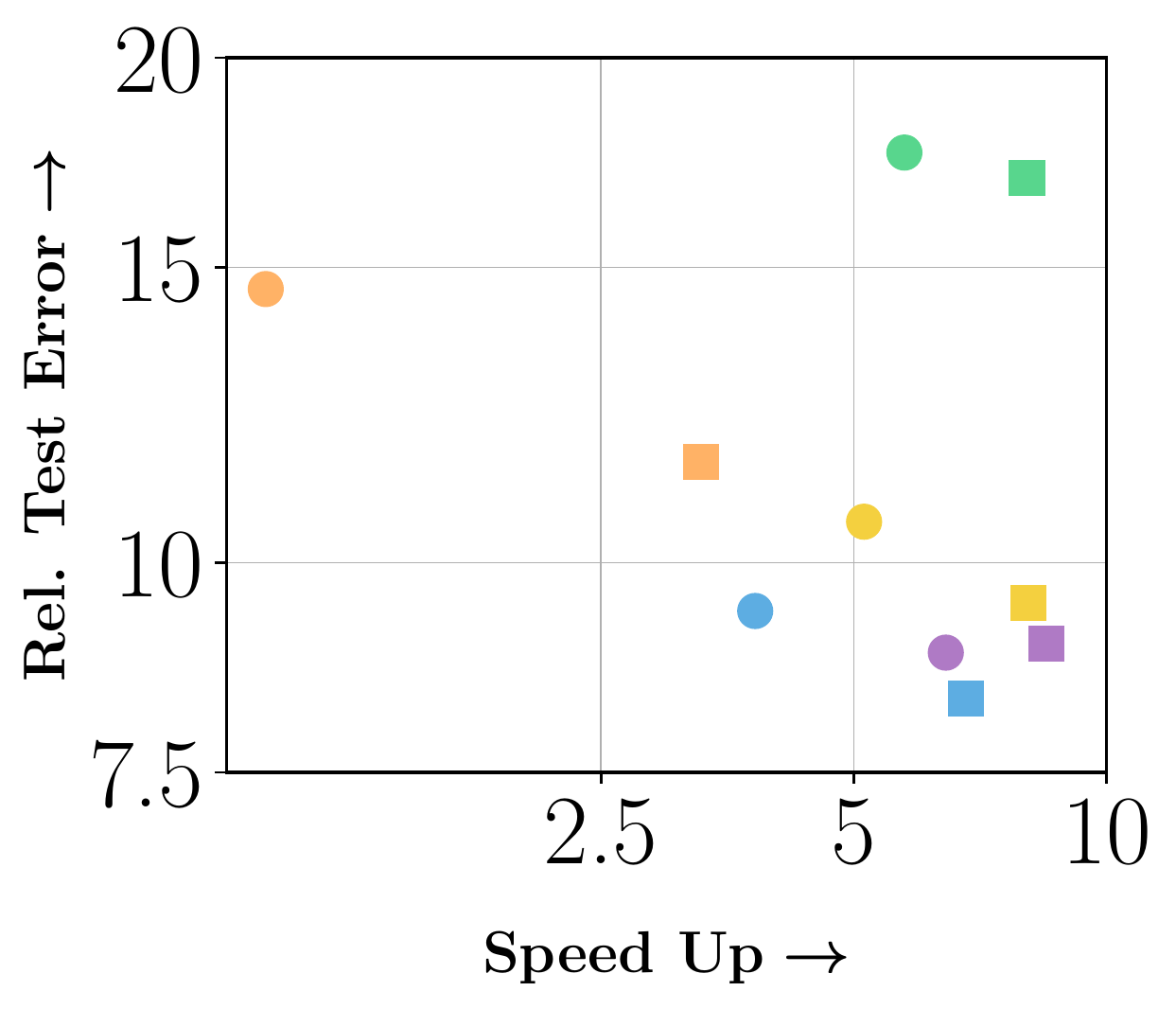}
\caption*{$\underbracket[1pt][1.0mm]{\hspace{3.2cm}}_{\substack{\vspace{-4.0mm}\\
\colorbox{white}{(b) \scriptsize Optimal R}}}$}
\phantomcaption
\label{fig:OptimalRplot}
\end{subfigure}
\begin{subfigure}[b]{0.21\textwidth}
\centering
\includegraphics[width=3.2cm, height=2.5cm]{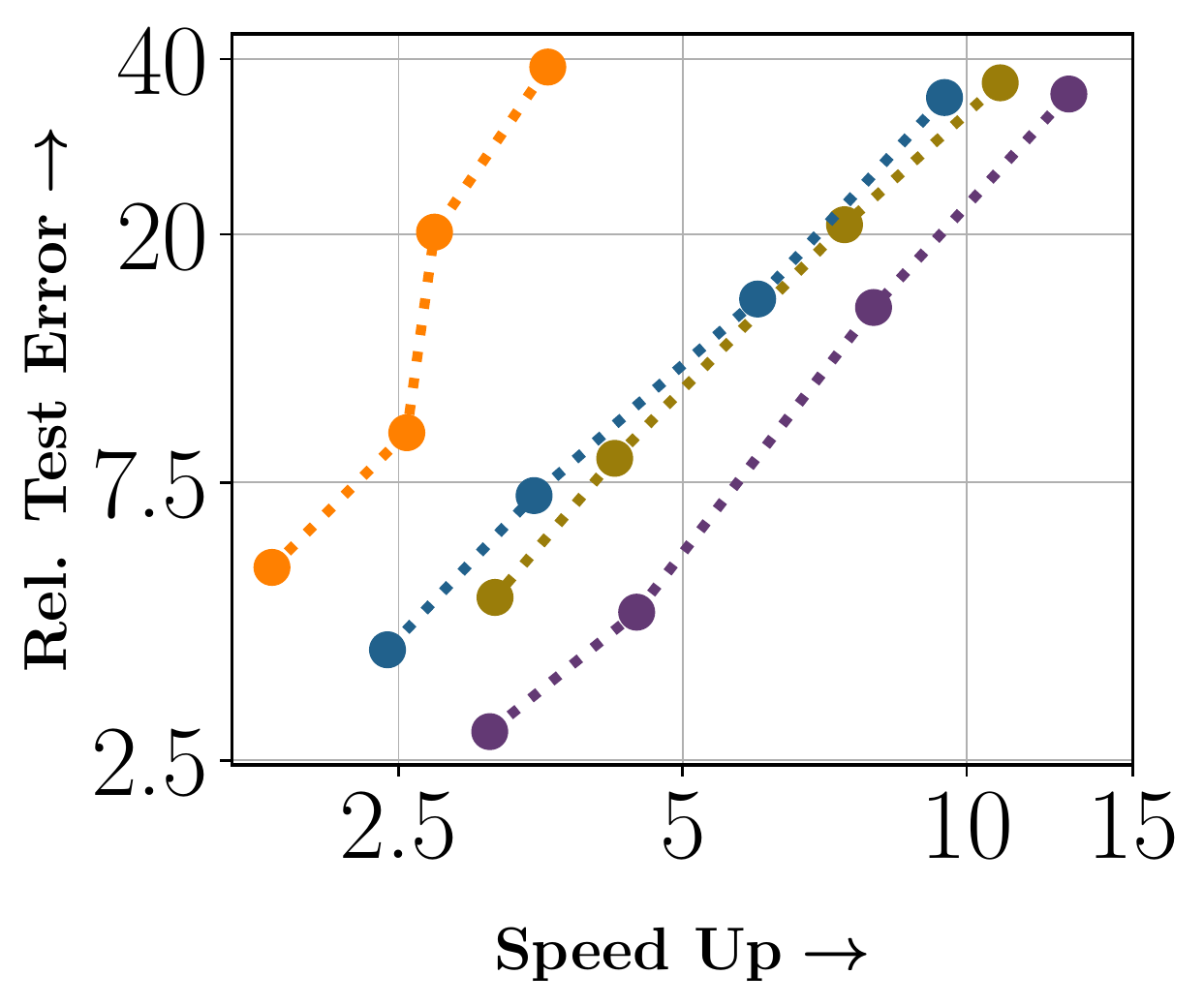}
\caption*{$\underbracket[1pt][1.0mm]{\hspace{3.2cm}}_{\substack{\vspace{-4.0mm}\\
\colorbox{white}{(c) \scriptsize PB vs Non-PB}}}$}
\phantomcaption
\label{fig:PBvsNonPB}
\end{subfigure}
\begin{subfigure}[b]{0.27\textwidth}
\centering
\includegraphics[width=4.5cm, height=2.5cm]{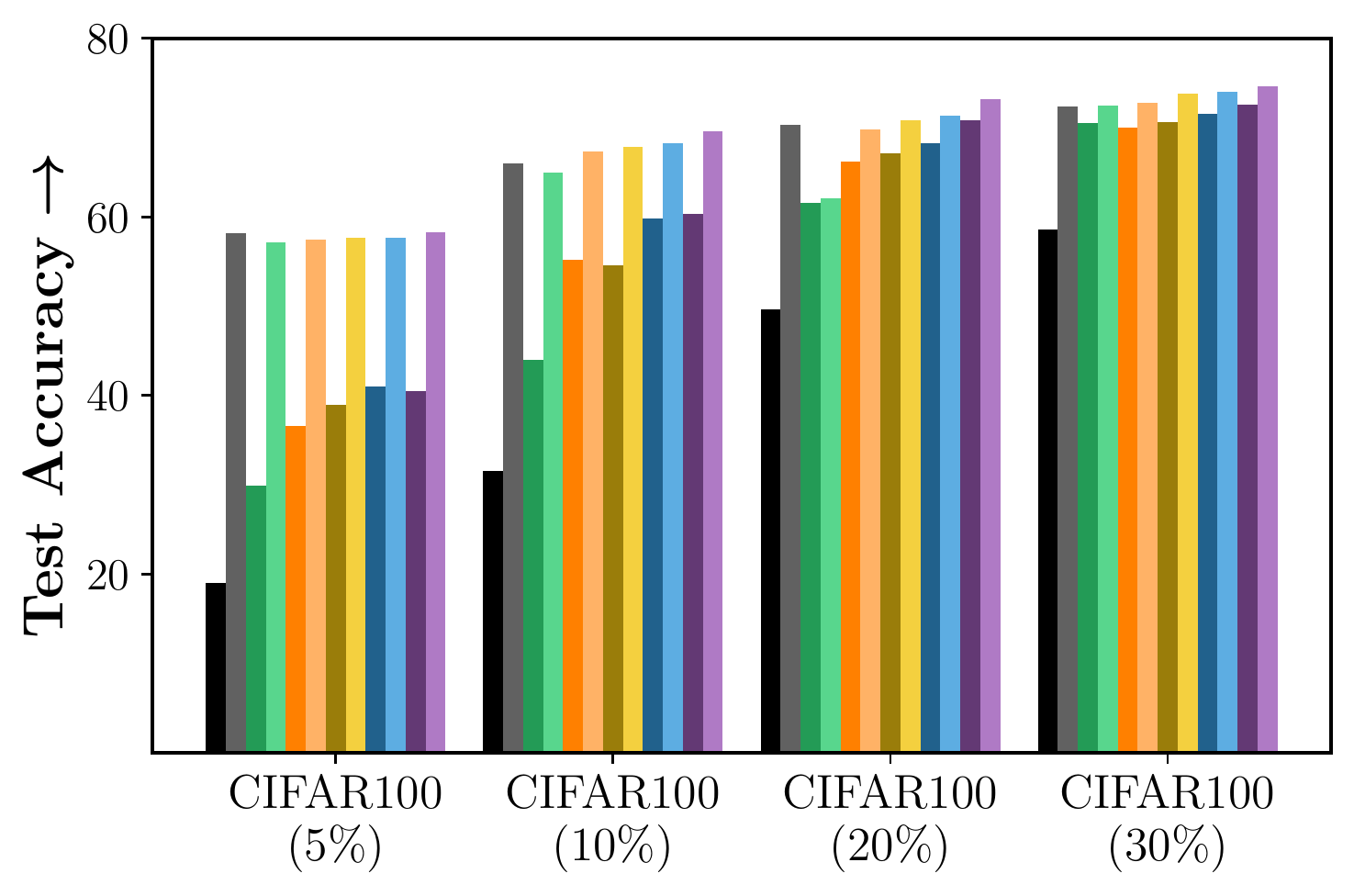}
\caption*{$\underbracket[1pt][1.0mm]{\hspace{4.5cm}}_{\substack{\vspace{-4.0mm}\\
\colorbox{white}{(d) \scriptsize Warm vs Non-Warm}}}$}
\phantomcaption
\label{fig:WarmvsNonWarm}
\end{subfigure}
\hspace{-0.6cm}
\begin{subfigure}[b]{0.22\textwidth}
\centering
\includegraphics[width=3.3cm, height=2.5cm]{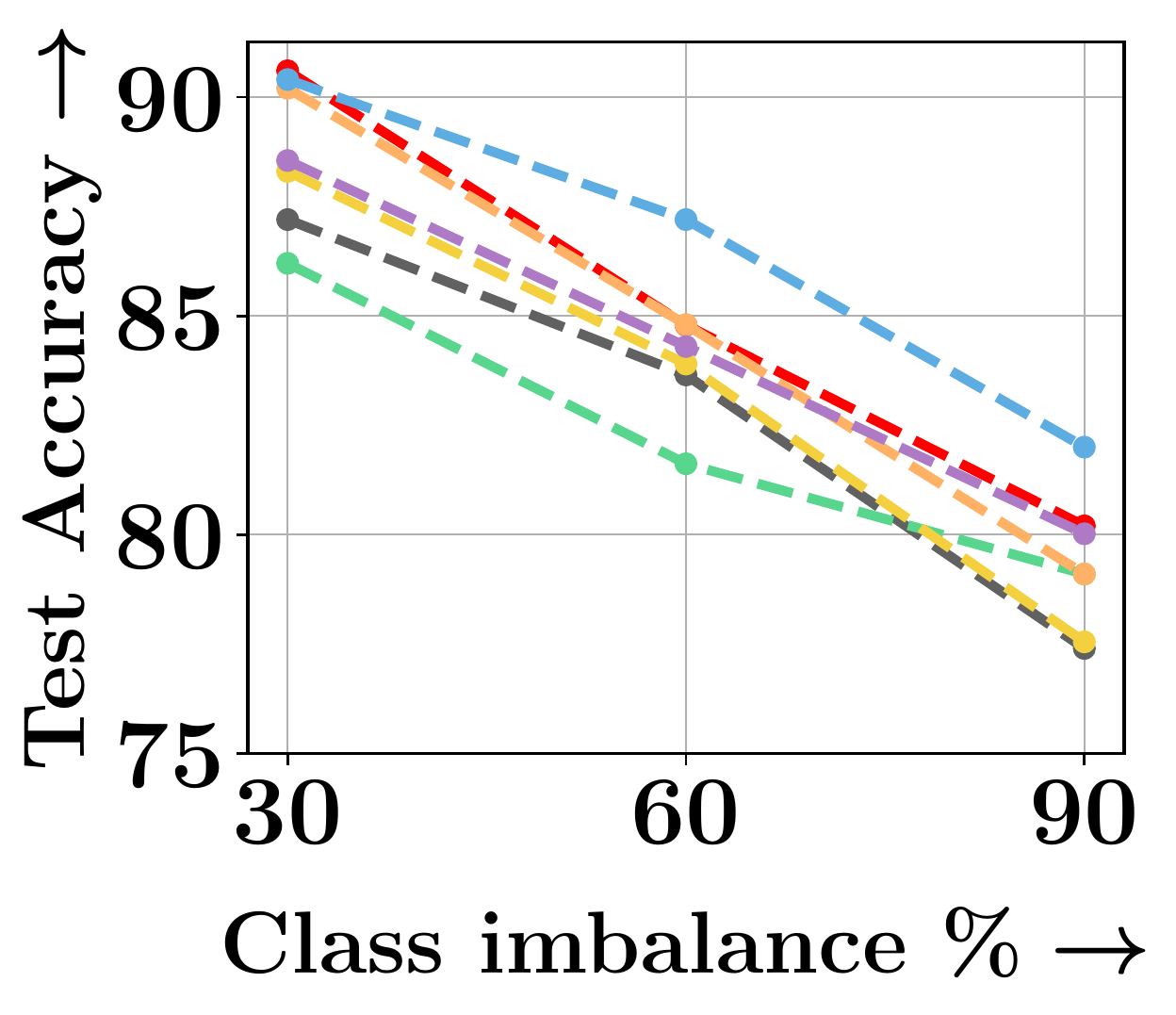}
\caption*{$\underbracket[1pt][1.0mm]{\hspace{3.9cm}}_{\substack{\vspace{-4.0mm}\\
\colorbox{white}{(e) \scriptsize CIFAR10 Varying Imbalance}}}$}
\phantomcaption
\label{fig:classimb}
\end{subfigure}
\begin{subfigure}[b]{0.22\textwidth}
\centering
\includegraphics[width=3.2cm, height=2.5cm]{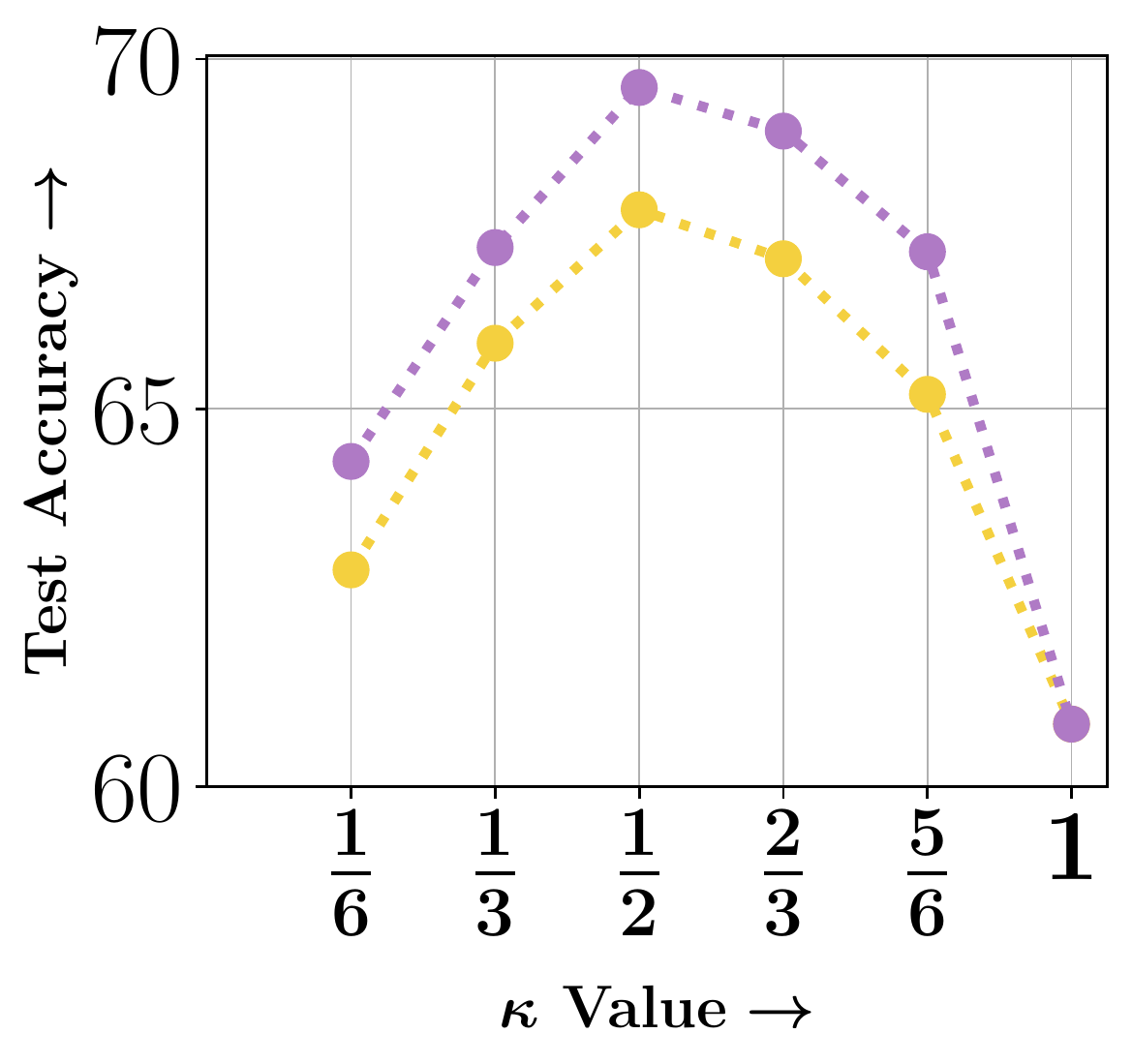}
\caption*{$\underbracket[1pt][1.0mm]{\hspace{3.2cm}}_{\substack{\vspace{-4.0mm}\\
\colorbox{white}{(f) \scriptsize Kappa Analysis}}}$}
\phantomcaption
\label{fig:Kappa}
\end{subfigure}
%\caption{CIFAR10 Class imbalance}
%\caption{CIFAR10 Lambda Accuracy}
\begin{subfigure}[b]{0.4\textwidth}
\centering
\includegraphics[width=3.2cm, height=2.5cm]{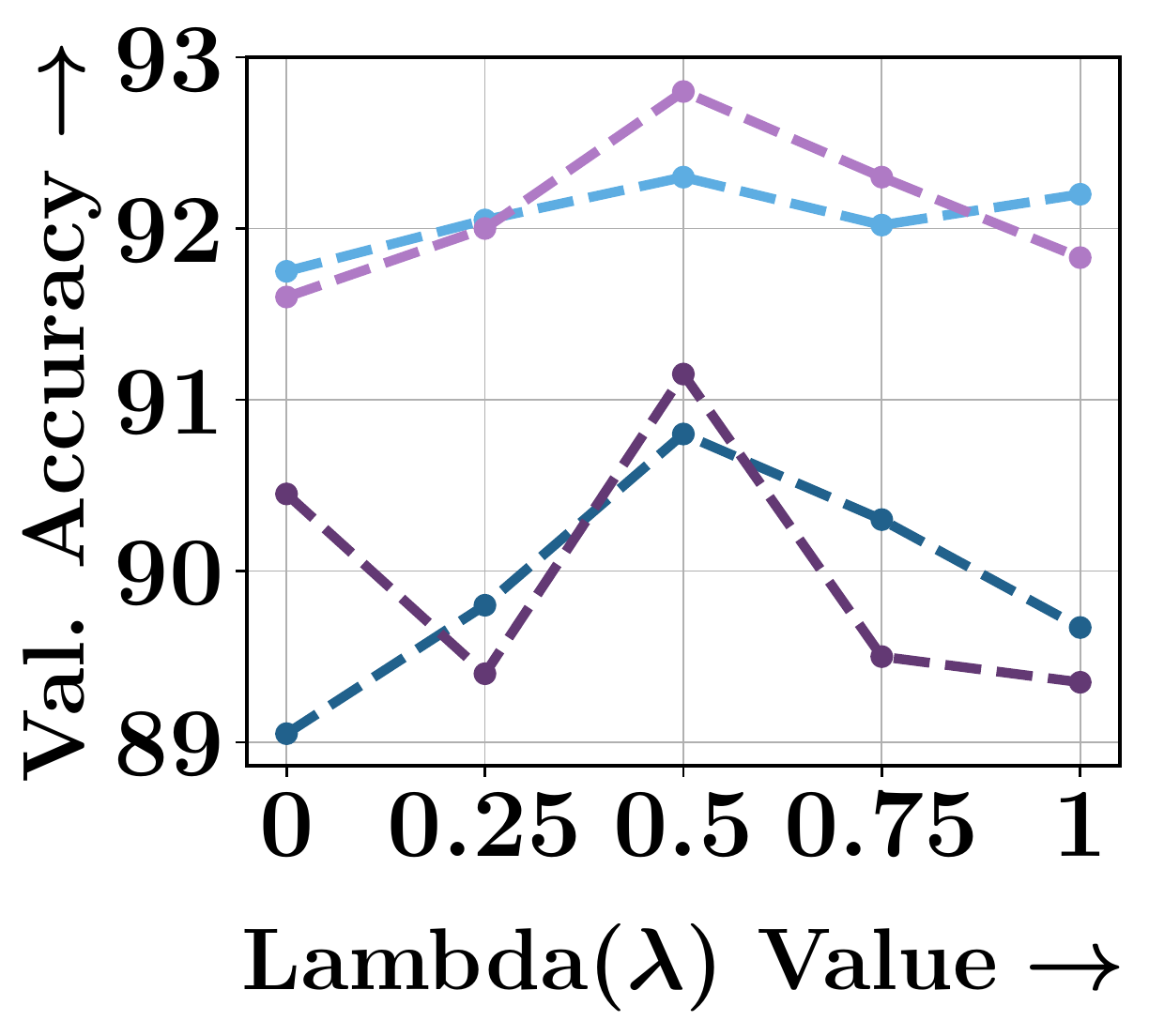} 
\hfill
\includegraphics[width=3.2cm, height=2.5cm]{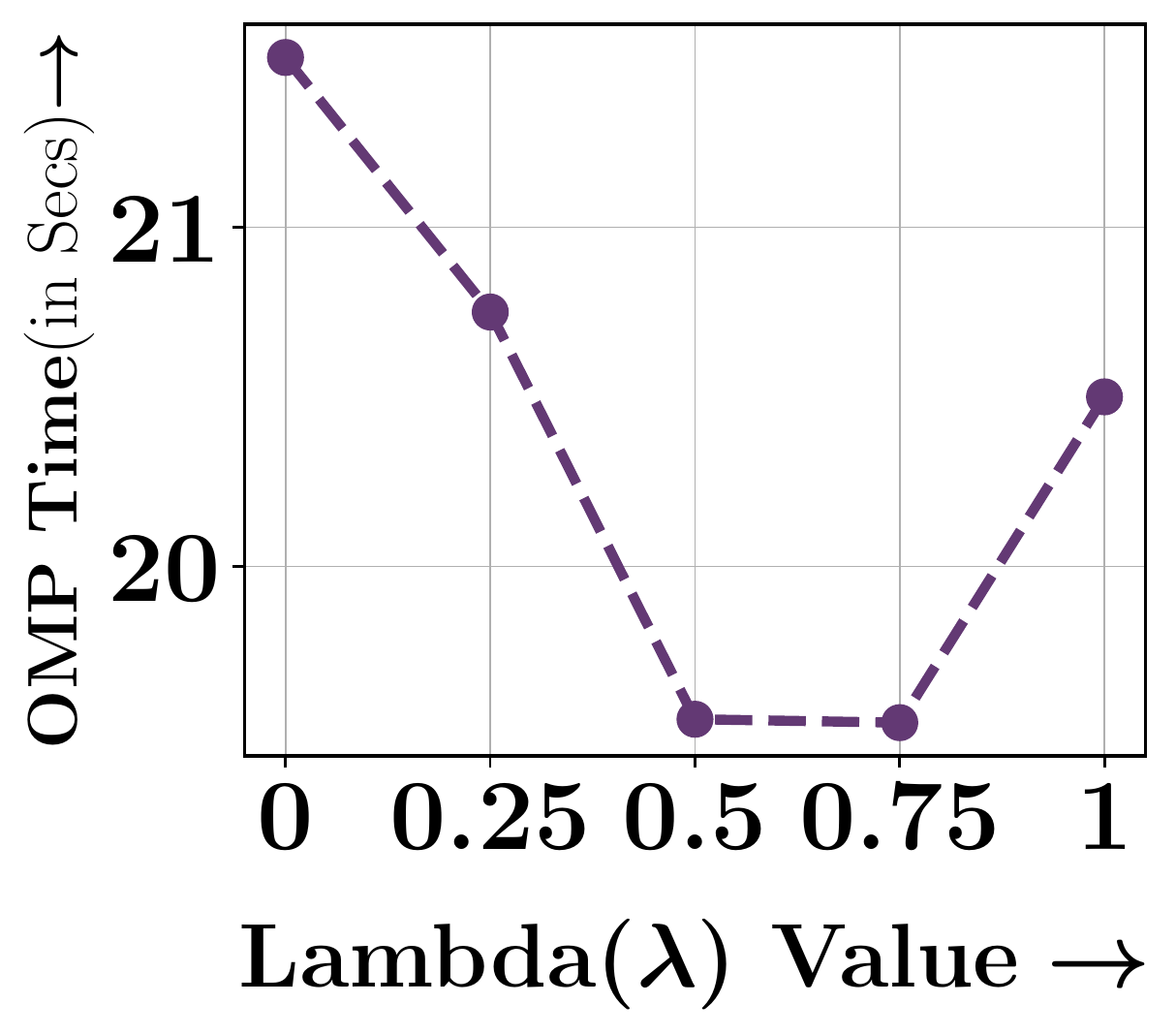} 
\caption*{$\underbracket[1pt][1.0mm]{\hspace{6.6cm}}_{\substack{\vspace{-4.0mm}\\
\colorbox{white}{\scriptsize (g) $\lambda$ Analysis(CIFAR10, 10\%)}}}$}
\phantomcaption
\label{fig:lambda}
\end{subfigure}\quad
%\caption{CIFAR10 Lambda Timing}
%\caption{CIFAR10 Small Models}
\caption{{Sub-figure (a) compares the effect of varying $R$ (5, 10 and 20) for different strategies (5\% CIFAR-100). Sub-figure (b) compares $R = 5$ with a 5\% subset (circles) and $R = 20$ with 10\% subset (squares) showing that the latter is more efficient and accurate. Sub-figure (c) compares the per-batch (PB) versions with non-PB versions showing that the former has a better accuracy-efficiency trade-off. Sub-figure (d) compares the warm-start variants with the variants without warm-start for different subset sizes of CIFAR-100. Sub-figure (e) shows the performance of different selection strategies for 30\% CIFAR-10 with varying percentages of imbalanced classes: 30\%, 60\%, and 90\%. \textsc{Grad-Match-Warm} outperforms all baselines, including full training (which under-performs due to a high imbalance). Sub-figure (f) shows the effect of the warm-start parameter $\kappa$ for CIFAR-100. Sub-figure (g) shows the effect of the regularization parameter $\lambda$ on \model{} and its variants for 10\% CIFAR-10.}}
\label{fig:ablation_studies}
\end{figure*}

\noindent \textbf{Data selection setting: } Since the goal of our experiments is efficiency, we use smaller subset sizes. For MNIST, we use sizes of [1\%, 3\%, 5\%, 10\%], for ImageNet-2012 we use [5\%, 10\%, 30\%], while for the others, we use [5\%, 10\%, 20\%, 30\%]. For the warm versions, we set $\kappa = 1/2$ (i.e. 50\% warm-start and 50\% data selection). Also, we set $R = 20$ in all experiments. In our ablation study experiments, we study the effect of varying $R$ and $\kappa$.

\noindent \textbf{Speedups and energy gains compared to full training: } In Figures~\ref{fig:CIFAR100},\ref{fig:MNIST},\ref{fig:CIFAR10},\ref{fig:SVHN},\ref{fig:imagenet}, we present scatter plots of relative error {\em vs.} speedups, both {\em w.r.t} full training. Figures~\ref{fig:CIFAR100Energy},\ref{fig:CIFAR10Energy} show scatter plots of relative error {\em vs.} energy efficiency, again {\em w.r.t} full training. In each case, we also include the cost of subset selection and subset training while computing the wall-clock time or energy consumed. For calculating the energy consumed by the GPU/CPU cores, we use pyJoules\footnote{\scriptsize{\url{https://pypi.org/project/pyJoules/}}.}. As a first takeaway, we note that \textsc{Grad-Match} and its variants, achieve significant speedup (single GPU) and energy savings when compared to full training. In particular ({\em c.f.}, Figure~\ref{fig:mainresults}) for CIFAR-10, \textsc{Grad-MatchPB-Warm} achieves a 7x, 4.2x and 3x speedup and energy gains (with 10\%, 20\%, 30\% subsets) with an accuracy drop of only 2.8\%, 1.5\% and 0.9\% respectively. For CIFAR-100, \textsc{Grad-MatchPB-Warm} achieves a 4.8x and 3x speedup with an accuracy loss of 2.1\% and 0.7\% respectively, while for ImageNet, (30\% subset), \textsc{Grad-MatchPB-Warm} achieves a 3x speedup with an accuracy loss of 1.3\%. The gains are even more significant for MNIST.

\noindent \textbf{Comparison to other baselines: } \textsc{Grad-MatchPB-Warm} not only outperforms random selection and \textsc{Full-EarlyStop} consistently, but also outperforms variants of \textsc{Craig}, \textsc{CraigPB}, and \textsc{Glister}. Furthermore, \textsc{Glister} and \textsc{Craig} could not run on ImageNet due to large memory requirements and running time. \textsc{Grad-Match}, \textsc{Grad-MatchPB}, and \textsc{CraigPB} were the only variants which could scale to ImageNet. Furthermore, \textsc{Glister} and \textsc{Craig} also perform poorly on CIFAR-100. We see that \textsc{Grad-MatchPB-Warm} almost consistently achieves best speedup-accuracy tradeoff ({\em i.e.}, the bottom right of the plots) on all datasets. We note that the performance gain provided by the variants of \textsc{Grad-Match} compared to other baselines like \textsc{Glister}, \textsc{Craig} and \textsc{Full-EarlyStop} is statistically significant (Wilcoxon signed-rank test~\cite{wilcoxon1992individual} with a $p$ value $= 0.01$). More details on the comparison (along with a detailed table of numbers) are in Appendix~\ref{app-exp}.
%performed a Wilcoxon signed-rank test \cite{wilcoxon1992individual} to determine whether there is a significant difference between \model{} and other baselines, and  %statistically, with the null hypothesis being that there is no difference between \model and other baselines. We observe that the one-tailed hypothesis test is significant at p=0.01, rejecting the null hypothesis. 
%This implies that \model{} significantly outperforms other baselines like \textsc{Glister}, \textsc{Craig}. 
%Note that in the case of ImageNet, the subset selection techniques (including \textsc{Grad-Match} do not perform as well as \textsc{Full-EarlyStop}, but with larger subset sizes, variants of \textsc{Grad-Match} significantly outperform \textsc{Full-EarlyStop}. Also note that we were unable to run \textsc{Glister} and \textsc{Craig} on ImageNet due to large memory requirements and running time. . The energy savings by subset selection are very similar to the speedups, which makes sense since the training is being performed on subsets. Furthermore, this also means that the subset selection itself is fast and energy-efficient, thereby ensuring that \textsc{Grad-Match} and its variants are energy and time-efficient.

\noindent \textbf{Convergence and running time: } %\todo{While fixed number of epochs is fair wrt time (even while using a standard recipes for CIFAR), a criticism associated with fixed epochs is that some models might take more number of epochs for yielding similar accuracies as another model. In which case validation based convergence might be more acceptable. }
Next, we compare the end-to-end training performance through a convergence plot. We plot test-accuracy versus training time in Figure~\ref{fig:CIFAR100Conv}. The plot shows that \model\ and specifically \textsc{Grad-MatchPB-Warm} is more efficient compared to other algorithms (including variants of \textsc{Glister} and \textsc{Craig}), and also converges faster than full training. Figure~\ref{fig:CIFAR100Ext} shows the extended convergence of \textsc{Grad-MatchPB-Warm} on 30\% CIFAR-100 subset, where the \model\ is allowed to train for as few more epochs to achieve comparable accuracy with Full training at the cost of losing some efficiency. The results show that \textsc{Grad-MatchPB-Warm} achieves similar performance to full training while being 2.5x faster, after running for just 30 to 50 additional epochs. Note that the points marked by \textbf{*} in Figure~\ref{fig:CIFAR100Ext} denotes the standard training endpoint (i.e., 300 epochs) used for all experiments using CIFAR-100.

\noindent \textbf{Comparison to smaller models: }
We compare the speedups achieved by \model{} to the speedups achieved using smaller models for training to understand the importance of data subset selection. We perform additional experiments ({\em c.f.}, Figure.~\ref{fig:CIFAR10_MobileNet}) on the CIFAR-10 dataset using MobileNet-V1 and MobileNet-V2 models as two proxies for small models. The results show that \textsc{Grad-MatchPB-Warm} outperforms the smaller models on both test accuracy and speedup (e.g., MobileNetV2 achieves less than 2x speedup with 1\% accuracy drop). 

\noindent \textbf{Data selection with class imbalance: }
We check the robustness of \model\ and its variants for generalization by comparing the test accuracies achieved on a clean test dataset when class imbalance is present in the training dataset. Following~\cite{killamsetty2021glister}, we form a dataset by making 30\% of the classes imbalanced by reducing the number of data points by 90\%. We present  results on CIFAR10 and MNIST in Figures~\ref{fig:CIFAR10Imb},\ref{fig:MNISTImb} respectively. We use the (clean) validation loss for gradient matching in the class imbalance scenario since the training data is biased.
%as we observed that using training loss gives better results when the train/test distribution is the same, whereas the clean validation loss gives better results in cases where training data is imbalanced or noisy.
The results show that \model\ and its variants outperform other baselines in all cases except for the 30\% MNIST case (where \textsc{Glister}, which also uses a clean validation set, performs better). Furthermore, in the case of MNIST with imbalance, \textsc{Grad-Match-warm} even outperforms training on the entire dataset.  Figure~\ref{fig:classimb} shows the performance on 10\% CIFAR-10 with varying percentages of imbalanced classes: 30\% (as shown in Figure~\ref{fig:CIFAR10Imb}), 60\% and 90\%. Grad-Match-Warm outperforms all baselines, including full training (which under-performs due to a high imbalance). The trend is similar when we vary the degree of imbalance as well.

\noindent \textbf{Ablation study results.} %for $R$, per-batch, warm-start, $\lambda$ and $\kappa$:} 
Next, we study the effect of $R$, per-batch gradients, warm-start, $\lambda$, $\kappa$ and other hyper-parameters. We start with the effect of $R$ on the performance of \model\ and its variants. We study the result on CIFAR-100 dataset at 5\% subset for varying values of $R$ (5,10,20) in the Figure~\ref{fig:Rplot-cifar100} (with leftmost point corresponding to R=5 and the rightmost to R=20).The first takeaway is that, as expected, \model\ and its variants outperform other baselines for different R values. Secondly, this also helps us understand the accuracy-efficiency trade-off with different values of $R$. From Figure~\ref{fig:OptimalRplot}, we see that a 10\% subset with $R = 20$ yields accuracy similar to a 5\% subset with $R = 5$. However, across the different algorithms, we observe that $R = 20$ is more efficient (from a time perspective) because of fewer subset selection runs. We then compare the PB variants of \textsc{Craig} and \textsc{Grad-Match} with their non-PB variants ({\em c.f.}, Figure~\ref{fig:PBvsNonPB}). We see that the PB versions are efficient and lie consistently to the bottom right ({\em, i.e.}, lesser relative test accuracy and higher speedups) than non-PB counterparts. One of the main reasons for this is that the subset selection time for the PB variants is almost half that of the non-PB variant ({\em c.f.}, Appendix~\ref{app:dss}), with similar relative errors. Next, we study the effect of warm-start along with data selection. As shown in Figure~\ref{fig:WarmvsNonWarm}, warm-start, in the beginning, helps the model come to a reasonable starting point for data selection, something which just random sets do not offer. We also observe that the effect of warm-start is more significant for smaller subset sizes (compared to larger sizes) in achieving large accuracy gains compared to the non-warm start versions. Figure~\ref{fig:Kappa} shows the effect of varying $\kappa$ ({\em i.e.}, the warm-start fraction) for 10\% of CIFAR-100. We observe that setting $\kappa = \frac{1}{2}$ generally performs the best. Setting a small value of $\kappa$ leads to sub-optimal performance because of not having a good starting point, while with larger values of $\kappa$ we do not have enough data selection and get results closer to the early stopping. The regularization parameter $\lambda$ prevents OMP from over-fitting (e.g., not assigning large weights to individual samples or mini-batches) since the subset selection is performed only every 20 epochs. Hence variants of \textsc{Grad-Match} performs poorly for small lambda values (e.g., $\lambda$=0) as shown in Figure.~\ref{fig:lambda}. Similarly, \textsc{Grad-Match} and its variants perform poorly for large $\lambda$ values as the OMP algorithm performs sub-optimally due to stronger restrictions on the possible sample weights.  In our experiments, we found that $\lambda=0.5$ achieves the accuracy and efficiency ({\em c.f.}, Figure.~\ref{fig:lambda}), and this holds consistently across subset sizes and datasets. Furthermore, we observed that $\epsilon$ does not significantly affect the performance of \textsc{Grad-Match} and its variants as long as $\epsilon$ is small (e.g., $\epsilon \leq 0.01$).

\section{Conclusions}
We introduce a \emph{Gradient Matching} framework \textsc{Grad-Match}, which is inspired by the convergence analysis of adaptive data selection strategies. \textsc{Grad-Match} optimizes an error term, which measures how well the weighted subset matches either the full gradient or the validation set gradients. We study the algorithm's theoretical properties (convergence rates and approximation bounds, connections to weak-submodularity) and finally demonstrate our algorithm's efficacy by demonstrating that it achieves the best speedup-accuracy trade-off and is more energy-efficient through experiments on several datasets.

\section*{Acknowledgements}
We thank anonymous reviewers, Baharan Mirzasoleiman and Nathan Beck for providing constructive feedback. Durga Sivasubramanian is supported by the Prime Minister's Research Fellowship. Ganesh and Abir are also grateful to IBM Research, India (specifically the IBM AI Horizon Networks - IIT Bombay initiative) for their support and sponsorship. Abir also acknowledges the DST Inspire Award and IITB Seed Grant. Rishabh acknowledges UT Dallas startup funding grant.

\bibliographystyle{icml2021}
\bibliography{dss}

\appendix

\onecolumn
\begin{center}
    \Huge{Appendix}
\end{center}
\section{Summary of Notation}\label{app-notation-summary}
\begin{table*}[!h]
\centering
\arrayrulecolor[rgb]{0.192,0.192,0.192}
%\begin{adjustbox}{max width=0.48\textwidth}
\begin{tabular}{|l|l|p{0.5\textwidth}|} 
\toprule
\hline 
\hline 
\multicolumn{1}{|l|}{Topic} & Notation & Explanation \\ \hline 
\toprule
&  $U$ & Set of $N$ instances in training set \\ 
\multicolumn{1}{|c|}{Data (sub)Sets and indices}
& $V$ &  Set of $M$ instances in validation set \\ 
&  $\Xcal^t$ & Subset of instances from $U$ at the $t^{th}$ epoch\\ 
& $W$ & A generic reference to both $U$ and $V$ \\
& $\pi_t^i \in \Xcal$ & Assignment of element $i\in W$ to an element of $\Xcal$ \\
\hline\hline
& $\theta^*$ & Optimal model parameter (vector) \\ 
\multicolumn{1}{|c|}{Parameters}
&  $\theta_t$ & Updated parameter (vector) at the $t^{th}$ epoch \\ 
& $\mathbf{w}^t$ & Vector weights associated with each data point in $\Xcal^t$ (at the $t^{th}$ epoch) \\
\hline\hline
\multicolumn{1}{|c|}{Loss Functions}
& $L_T$ & Training loss which when evaluated on $x_i \in U$ is referred to as $L_T^i$ \\
& $L_V$ & Validation loss which when evaluated on $x_j \in V$ is referred to as $L_V^j$ \\ 
& $L$ & Generic reference to the loss function which when evaluated on $x_i$ is referred to as $L^i$.   \\ 
& $\mbox{Err}({\mathbf w}^t, \Xcal^t, L, L_T, \theta_t)$ & ${\left\Vert \sum_{i \in \Xcal^t} w^t_i \nabla_{\theta}L_T^i(\theta_t) -  \nabla_{\theta}L(\theta_t)\right\Vert}$ \\
& $E(\Xcal)$ & $\min_{\wb} \mbox{Err}(\wb, \Xcal, L, L_T, \theta_t)$ \\
& $F(\Xcal)$ & $L_{\max} - E(\Xcal)$ where $L_{\max}$ is an upperbound on $E(\Xcal)$\\
& $\hat{E}(\Xcal)$ & The upper bound $\min_{\mathbf w} \mbox{Err}( {\mathbf w}, \Xcal, L, L_T, \theta_t) \leq \hat{E}(\Xcal)$ minimized in Section \\
& $\hat{F}(\Xcal)$ & The facility location lower bound function $\sum_{i\in W} \max_{j \in \Xcal} \big(L_{\max} -   \| \nabla_{\theta} L^i(\theta_t) - \nabla_{\theta} L_T^j(\theta_t)\| \big) $ to be maximized\\ 
& $E_{\lambda}(\Xcal,\mathbf{w})$ & Regularized version of $E(\Xcal)$ defined as $\big\| \sum_{i\in \Xcal} {\mathbf{w} \nabla_{\theta} L_T ^i (\theta) -\nabla_{\theta} L(\theta) \big\|^2 + \lambda ||{\mathbf w}|| ^2}$. See Section \\
& $E_{\lambda}(\Xcal)$ & $\min_{{\mathbf w}} E_{\lambda}(\Xcal,\mathbf{w})$ \\
& $F_{\lambda}(\Xcal)$ & $L_{\max} - \min_{\mathbf w} E_{\lambda}(\Xcal,\wb)$ which we prove to be $\gamma$-weakly submodular in Section and subsequently maximize\\
\hline\hline

& $\sigma_T$ & Upperbound on the gradient of $L_T^i$ \\ 
\multicolumn{1}{|c|}{Hyperparameters}
& $\sigma_V$ & Upperbound on the gradient of $L_V^j$ \\
& $k$ & Size of selected subset of points \\ 
& $R$ & The number of training epochs after which data selection is periodically performed \\ 
& $\alpha_t$ & The learning rate schedule at the $t^{th}$ epoch \\
\hline\hline

\bottomrule
\end{tabular}
%\end{adjustbox}
\arrayrulecolor{black}
\caption{Organization of the notations used througout this paper}
\label{tab:main-notations}
\end{table*}

\section{Proofs of the Technical Results}
\subsection{Proof of Theorem~\ref{thm:convergence-result}} 
\label{app-conv-res-proof}
We begin by first stating and then proving Theorem~\ref{thm:convergence-result}.
\begin{theorem-nono}\label{thm:convergence-result-fullgrad}
Any adaptive data selection algorithm (run with full gradient descent), defined via weights $\wb^t$ and subsets $\Xcal^t$ for $t = 1, \cdots, T$, enjoys the following guarantees:\vspace{2mm}\\
\noindent (1). If $L_T$ is Lipschitz continuous with parameter $\sigma_T$, optimal model parameters $\theta^*$, and $\alpha = \frac{D}{\sigma_T \sqrt{T}}$, then $\min_{t = 1:T} L(\theta_t) - L(\theta^*) \leq \frac{D\sigma_T}{\sqrt{T}} + \frac{D}{T}\sum_{t=1}^{T-1} \mbox{Err}(\wb^t, \Xcal^t, L, L_T, \theta_t)$. \vspace{2mm}\\
\noindent (2) If $L_T$ is Lipschitz smooth with parameter $\mathcal{L}_T$, optimal model parameters $\theta^*$, and $L_T^i$ satisfies $0 \leq L_T^i(\theta) \leq \beta_T$, $\forall i$, then setting $\alpha = \frac{1}{\mathcal{L}_T}$, we have $\min_{t = 1:T} L(\theta_t) - L(\theta^*) \leq \frac{D^2 \mathcal{L}_T + 2\beta_T}{2T}  +\frac{D}{T}\sum_{t=1}^{T-1} \mbox{Err}(\wb^t, \Xcal^t, L, L_T, \theta_t)$. \vspace{2mm}\\
 \noindent (3) If $L_T$ is Lipschitz continuous with parameter $\sigma_T$, optimal model parameters $\theta^*$, and $L$ is strongly convex with parameter $\mu$, then setting a learning rate $\alpha_t = \frac{2}{\mu(1+t)}$, we have $\min_{t = 1:T} L(\theta_t) - L(\theta^*) \leq \frac{2{\sigma_T}^2}{\mu (T+1)} + \sum_{t=1}^{t=T} \frac{2Dt}{T(T+1)} \mbox{Err}(\wb^t, \Xcal^t, L, L_T, \theta_t)$.
\end{theorem-nono}
\begin{proof}
Suppose the gradients of validation loss and training loss are sigma bounded by $\sigma_V$ and $\sigma_T$ respectively. Let $\theta_t$ be the model parameters at epoch $t$ and $\theta^*$ be the optimal model parameters.

Let, $L_w(\theta_t) = \underset{i \in \Xcal^t}{\sum} w^t_i L_T^i(\theta_t)$ be the weighted subset training loss parameterized by model parameters $\theta_t$  at time step t. Let $\alpha_t$ be the learning rate used at epoch $t$.

From the definition of Gradient Descent, we have:
\begin{equation}
\begin{aligned}
{\nabla_{\theta}L_w(\theta_t)}^T(\theta_t - \theta^*) = \frac{1}{\alpha_t}{(\theta_t - \theta_{t+1})}^T(\theta_t - \theta^*)
\end{aligned}
\end{equation}
\begin{equation}
\begin{aligned}
{\nabla_{\theta}L_w(\theta_t)}^T(\theta_t - \theta^*) = \frac{1}{2\alpha_t}\left({\left\Vert\theta_t - \theta_{t+1}\right\Vert}^2 + {\left\Vert\theta_{t} - \theta^{*}\right\Vert}^2 - {\left\Vert\theta_{t+1} - \theta^{*}\right\Vert}^2\right)
\end{aligned}
\end{equation}

\begin{equation}
\begin{aligned}
\label{eq-24}
{\nabla_{\theta}L_w(\theta_t)}^T(\theta_t - \theta^*) = \frac{1}{2\alpha_t}\left({\left\Vert \alpha_t \sum_{i \in \Xcal^t} w^t_i \nabla_{\theta}L_T^i(\theta_t)\right\Vert}^2 + {\left\Vert\theta_{t} - \theta^{*}\right\Vert}^2 - {\left\Vert\theta_{t+1} - \theta^{*}\right\Vert}^2 \right)
\end{aligned}
\end{equation}

We can rewrite the function ${\nabla_{\theta}L_w(\theta_t)}^T(\theta_t - \theta^*)$ as follows:
\begin{equation}
\begin{aligned}
\label{eq-25}
{\nabla_{\theta}L_w(\theta_t)}^T(\theta_t - \theta^*) = {\nabla_{\theta}L_w(\theta_t)}^T(\theta_t - \theta^*) -{\nabla_{\theta}L(\theta_t)}^T(\theta_t - \theta^*) + {\nabla_{\theta}L(\theta_t)}^T(\theta_t - \theta^*)
\end{aligned}
\end{equation}

Combining the equations~\eqref{eq-24} and~\eqref{eq-25} we have,
\begin{equation}
\begin{aligned}
{\nabla_{\theta}L_w(\theta_t)}^T(\theta_t - \theta^*) -{\nabla_{\theta}L(\theta_t)}^T(\theta_t - \theta^*) + {\nabla_{\theta}L(\theta_t)}^T(\theta_t - \theta^*) = \frac{1}{2\alpha_t}\left({\left\Vert \alpha_t \sum_{i \in \Xcal^t} w^t_i \nabla_{\theta}L_T^i(\theta_t)\right\Vert}^2 + {\left\Vert\theta_{t} - \theta^{*}\right\Vert}^2 - {\left\Vert\theta_{t+1} - \theta^{*}\right\Vert}^2 \right)
\end{aligned}
\end{equation}

\begin{equation}
\begin{aligned}
\label{gd-cndtn}
{\nabla_{\theta}L(\theta_t)}^T(\theta_t - \theta^*) = \frac{1}{2\alpha_t}\left({\left\Vert \alpha_t \sum_{i \in \Xcal^t} w^t_i \nabla_{\theta}L_T^i(\theta_t)\right\Vert}^2 + {\left\Vert\theta_{t} - \theta^{*}\right\Vert}^2 - {\left\Vert\theta_{t+1} - \theta^{*}\right\Vert}^2 \right) - {\left(\nabla_{\theta}L_w(\theta_t) - \nabla_{\theta}L(\theta_t)\right)}^T(\theta_t - \theta^*) 
\end{aligned}
\end{equation}

Summing up equation~\eqref{gd-cndtn} for different values of $t \in [0, T-1]$ and assuming a constant learning rate of $\alpha_t = \alpha$, we have:
\begin{equation}
\begin{aligned}
\sum_{t=0}^{T-1}{\nabla_{\theta}L(\theta_t)}^T(\theta_t - \theta^*) = & \frac{1}{2\alpha}{\left\Vert\theta_{0} - \theta^{*}\right\Vert}^2 - {\left\Vert\theta_{T} - \theta^{*}\right\Vert}^2 + \sum_{t=0}^{T-1}(\frac{1}{2\alpha}{\left\Vert \alpha \sum_{i \in \Xcal^t} w^t_i \nabla_{\theta}L_T^i(\theta_t)\right\Vert}^2) \nonumber\\ 
&+ \sum_{t=0}^{T-1}\left({\left(\nabla_{\theta}L_w(\theta_t) -  \nabla_{\theta}L(\theta_t)\right)}^T(\theta_t - \theta^*) \right)
\end{aligned}
\end{equation}

Since ${\left\Vert\theta_{T} - \theta^{*}\right\Vert}^2 \geq 0$, we have:
\begin{equation}
\begin{aligned}
\sum_{t=0}^{T-1}{\nabla_{\theta}L(\theta_t)}^T(\theta_t - \theta^*) \leq \frac{1}{2\alpha}{\left\Vert\theta_{0} - \theta^{*}\right\Vert}^2 + \sum_{t=0}^{T-1}(\frac{1}{2\alpha}{\left\Vert \alpha \sum_{i \in \Xcal^t} w^t_i \nabla_{\theta}L_T^i(\theta_t)\right\Vert}^2) + \sum_{t=0}^{T-1}\left({\left(\nabla_{\theta}L_w(\theta_t) -  \nabla_{\theta}L(\theta_t)\right)}^T(\theta_t - \theta^*) \right)
\label{gd-equation}
\end{aligned}
\end{equation}

\begin{case}
\textbf{$L_T$ is lipschitz continuous with parameter $\sigma_T$ and $L$ is a convex function}

From convexity of function $L(\theta)$, we know $L(\theta_t) - L(\theta^*) \leq {\nabla_{\theta}L(\theta_t)}^T(\theta_t - \theta^*)$. Combining this with Equation~\ref{gd-equation} we have,

\begin{equation}
\begin{aligned}
\sum_{t=0}^{T-1} L(\theta_t) - L(\theta^*) \leq \frac{1}{2\alpha}{\left\Vert\theta_{0} - \theta^{*}\right\Vert}^2 + \sum_{t=0}^{T-1}(\frac{1}{2\alpha}{\left\Vert \alpha \sum_{i \in \Xcal^t} w^t_i \nabla_{\theta}L_T^i(\theta_t)\right\Vert}^2) + \sum_{t=0}^{T-1}\left({\left(\nabla_{\theta}L_w(\theta_t) -  \nabla_{\theta}L(\theta_t)\right)}^T(\theta_t - \theta^*) \right)
\end{aligned}
\end{equation}

Since, $\left\Vert L_T(\theta)\right\Vert \leq \sigma_T$, we have ${\left\Vert \alpha \sum_{i \in \Xcal^t} w^t_i \nabla_{\theta}L_T^i(\theta_t)\right\Vert} \leq \sum_{i=1}^{|\Xcal^t|}w_i^t\sigma$. Assuming that the weights at every iteration are normalized such that $\underset{t \in [1, T]}{\forall} \sum_{i=1}^{|\Xcal^t|}w_i^t = 1$ and the training and validation loss gradients are normalized as well, we have ${\left\Vert \alpha \sum_{i \in \Xcal^t} w^t_i \nabla_{\theta}L_T^i(\theta_t)\right\Vert} \leq \sigma_T$. Also assuming that $\left\Vert\theta - \theta^{*}\right\Vert \leq D$, we have,

\begin{equation}
\begin{aligned}
\sum_{t=0}^{T-1} L(\theta_t) - L(\theta^*) \leq \frac{D^2}{2\alpha} + \frac{T \alpha \sigma_T^2}{2} + \sum_{t=0}^{T-1}D\left({\left\Vert\nabla_{\theta}L_w(\theta_t) -  \nabla_{\theta}L(\theta_t)\right\Vert} \right)
\end{aligned}
\end{equation}

\begin{equation}
\begin{aligned}
\label{avg_loss}
\frac{\sum_{t=0}^{T-1}L(\theta_t) - L(\theta^*)}{T} \leq \frac{D^2}{2\alpha T} + \frac{\alpha \sigma_T^2}{2} + \sum_{t=0}^{T-1}\frac{D}{T}\left({\left\Vert\nabla_{\theta}L_w(\theta_t) - \nabla_{\theta}L(\theta_t)\right\Vert} \right)
\end{aligned}
\end{equation}

Since, $\min{(L(\theta_t) - L(\theta^*))} \leq \frac{\sum_{t=0}^{T-1}L(\theta_t) - L(\theta^*)}{T}$, we have:
\begin{equation}
\begin{aligned}
\min{(L(\theta_t) - L(\theta^*))} \leq \frac{D^2}{2\alpha T} + \frac{\alpha \sigma_T^2}{2} + \sum_{t=0}^{T-1}\frac{D}{T}\left({\left\Vert\nabla_{\theta}L_w(\theta_t) -  \nabla_{\theta}L(\theta_t)\right\Vert} \right)
\end{aligned}
\end{equation}

Substituting $L_w(\theta_t) = \underset{i \in \Xcal^t}{\sum} w^t_i L_T^i(\theta_t)$ in the above equation we have,
\begin{equation}
\begin{aligned}
\min{(L(\theta_t) - L(\theta^*))} \leq \frac{D^2}{2\alpha T} + \frac{\alpha \sigma_T^2}{2} + \sum_{t=0}^{T-1}\frac{D}{T}\left({\left\Vert\underset{i \in \Xcal^t}{\sum} w^t_i \nabla_{\theta}L_T^i(\theta_t) -  \nabla_{\theta}L(\theta_t)\right\Vert} \right)
\end{aligned}
\end{equation}

Choosing $\alpha = \frac{D}{\sigma_T \sqrt{T}}$, we have:
\begin{equation}
\begin{aligned}
\label{min_loss}
\min{(L(\theta_t) - L(\theta^*))} \leq \frac{D\sigma_T}{\sqrt{T}} + \sum_{t=0}^{T-1}\frac{D}{T}\left({\left\Vert\underset{i \in \Xcal^t}{\sum} w^t_i \nabla_{\theta}L_T^i(\theta_t) -  \nabla_{\theta}L(\theta_t)\right\Vert} \right)
\end{aligned}
\end{equation}
\end{case}

\begin{case}
\textbf{$L_T$ is lipschitz smooth with parameter $\mathcal{L}_T$, and $\forall i$,  $L_T^i$ satisfies $0 \leq L_T^i(\theta) \leq \beta_T$}

Since $L_w(\theta_t) = \underset{i \in \Xcal^t}{\sum} w^t_i L_T^i(\theta_t)$, from the additive property of lipschitz smooth functions we can say that $L_w(\theta_t)$ is also lipschitz smooth with constant $\underset{i \in \Xcal^t}{\sum} w^t_i \mathcal{L}_T$. Assuming that the weights at every iteration are normalized such that $\underset{t \in [0, T]}{\forall} \sum_{i=1}^{|\Xcal^t|}w_i^t = 1$, we can say that $L_w(\theta_t)$ is lipschitz smooth with constant $\mathcal{L}_T$.

From lipschitz smoothness of function $L_w(\theta)$, we have:
\begin{equation}
\begin{aligned}
L_w(\theta_{t+1}) \leq L_w(\theta_t) + \nabla_{\theta}L_w(\theta_t)^T(\theta_{t+1} - \theta_{t}) + \frac{L}{2}{\left\Vert\theta_{t+1} - \theta_t \right\Vert}^2
\end{aligned}
\end{equation}

Since $\theta_{t+1} - \theta_t = -\alpha \nabla_{\theta}L_w(\theta_t)$, we have:
\begin{equation}
\begin{aligned}
L_w(\theta_{t+1}) \leq L_w(\theta_t) - \alpha \nabla_{\theta}L_w(\theta_t)^T\nabla_{\theta}L_w(\theta_t) + \frac{\mathcal{L}_{T}}{2}{\left\Vert \alpha \nabla_{\theta}L_w(\theta_t) \right\Vert}^2
\end{aligned}
\end{equation}

\begin{equation}
\begin{aligned}
L_w(\theta_{t+1}) \leq L_w(\theta_t) + \frac{\mathcal{L}_T \alpha^2 - 2 \alpha}{2}{\left\Vert \nabla_{\theta}L_w(\theta_t) \right\Vert}^2
\end{aligned}
\end{equation}

Choosing $\alpha = \frac{1}{\mathcal{L}_T}$, we have:
\begin{equation}
\begin{aligned}
L_w(\theta_{t+1}) \leq L_w(\theta_t) - \frac{1}{2\mathcal{L}_T}{\left\Vert \nabla_{\theta}L_w(\theta_t) \right\Vert}^2
\end{aligned}
\end{equation}

Since $\nabla_{\theta}L_w(\theta_T) = \underset{i \in \Xcal^t}{\sum} w^t_i \nabla_{\theta}L_T^i(\theta_t)$, we have:
\begin{equation}
\begin{aligned}
L_w(\theta_{t+1}) \leq L_w(\theta_t) - \frac{1}{2\mathcal{L}_T}{\left\Vert \underset{i \in \Xcal^t}{\sum} w^t_i \nabla_{\theta}L_T^i(\theta_t) \right\Vert}^2
\end{aligned}
\end{equation}

\begin{equation}
\begin{aligned}
\frac{1}{2\mathcal{L}_T}{\left\Vert \underset{i \in \Xcal^t}{\sum} w^t_i \nabla_{\theta}L_T^i(\theta_t) \right\Vert}^2 \leq L_w(\theta_t) - L_w(\theta_{t+1}) 
\end{aligned}
\end{equation}

Summing the above equation for different values of $t$ in $[0, T-1]$, we have:

\begin{equation}
\begin{aligned}
\sum_{t=0}^{t=T-1}\frac{1}{2\mathcal{L}_T}{\left\Vert \underset{i \in \Xcal^t}{\sum} w^t_i \nabla_{\theta}L_T^i(\theta_t) \right\Vert}^2 \leq \sum_{t=0}^{t=T-1} (L_w(\theta_t) - L_w(\theta_{t+1}))
\end{aligned}
\end{equation}

\begin{equation}
\begin{aligned}
\sum_{t=0}^{t=T-1}\frac{1}{2\mathcal{L}_T}{\left\Vert \underset{i \in \Xcal^t}{\sum} w^t_i \nabla_{\theta}L_T^i(\theta_t) \right\Vert}^2 \leq L_w(\theta_0) - L_w(\theta_T)
\label{smooth-gd-cndtn}
\end{aligned}
\end{equation}

Substituting $\alpha=\frac{1}{\mathcal{L}_T}$ in Equation~\ref{gd-equation}, we have:

\begin{equation}
\begin{aligned}
\sum_{t=0}^{T-1}{\nabla_{\theta}L(\theta_t)}^T(\theta_t - \theta^*) \leq \frac{\mathcal{L}_T}{2}{\left\Vert\theta_{0} - \theta^{*}\right\Vert}^2 + \sum_{t=0}^{T-1}(\frac{1}{2\mathcal{L}_T}{\left\Vert \sum_{i \in \Xcal^t} w^t_i \nabla_{\theta}L_T^i(\theta_t)\right\Vert}^2) + \sum_{t=0}^{T-1}\left({\left(\nabla_{\theta}L_w(\theta_t) -  \nabla_{\theta}L(\theta_t)\right)}^T(\theta_t - \theta^*) \right)
\end{aligned}
\end{equation}

Substituting Equation~\ref{smooth-gd-cndtn}, we have:

\begin{equation}
\begin{aligned}
\sum_{t=0}^{T-1}{\nabla_{\theta}L(\theta_t)}^T(\theta_t - \theta^*) \leq \frac{\mathcal{L}_T}{2}{\left\Vert\theta_{0} - \theta^{*}\right\Vert}^2 + L_w(\theta_0) - L_w(\theta_T) + \sum_{t=0}^{T-1}\left({\left(\nabla_{\theta}L_w(\theta_t) -  \nabla_{\theta}L(\theta_t)\right)}^T(\theta_t - \theta^*) \right)
\end{aligned}
\end{equation}

\begin{equation}
\begin{aligned}
\sum_{t=0}^{T-1}{\nabla_{\theta}L(\theta_t)}^T(\theta_t - \theta^*) \leq \frac{\mathcal{L}_T}{2}{\left\Vert\theta_{0} - \theta^{*}\right\Vert}^2 + L_w(\theta_0) + \sum_{t=0}^{T-1}\left({\left(\nabla_{\theta}L_w(\theta_t) -  \nabla_{\theta}L(\theta_t)\right)}^T(\theta_t - \theta^*) \right)
\end{aligned}
\end{equation}

Since $L_T(\theta)$ is bounded by $\beta_T$, we have $L_w(\theta) = \underset{i \in \Xcal^t}{\sum} w^t_i L_T^i(\theta)$ bounded by $\beta_T$ as the weights are normalized to $1$ every epoch (i.e., $\underset{t \in [0, T]}{\forall} \sum_{i=1}^{|\Xcal^t|}w_i^t = 1$).

\begin{equation}
\begin{aligned}
\sum_{t=0}^{T-1}{\nabla_{\theta}L(\theta_t)}^T(\theta_t - \theta^*) \leq \frac{\mathcal{L}_T}{2}{\left\Vert\theta_{0} - \theta^{*}\right\Vert}^2 + \beta_T + \sum_{t=0}^{T-1}\left({\left(\nabla_{\theta}L_w(\theta_t) -  \nabla_{\theta}L(\theta_t)\right)}^T(\theta_t - \theta^*) \right)
\end{aligned}
\end{equation}

Dividing the above equation by T, we have:

\begin{equation}
\begin{aligned}
\frac{\sum_{t=0}^{T-1}{\nabla_{\theta}L(\theta_t)}^T(\theta_t - \theta^*)}{T} \leq \frac{\mathcal{L}_T}{2T}{\left\Vert\theta_{0} - \theta^{*}\right\Vert}^2 + \frac{\beta_T}{T} + \frac{\sum_{t=0}^{T-1}\left({\left(\nabla_{\theta}L_w(\theta_t) -  \nabla_{\theta}L(\theta_t)\right)}^T(\theta_t - \theta^*) \right)}{T}
\end{aligned}
\end{equation}

Since $\left\Vert\theta - \theta^{*}\right\Vert \leq D$, we have,
\begin{equation}
\begin{aligned}
\frac{\sum_{t=0}^{T-1}{\nabla_{\theta}L(\theta_t)}^T(\theta_t - \theta^*)}{T} \leq \frac{D^2 \mathcal{L}_T}{2T} + \frac{\beta_T}{T} + \frac{D}{T}\sum_{t=0}^{T-1}\left({\left \Vert \nabla_{\theta}L_w(\theta_t) -  \nabla_{\theta}L(\theta_t)\right \Vert} \right)
\end{aligned}
\end{equation}

From convexity of function $L(\theta)$, we know $L(\theta_t) - L(\theta^*) \leq {\nabla_{\theta}L(\theta_t)}^T(\theta_t - \theta^*)$. Combining this with above equation we have,
\begin{equation}
\begin{aligned}
\frac{\sum_{t=0}^{T-1}L(\theta_t) - L(\theta^*)}{T} \leq \frac{D^2 \mathcal{L}_T}{2T} + \frac{\beta_T}{T} + \frac{D}{T}\sum_{t=0}^{T-1}\left({\left \Vert \nabla_{\theta}L_w(\theta_t) -  \nabla_{\theta}L(\theta_t)\right \Vert} \right)
\end{aligned}
\end{equation}

Since, $\min{(L(\theta_t) - L(\theta^*))} \leq \frac{\sum_{t=0}^{T-1}L(\theta_t) - L(\theta^*)}{T}$, we have:
\begin{equation}
\begin{aligned}
\min{(L(\theta_t) - L(\theta^*))} \leq \frac{D^2 \mathcal{L}_T}{2T} + \frac{\beta_T}{T} + \frac{D}{T}\sum_{t=0}^{T-1}\left({\left \Vert \nabla_{\theta}L_w(\theta_t) -  \nabla_{\theta}L(\theta_t)\right \Vert} \right)
\end{aligned}
\end{equation}

Substituting $L_w(\theta_t) = \underset{i \in \Xcal^t}{\sum} w^t_i L_T^i(\theta_t)$ in the above equation we have,
\begin{equation}
\begin{aligned}
\min{(L(\theta_t) - L(\theta^*))} \leq \frac{D^2 \mathcal{L}_T}{2T} + \frac{\beta_T}{T} + \frac{D}{T}\sum_{t=0}^{T-1}\left({\left \Vert \underset{i \in \Xcal^t}{\sum} w^t_i \nabla_{\theta}L_T^i(\theta_t) -  \nabla_{\theta}L(\theta_t)\right \Vert} \right)
\end{aligned}
\end{equation}
\end{case}

\begin{case}
\textbf{$L_T$ is Lipschitz continuous (parameter $\sigma_T$) and $L$ is strongly convex with parameter $\mu$}

Let the learning at time step $t$ be $\alpha_t$

From Equation~\ref{gd-cndtn}, we have:
\begin{equation}
\begin{aligned}
{\nabla_{\theta}L(\theta_t)}^T(\theta_t - \theta^*) = \frac{1}{2\alpha_t}\left({\left\Vert \alpha_t \sum_{i \in \Xcal^t} w^t_i \nabla_{\theta}L_T^i(\theta_t)\right\Vert}^2 + {\left\Vert\theta_{t} - \theta^{*}\right\Vert}^2 - {\left\Vert\theta_{t+1} - \theta^{*}\right\Vert}^2 \right) - {\left(\nabla_{\theta}L_w(\theta_t) - \nabla_{\theta}L(\theta_t)\right)}^T(\theta_t - \theta^*) 
\end{aligned}
\end{equation}

From the strong convexity of loss function $L$, we have:
\begin{equation}
\begin{aligned}
\nabla_{\theta}L(\theta_t)^T(\theta_t - \theta^*) \geq L(\theta_t) - L(\theta^*) + \frac{\mu}{2}{\left\Vert\theta_t - \theta^*\right\Vert}^2
\end{aligned}
\label{strong-convexity}
\end{equation}

Combining the above two equations, we have:
\begin{equation}
\begin{aligned}
L(\theta_t) - L(\theta^*) = \frac{1}{2\alpha_t}\left({\left\Vert \alpha_t \sum_{i \in \Xcal^t} w^t_i \nabla_{\theta}L_T^i(\theta_t)\right\Vert}^2 + {\left\Vert\theta_{t} - \theta^{*}\right\Vert}^2 - {\left\Vert\theta_{t+1} - \theta^{*}\right\Vert}^2 \right) - {\left(\nabla_{\theta}L_w(\theta_t) - \nabla_{\theta}L(\theta_t)\right)}^T(\theta_t - \theta^*) - \frac{\mu}{2}{\left\Vert\theta_t - \theta^*\right\Vert}^2
\end{aligned}
\end{equation}

\begin{equation}
\begin{aligned}
L(\theta_t) - L(\theta^*) = \frac{\alpha_t}{2}{\left\Vert \sum_{i \in \Xcal^t} w^t_i \nabla_{\theta}L_T^i(\theta_t)\right\Vert}^2 + \frac{\alpha_t^{-1}-\mu}{2}{\left\Vert\theta_{t} - \theta^{*}\right\Vert}^2 - \frac{\alpha_t^{-1}}{2}{\left\Vert\theta_{t+1} - \theta^{*}\right\Vert}^2  - {\left(\nabla_{\theta}L_w(\theta_t) - \nabla_{\theta}L(\theta_t)\right)}^T(\theta_t - \theta^*)
\end{aligned}
\end{equation}

Setting an learning rate of $\alpha_t = \frac{2}{\mu(t+1)}$ and multiplying by $t$ on both sides, we have:
\begin{equation}
\begin{aligned}
t(L(\theta_t) - L(\theta^*)) &= \frac{t}{\mu(t+1)}{\left\Vert \sum_{i \in \Xcal^t} w^t_i \nabla_{\theta}L_T^i(\theta_t)\right\Vert}^2 + \frac{\mu t(t-1)}{4}{\left\Vert\theta_{t} - \theta^{*}\right\Vert}^2\\
&- \frac{\mu t(t+1)}{4}{\left\Vert\theta_{t+1} - \theta^{*}\right\Vert}^2  - t {\left(\nabla_{\theta}L_w(\theta_t) - \nabla_{\theta}L(\theta_t)\right)}^T(\theta_t - \theta^*)
\end{aligned}
\end{equation}

Since, $\left\Vert L_T(\theta)\right\Vert \leq \sigma_T$, we have ${\left\Vert \alpha \sum_{i \in \Xcal^t} w^t_i \nabla_{\theta}L_T^i(\theta_t)\right\Vert} \leq \sum_{i=1}^{|\Xcal^t|}w_i^t \sigma_T$. Assuming that the weights at every iteration are normalized such that $\underset{t \in [1, T]}{\forall} \sum_{i=1}^{|\Xcal^t|}w_i^t = 1$ and the training and validation loss gradients are normalized as well, we have ${\left\Vert \alpha \sum_{i \in \Xcal^t} w^t_i \nabla_{\theta}L_T^i(\theta_t)\right\Vert} \leq \sigma_T$. Also assuming that $\left\Vert\theta - \theta^{*}\right\Vert \leq D$, we have,

\begin{equation}
\begin{aligned}
t(L(\theta_t) - L(\theta^*)) &= \frac{{\sigma_T}^2 t}{\mu(t+1)} + \frac{\mu t(t-1)}{4}{\left\Vert\theta_{t} - \theta^{*}\right\Vert}^2\\
&- \frac{\mu t(t+1)}{4}{\left\Vert\theta_{t+1} - \theta^{*}\right\Vert}^2  + Dt \left \Vert \nabla_{\theta}L_w(\theta_t) - \nabla_{\theta}L(\theta_t)\right\Vert
\end{aligned}
\end{equation}

Summing the above equation from $t=1,\cdots,T$, we have:
\begin{equation}
\begin{aligned}
\sum_{t=1}^{t=T}t(L(\theta_t) - L(\theta^*)) &= \sum_{t=1}^{t=T}\frac{{\sigma_T}^2 t}{\mu(t+1)} + \sum_{t=1}^{t=T}\frac{\mu t(t-1)}{4}{\left\Vert\theta_{t} - \theta^{*}\right\Vert}^2\\
&- \sum_{t=1}^{t=T} \frac{\mu t(t+1)}{4}{\left\Vert\theta_{t+1} - \theta^{*}\right\Vert}^2  + \sum_{t=1}^{t=T} Dt \left \Vert \nabla_{\theta}L_w(\theta_t) - \nabla_{\theta}L(\theta_t)\right\Vert
\end{aligned}
\end{equation}

\begin{equation}
\begin{aligned}
\sum_{t=1}^{t=T}t(L(\theta_t) - L(\theta^*)) &\leq \sum_{t=1}^{t=T}\frac{{\sigma_T}^2}{\mu} + \sum_{t=1}^{t=T}\frac{\mu t(t-1)}{4}{\left\Vert\theta_{t} - \theta^{*}\right\Vert}^2\\
&- \sum_{t=1}^{t=T} \frac{\mu t(t+1)}{4}{\left\Vert\theta_{t+1} - \theta^{*}\right\Vert}^2  + \sum_{t=1}^{t=T} Dt \left \Vert \nabla_{\theta}L_w(\theta_t) - \nabla_{\theta}L(\theta_t)\right\Vert
\end{aligned}
\end{equation}

\begin{equation}
\begin{aligned}
\sum_{t=1}^{t=T}t(L(\theta_t) - L(\theta^*)) &\leq \frac{{\sigma_T}^2 T}{\mu } + \frac{\mu}{4}(0 - T(T+1){\left\Vert\theta_{T+1} - \theta^{*}\right\Vert}^2)\\
&+ \sum_{t=1}^{t=T} Dt \left \Vert \nabla_{\theta}L_w(\theta_t) - \nabla_{\theta}L(\theta_t)\right\Vert
\end{aligned}
\end{equation}

Since $\frac{\mu}{4}(0 - T(T+1){\left\Vert\theta_{T+1} - \theta^{*}\right\Vert}^2) \leq 0$, we have:

\begin{equation}
\begin{aligned}
\sum_{t=1}^{t=T}t(L(\theta_t) - L(\theta^*)) &\leq \frac{{\sigma_T}^2 T}{\mu } + \sum_{t=1}^{t=T} Dt \left \Vert \nabla_{\theta}L_w(\theta_t) - \nabla_{\theta}L(\theta_t)\right\Vert
\end{aligned}
\end{equation}

Since, $L(\theta_t) - L(\theta^*) \leq \min_{t = 1:T} L(\theta_t) - L(\theta^*)$ and multiplying the above equation by $\frac{2}{T(T+1)}$, we have: 

\begin{equation}
\begin{aligned}
\frac{2}{T(T+1)}\sum_{t=1}^{t=T}t(\min_{t = 1:T} L(\theta_t) - L(\theta^*)) \leq \frac{2}{T(T+1)} \frac{{\sigma_T}^2 T}{\mu } + \frac{2}{T(T+1)} \sum_{t=1}^{t=T} Dt \left \Vert \nabla_{\theta}L_w(\theta_t) - \nabla_{\theta}L(\theta_t)\right\Vert
\end{aligned}
\end{equation}

This in turn implies:
\begin{equation}
\begin{aligned}
\min_{t = 1:T} L(\theta_t) - L(\theta^*) &\leq \frac{{\sigma_T}^2 2}{\mu (T+1)} + \sum_{t=1}^{t=T} \frac{2Dt}{T(T+1)} \left \Vert \nabla_{\theta}L_w(\theta_t) - \nabla_{\theta}L(\theta_t)\right\Vert
\end{aligned}
\end{equation}
\end{case}
\end{proof}

\subsection{Convergence Analysis with Stochastic Gradient Descent}
\label{app-conv-res-sgd}
\begin{theorem-nono}\label{thm:sgd-convergence-result}
Denote $L_V$ as the validation loss, $L_T$ as the full training loss, and that the parameters satisfy $||\theta||^2 \leq D^2$. Let $L$ denote either the training or validation loss (with gradient bounded by $\sigma$). Any adaptive data selection algorithm, defined via weights $\wb^t$ and subsets $\Xcal^t$ for $t = 1, \cdots, T$, and run with a learning rate $\alpha$ using stochastic gradient descent enjoys the following convergence bounds:
\begin{itemize}
    \item if $L_T$ is lipschitz continuous and $\alpha = \frac{D}{\sigma_T \sqrt{T}}$, then $\mathop{\mathbb{E}}\left(\min_{t = 1:T} L(\theta_t)\right) - L(\theta^*) \leq \frac{D\sigma_T}{\sqrt{T}} + \frac{D}{T}\sum_{t=1}^{T-1} \mathop{\mathbb{E}}(\mbox{Err}(\wb^t, \Xcal^t, L, L_T, \theta_t))$
    \item if $L_T$ is Lipschitz continuous, $L$ is strongly convex with a strong convexity parameter $\mu$, then setting a learning rate $\alpha_t = \frac{2}{\mu(1+t)}$, then $\mathop{\mathbb{E}}\left(\min_{t = 1:T} L(\theta_t)\right) - L(\theta^*) \leq \frac{{\sigma_T}^2 2}{\mu (T+1)} + \sum_{t=1}^{t=T} \frac{2D}{T(T+1)}\mathop{\mathbb{E}} \left(t\mbox{Err}(\wb^t, \Xcal^t, L, L_T, \theta_t)\right)$
\end{itemize}
\small{
where: \\
$\mbox{Err}(\wb^t, \Xcal^t, L, L_T, \theta_t) = {\left\Vert \sum_{i \in \Xcal^t} w^t_i \nabla_{\theta}L_T^i(\theta_t) -  \nabla_{\theta}L(\theta_t)\right\Vert}$}
\end{theorem-nono}
\normalsize
\begin{proof}
Suppose the gradients of validation loss and training loss are sigma bounded by $\sigma_V$ and $\sigma_T$ respectively. Let $\theta_t$ be the model parameters at epoch $t$ and $\theta^*$ be the optimal model parameters. Let $\alpha_t$ is the learning rate at epoch $t$.

Let $L_w(\theta_t) = \underset{i \in \Xcal^t}{\sum} w^t_i L_T^i(\theta_t)$ be the weighted training loss where $L_T^i(\theta_t)$ is the training loss of the $i^{th}$ instance in the subset $\Xcal^t$. 

Let $\nabla_{\theta}L_w^i(\theta_t)$ be the weighted training loss gradient of the $i^{th}$ instance in the subset $\Xcal^t$, we have:

$$\nabla_{\theta}L_w^i(\theta_t) = w^t_i \nabla_{\theta}L_T^i(\theta_t)$$

For a particular $\theta_t$, conditional expectation of $\nabla_{\theta}L_w^i(\theta_t)$ given $\theta=\theta_t$ over the random choice of $i$ (i.e., randomly selecting $i^{th}$ sample from the subset $\Xcal^t$) yields:
\begin{equation}
    \begin{aligned}
        \mathop{\mathbb{E}}(\nabla_{\theta}L_w^i(\theta_t) \mid \theta = \theta_t) &= \underset{i \in \Xcal^t}{\sum}\nabla_{\theta}w^t_i L_T^i(\theta_t)\\
        &= \nabla_{\theta}L_w(\theta_T)
    \end{aligned}
\end{equation}

In the above equation, $\wb^t$ can be assumed as weighted probability distribution as $\underset{i \in \Xcal^t}{\sum}w_i^t = 1$.

Similarly conditional expectation of $\nabla_{\theta}L_w^i(\theta_t)^T(\theta_t - \theta^*)$ given $\theta=\theta_t$ is,
\begin{equation}
    \begin{aligned}
        \mathop{\mathbb{E}}(\nabla_{\theta}L_w^i(\theta_t)^T(\theta_t - \theta^*) \mid \theta = \theta_t) 
        = \nabla_{\theta}L_w(\theta_T)^T(\theta_t- \theta^*)
    \end{aligned}
\end{equation}

Using the fact that $\theta = \theta_t$ can occur for $\theta$ in some finite set $\Theta$ (i.e., one element for every choice of samples through out all iterations), we have:
\begin{equation}
    \begin{aligned}
    \label{sgd-expectation}
        \mathop{\mathbb{E}}(\nabla_{\theta}L_w^i(\theta_t)^T(\theta_t - \theta^*)) 
        &= \sum_{\theta_t \in \Theta} \mathop{\mathbb{E}}\left(\nabla_{\theta}L_w^i(\theta_t)^T(\theta_t - \theta^*)\right)\operatorname{prob}(\theta = \theta_t)\\
        &= \sum_{\theta_t \in \Theta} \nabla_{\theta}L_w(\theta_T)^T(\theta_t- \theta^*) \operatorname{prob}(\theta = \theta_t)\\
        &= \mathop{\mathbb{E}}\left(\nabla_{\theta}L_w(\theta_T)^T(\theta_t- \theta^*)\right)
    \end{aligned}
\end{equation}

From the definition of stochastic gradient descent, we have:
\begin{equation}
\begin{aligned}
{\nabla_{\theta}L_w^i(\theta_t)}^T(\theta_t - \theta^*) = \frac{1}{\alpha_t}{(\theta_t - \theta_{t+1})}^T(\theta_t - \theta^*)
\end{aligned}
\end{equation}
\begin{equation}
\begin{aligned}
{\nabla_{\theta}L_w^i(\theta_t)}^T(\theta_t - \theta^*) = \frac{1}{2\alpha_t}\left({\left\Vert\theta_t - \theta_{t+1}\right\Vert}^2 + {\left\Vert\theta_{t} - \theta^{*}\right\Vert}^2 - {\left\Vert\theta_{t+1} - \theta^{*}\right\Vert}^2\right)
\end{aligned}
\end{equation}

\begin{equation}
\begin{aligned}
\label{sgd-eq1}
{\nabla_{\theta}L_w^i(\theta_t)}^T(\theta_t - \theta^*) = \frac{1}{2\alpha_t}\left({\left\Vert \alpha_t \nabla_{\theta}L_w^i(\theta_t)\right\Vert}^2 + {\left\Vert\theta_{t} - \theta^{*}\right\Vert}^2 - {\left\Vert\theta_{t+1} - \theta^{*}\right\Vert}^2 \right)
\end{aligned}
\end{equation}

We can rewrite the function ${\nabla_{\theta}L_w^i(\theta_t)}^T(\theta_t - \theta^*)$ as follows:
\begin{equation}
\begin{aligned}
\label{sgd-eq2}
{\nabla_{\theta}L_w^i(\theta_t)}^T(\theta_t - \theta^*) = {\nabla_{\theta}L_w^i(\theta_t)}^T(\theta_t - \theta^*) -{\nabla_{\theta}L(\theta_t)}^T(\theta_t - \theta^*) + {\nabla_{\theta}L(\theta_t)}^T(\theta_t - \theta^*)
\end{aligned}
\end{equation}

Combining the equations Equation~\ref{sgd-eq1} ,Equation~\ref{sgd-eq2} we have,
\begin{equation}
\begin{aligned}
{\nabla_{\theta}L_w^i(\theta_t)}^T(\theta_t - \theta^*) -{\nabla_{\theta}L(\theta_t)}^T(\theta_t - \theta^*) + {\nabla_{\theta}L(\theta_t)}^T(\theta_t - \theta^*) = \frac{1}{2\alpha_t}\left({\left\Vert \alpha_t  \nabla_{\theta}L_w^i(\theta_t)\right\Vert}^2 + {\left\Vert\theta_{t} - \theta^{*}\right\Vert}^2 - {\left\Vert\theta_{t+1} - \theta^{*}\right\Vert}^2 \right)
\end{aligned}
\end{equation}

\begin{equation}
\begin{aligned}
 {\nabla_{\theta}L(\theta_t)}^T(\theta_t - \theta^*) = \frac{1}{2\alpha_t}\left({\left\Vert \alpha_t  \nabla_{\theta}L_w^i(\theta_t)\right\Vert}^2 + {\left\Vert\theta_{t} - \theta^{*}\right\Vert}^2 - {\left\Vert\theta_{t+1} - \theta^{*}\right\Vert}^2 \right) - {(\nabla_{\theta}L_w^i(\theta_t) - \nabla_{\theta}L(\theta_t))}^T(\theta_t - \theta^*) 
\end{aligned}
\end{equation}

Taking expectation on both sides of the above equation, we have:

\begin{equation}
\begin{aligned}
 \mathop{\mathbb{E}}({\nabla_{\theta}L(\theta_t)}^T(\theta_t - \theta^*)) &= \frac{1}{2\alpha_t}\mathop{\mathbb{E}}\left({\left\Vert \alpha_t  \nabla_{\theta}L_w^i(\theta_t)\right\Vert}^2 \right) + \frac{1}{2\alpha_t} \mathop{\mathbb{E}}\left({\left\Vert\theta_{t}
 - \theta^{*}\right\Vert}^2\right)\\
 &- \frac{1}{2\alpha_t} \mathop{\mathbb{E}}\left({\left\Vert\theta_{t+1} - \theta^{*}\right\Vert}^2\right) - \mathop{\mathbb{E}}\left({(\nabla_{\theta}L_w^i(\theta_t) - \nabla_{\theta}L(\theta_t))}^T(\theta_t - \theta^*)\right) 
\end{aligned}
\end{equation}

From Equation~\ref{sgd-expectation}, we know that $\mathop{\mathbb{E}}(\nabla_{\theta}L_w^i(\theta_t)^T(\theta_t - \theta^*)) = \mathop{\mathbb{E}}\left(\nabla_{\theta}L_w(\theta_T)^T(\theta_t- \theta^*)\right)$. Substituting it in the above equation, we have:
\begin{equation}
\begin{aligned}
\label{sgd-cndtn}
 \mathop{\mathbb{E}}({\nabla_{\theta}L(\theta_t)}^T(\theta_t - \theta^*)) &= \frac{1}{2\alpha_t}\mathop{\mathbb{E}}\left({\left\Vert \alpha_t  \nabla_{\theta}L_w(\theta_t)\right\Vert}^2 \right) + \frac{1}{2\alpha_t} \mathop{\mathbb{E}}\left({\left\Vert\theta_{t}
 - \theta^{*}\right\Vert}^2\right)\\
 &- \frac{1}{2\alpha_t} \mathop{\mathbb{E}}\left({\left\Vert\theta_{t+1} - \theta^{*}\right\Vert}^2\right) - \mathop{\mathbb{E}}\left({(\nabla_{\theta}L_w(\theta_t) - \nabla_{\theta}L(\theta_t))}^T(\theta_t - \theta^*)\right) 
\end{aligned}
\end{equation}

\begin{case}
\textbf{$L_T$ is lipschitz continuous with parameter $\sigma_T$ (i.e., $\left\Vert \nabla_{\theta}L_T(\theta)\right\Vert \leq \sigma_T$ )}

From convexity of function $L(\theta)$, we know $L(\theta_t) - L(\theta^*) \leq {\nabla_{\theta}L(\theta_t)}^T(\theta_t - \theta^*)$. Combining this with Equation~\ref{sgd-cndtn} we have,

\begin{equation}
\begin{aligned}
 \mathop{\mathbb{E}}(L(\theta_t) - L(\theta^*)) &\leq \frac{1}{2\alpha_t}\mathop{\mathbb{E}}\left({\left\Vert \alpha_t  \nabla_{\theta}L_w(\theta_t)\right\Vert}^2 \right) + \frac{1}{2\alpha_t} \mathop{\mathbb{E}}\left({\left\Vert\theta_{t}
 - \theta^{*}\right\Vert}^2\right)\\
 &- \frac{1}{2\alpha_t} \mathop{\mathbb{E}}\left({\left\Vert\theta_{t+1} - \theta^{*}\right\Vert}^2\right) - \mathop{\mathbb{E}}\left({(\nabla_{\theta}L_w(\theta_t) - \nabla_{\theta}L(\theta_t))}^T(\theta_t - \theta^*)\right) 
\end{aligned}
\end{equation}

Summing up the above equation from $t=0 \cdots T-1$, we have:
\begin{equation}
\begin{aligned}
 \sum_{t=0}^{T-1}\mathop{\mathbb{E}}(L(\theta_t) - L(\theta^*)) &\leq \sum_{t=0}^{T-1}\frac{1}{2\alpha_t}\mathop{\mathbb{E}}\left({\left\Vert \alpha_t  \nabla_{\theta}L_w(\theta_t)\right\Vert}^2 \right) + \frac{1}{2\alpha_t} \sum_{t=0}^{T-1}\mathop{\mathbb{E}}\left({\left\Vert\theta_{t}
 - \theta^{*}\right\Vert}^2\right)\\
 &- \frac{1}{2\alpha_t} \sum_{t=0}^{T-1}\mathop{\mathbb{E}}\left({\left\Vert\theta_{t+1} - \theta^{*}\right\Vert}^2\right) - \sum_{t=0}^{T-1}\mathop{\mathbb{E}}\left({(\nabla_{\theta}L_w(\theta_t) - \nabla_{\theta}L(\theta_t))}^T(\theta_t - \theta^*)\right) 
\end{aligned}
\end{equation}

Since $\mathop{\mathbb{E}}\left(\left\Vert\nabla_{\theta}L_T(\theta)\right\Vert\right) \leq \sigma_T$, we have $\mathop{\mathbb{E}}\left(\left\Vert\nabla_{\theta}L_w(\theta)\right\Vert\right) \leq \sigma_T$ as the weights are normalized to 1. Substituting it in the above equation, we have:

\begin{equation}
\begin{aligned}
 \sum_{t=0}^{T-1}\mathop{\mathbb{E}}(L(\theta_t) - L(\theta^*)) &\leq \sum_{t=0}^{T-1}\frac{\alpha_t \sigma_T^2}{2} + \frac{1}{2\alpha_t} \sum_{t=0}^{T-1}\mathop{\mathbb{E}}\left({\left\Vert\theta_{t}
 - \theta^{*}\right\Vert}^2\right)\\
 &- \frac{1}{2\alpha_t} \sum_{t=0}^{T-1}\mathop{\mathbb{E}}\left({\left\Vert\theta_{t+1} - \theta^{*}\right\Vert}^2\right) - \sum_{t=0}^{T-1}\mathop{\mathbb{E}}\left({(\nabla_{\theta}L_w(\theta_t) - \nabla_{\theta}L(\theta_t))}^T(\theta_t - \theta^*)\right) 
\end{aligned}
\end{equation}

\begin{equation}
\begin{aligned}
 \sum_{t=0}^{T-1}\mathop{\mathbb{E}}(L(\theta_t) - L(\theta^*)) &\leq \sum_{t=0}^{T-1}\frac{\alpha_t \sigma_T^2}{2} + \frac{1}{2\alpha_t} \mathop{\mathbb{E}}\left({\left\Vert\theta_{0}
 - \theta^{*}\right\Vert}^2\right)\\
 &- \frac{1}{2\alpha_t} \mathop{\mathbb{E}}\left({\left\Vert\theta_{T} - \theta^{*}\right\Vert}^2\right) - \sum_{t=0}^{T-1}\mathop{\mathbb{E}}\left({(\nabla_{\theta}L_w(\theta_t) - \nabla_{\theta}L(\theta_t))}^T(\theta_t - \theta^*)\right) 
\end{aligned}
\end{equation}

Since $\mathop{\mathbb{E}}\left({\left\Vert\theta_{T} - \theta^{*}\right\Vert}^2\right) \geq 0$, we have:

\begin{equation}
\begin{aligned}
 \sum_{t=0}^{T-1}\mathop{\mathbb{E}}(L(\theta_t) - L(\theta^*)) &\leq \sum_{t=0}^{T-1}\frac{\alpha_t \sigma_T^2}{2} + \frac{1}{2\alpha_t} \mathop{\mathbb{E}}\left({\left\Vert\theta_{0}
 - \theta^{*}\right\Vert}^2\right) - \sum_{t=0}^{T-1}\mathop{\mathbb{E}}\left({(\nabla_{\theta}L_w(\theta_t) - \nabla_{\theta}L(\theta_t))}^T(\theta_t - \theta^*)\right) 
\end{aligned}
\end{equation}

Also assuming that $\left\Vert\theta - \theta^{*}\right\Vert \leq D$, we have,

\begin{equation}
\begin{aligned}
 \sum_{t=0}^{T-1}\mathop{\mathbb{E}}(L(\theta_t) - L(\theta^*)) &\leq \frac{\alpha_t T \sigma_T^2}{2} + \frac{D^2}{2 \alpha_t} - D \sum_{t=0}^{T-1}\mathop{\mathbb{E}}\left(\left\Vert\nabla_{\theta}L_w(\theta_t) - \nabla_{\theta}L(\theta_t)\right\Vert\right) 
\end{aligned}
\end{equation}

Choosing a constant learning rate of $\alpha_t = \frac{D}{\sigma_T \sqrt{T}}$, we have:

\begin{equation}
\begin{aligned}
 \sum_{t=0}^{T-1}\mathop{\mathbb{E}}(L(\theta_t) - L(\theta^*)) &\leq \frac{D \sigma_T \sqrt{T}}{2} + \frac{D \sigma_T \sqrt{T}}{2} - D \sum_{t=0}^{T-1}\mathop{\mathbb{E}}\left(\left\Vert\nabla_{\theta}L_w(\theta_t) - \nabla_{\theta}L(\theta_t)\right\Vert\right) 
\end{aligned}
\end{equation}

Since, $L(\theta_t) - L(\theta^*) \leq \min_{t = 1:T} L(\theta_t) - L(\theta^*)$, we have: 

\begin{equation}
\begin{aligned}
 \sum_{t=0}^{T-1}\mathop{\mathbb{E}}(\min_{t = 1:T} L(\theta_t) - L(\theta^*)) &\leq \frac{D \sigma_T \sqrt{T}}{2} + \frac{D \sigma_T \sqrt{T}}{2} - D \sum_{t=0}^{T-1}\mathop{\mathbb{E}}\left(\left\Vert\nabla_{\theta}L_w(\theta_t) - \nabla_{\theta}L(\theta_t)\right\Vert\right) 
\end{aligned}
\end{equation}

Dividing the above equation by $T$ in the both sides, we have:
\begin{equation}
\begin{aligned}
 \frac{1}{T}\sum_{t=0}^{T-1}\mathop{\mathbb{E}}(\min_{t = 1:T} L(\theta_t) - L(\theta^*)) &\leq \sum_{t=0}^{T-1}\frac{D \sigma_T}{\sqrt{T}} - \frac{D}{T} \sum_{t=0}^{T-1}\mathop{\mathbb{E}}\left(\left\Vert\nabla_{\theta}L_w(\theta_t) - \nabla_{\theta}L(\theta_t)\right\Vert\right) 
\end{aligned}
\end{equation}

\begin{equation}
\begin{aligned}
 \mathop{\mathbb{E}}(\min_{t = 1:T} L(\theta_t) - L(\theta^*)) &\leq \sum_{t=0}^{T-1}\frac{D \sigma_T}{\sqrt{T}} - \frac{D}{T} \sum_{t=0}^{T-1}\mathop{\mathbb{E}}\left(\left\Vert\nabla_{\theta}L_w(\theta_t) - \nabla_{\theta}L(\theta_t)\right\Vert\right) 
\end{aligned}
\end{equation}

Substituting $\nabla_{\theta}L_w(\theta_t) = \underset{i \in \Xcal^t}{\sum} w^t_i \nabla_{\theta}L_T^i(\theta_t)$, we have:

\begin{equation}
\begin{aligned}
 \mathop{\mathbb{E}}\left(\min_{t = 1:T} L(\theta_t)\right) - L(\theta^*) &\leq \sum_{t=0}^{T-1}\frac{D \sigma_T}{\sqrt{T}} - \frac{D}{T} \sum_{t=0}^{T-1}\mathop{\mathbb{E}}\left(\left\Vert\underset{i \in \Xcal^t}{\sum} w^t_i \nabla_{\theta}L_T^i(\theta_t) - \nabla_{\theta}L(\theta_t)\right\Vert\right) 
\end{aligned}
\end{equation}
\end{case}

\begin{case}
\textbf{$L_T$ is Lipschitz continuous, $L$ is strongly convex with a strong convexity parameter $\mu$}

From strong convexity of function $L(\theta)$, we have:

\begin{equation}
    \begin{aligned}
         \mathop{\mathbb{E}}\left({\nabla_{\theta}L(\theta_t)}^T(\theta_t - \theta^*)\right) \geq \mathop{\mathbb{E}}\left(L(\theta_t) - L(\theta^*)\right) + \frac{\mu}{2}\mathop{\mathbb{E}}\left(\left\Vert\theta_t - \theta^*\right\Vert\right)
    \end{aligned}
\end{equation}

Combining the above equation with Equation~\ref{sgd-cndtn}, we have:

\begin{equation}
\begin{aligned}
 \mathop{\mathbb{E}}\left(L(\theta_t) - L(\theta^*)\right) + \frac{\mu}{2}\mathop{\mathbb{E}}\left(\left\Vert\theta_t - \theta^*\right\Vert\right) &\leq \frac{1}{2\alpha_t}\mathop{\mathbb{E}}\left({\left\Vert \alpha_t  \nabla_{\theta}L_w(\theta_t)\right\Vert}^2 \right) + \frac{1}{2\alpha_t} \mathop{\mathbb{E}}\left({\left\Vert\theta_{t}
 - \theta^{*}\right\Vert}^2\right)\\
 &- \frac{1}{2\alpha_t} \mathop{\mathbb{E}}\left({\left\Vert\theta_{t+1} - \theta^{*}\right\Vert}^2\right) - \mathop{\mathbb{E}}\left({(\nabla_{\theta}L_w(\theta_t) - \nabla_{\theta}L(\theta_t))}^T(\theta_t - \theta^*)\right) 
\end{aligned}
\end{equation}

\begin{equation}
\begin{aligned}
 \mathop{\mathbb{E}}\left(L(\theta_t) - L(\theta^*)\right)  &\leq \frac{1}{2\alpha_t}\mathop{\mathbb{E}}\left({\left\Vert \alpha_t  \nabla_{\theta}L_w(\theta_t)\right\Vert}^2 \right) + \frac{1}{2\alpha_t} \mathop{\mathbb{E}}\left({\left\Vert\theta_{t}
 - \theta^{*}\right\Vert}^2\right)\\
 &- \frac{1}{2\alpha_t} \mathop{\mathbb{E}}\left({\left\Vert\theta_{t+1} - \theta^{*}\right\Vert}^2\right) - \mathop{\mathbb{E}}\left({(\nabla_{\theta}L_w(\theta_t) - \nabla_{\theta}L(\theta_t))}^T(\theta_t - \theta^*)\right)  - \frac{\mu}{2}\mathop{\mathbb{E}}\left(\left\Vert\theta_t - \theta^*\right\Vert\right)
\end{aligned}
\end{equation}

\begin{equation}
\begin{aligned}
 \mathop{\mathbb{E}}\left(L(\theta_t) - L(\theta^*)\right)  &\leq \frac{1}{2\alpha_t}\mathop{\mathbb{E}}\left({\left\Vert \alpha_t  \nabla_{\theta}L_w(\theta_t)\right\Vert}^2 \right) + \frac{\alpha_t^{-1} - \mu}{2} \mathop{\mathbb{E}}\left({\left\Vert\theta_{t}
 - \theta^{*}\right\Vert}^2\right)\\
 &- \frac{\alpha_t^{-1}}{2} \mathop{\mathbb{E}}\left({\left\Vert\theta_{t+1} - \theta^{*}\right\Vert}^2\right) - \mathop{\mathbb{E}}\left({(\nabla_{\theta}L_w(\theta_t) - \nabla_{\theta}L(\theta_t))}^T(\theta_t - \theta^*)\right)
\end{aligned}
\end{equation}

Setting an learning rate of $\alpha_t = \frac{2}{\mu(t+1)}$, multiplying by $t$ on both sides and substituting $\nabla_{\theta}L_w(\theta_t) = \sum_{i \in \Xcal^t} w^t_i \nabla_{\theta}L_T^i(\theta_t)$, we have:
\begin{equation}
\begin{aligned}
\mathop{\mathbb{E}}\left(t(L(\theta_t) - L(\theta^*))\right) &= \mathop{\mathbb{E}}\left(\frac{t}{\mu(t+1)}{\left\Vert \sum_{i \in \Xcal^t} w^t_i \nabla_{\theta}L_T^i(\theta_t)\right\Vert}^2\right) + \mathop{\mathbb{E}}\left(\frac{\mu t(t-1)}{4}{\left\Vert\theta_{t} - \theta^{*}\right\Vert}^2\right)\\
&- \mathop{\mathbb{E}}\left(\frac{\mu t(t+1)}{4}{\left\Vert\theta_{t+1} - \theta^{*}\right\Vert}^2\right)  - \mathop{\mathbb{E}}\left(t {\left(\nabla_{\theta}L_w(\theta_t) - \nabla_{\theta}L(\theta_t)\right)}^T(\theta_t - \theta^*)\right)
\end{aligned}
\end{equation}

Since, $\left\Vert L_T(\theta)\right\Vert \leq \sigma_T$, we have ${\left\Vert \alpha \sum_{i \in \Xcal^t} w^t_i \nabla_{\theta}L_T^i(\theta_t)\right\Vert} \leq \sum_{i=1}^{|\Xcal^t|}w_i^t \sigma_T$. Assuming that the weights at every iteration are normalized such that $\underset{t \in [1, T]}{\forall} \sum_{i=1}^{|\Xcal^t|}w_i^t = 1$ and the training and validation loss gradients are normalized as well, we have ${\left\Vert \alpha \sum_{i \in \Xcal^t} w^t_i \nabla_{\theta}L_T^i(\theta_t)\right\Vert} \leq \sigma_T$. Also assuming that $\left\Vert\theta - \theta^{*}\right\Vert \leq D$, we have,

\begin{equation}
\begin{aligned}
\mathop{\mathbb{E}}\left(t(L(\theta_t) - L(\theta^*))\right) &= \frac{{\sigma_T}^2 t}{\mu(t+1)} + \frac{\mu t(t-1)}{4}\mathop{\mathbb{E}}\left({\left\Vert\theta_{t} - \theta^{*}\right\Vert}^2\right)\\
&- \frac{\mu t(t+1)}{4}\mathop{\mathbb{E}}\left({\left\Vert\theta_{t+1} - \theta^{*}\right\Vert}^2\right)  + \mathop{\mathbb{E}}\left(Dt \left \Vert \nabla_{\theta}L_w(\theta_t) - \nabla_{\theta}L(\theta_t)\right\Vert\right)
\end{aligned}
\end{equation}

Summing the above equation from $t=1,\cdots,T$, we have:
\begin{equation}
\begin{aligned}
\sum_{t=1}^{t=T}\mathop{\mathbb{E}}\left(t(L(\theta_t) - L(\theta^*))\right) &= \sum_{t=1}^{t=T}\frac{{\sigma_T}^2 t}{\mu(t+1)} + \sum_{t=1}^{t=T}\frac{\mu t(t-1)}{4}\mathop{\mathbb{E}}\left({\left\Vert\theta_{t} - \theta^{*}\right\Vert}^2\right)\\
&- \sum_{t=1}^{t=T} \frac{\mu t(t+1)}{4}\mathop{\mathbb{E}}\left({\left\Vert\theta_{t+1} - \theta^{*}\right\Vert}^2\right)  + \sum_{t=1}^{t=T} D\mathop{\mathbb{E}}\left(t \left \Vert \nabla_{\theta}L_w(\theta_t) - \nabla_{\theta}L(\theta_t)\right\Vert\right)
\end{aligned}
\end{equation}

\begin{equation}
\begin{aligned}
\sum_{t=1}^{t=T}\mathop{\mathbb{E}}\left(t(L(\theta_t) - L(\theta^*))\right) &\leq \sum_{t=1}^{t=T}\frac{{\sigma_T}^2}{\mu} + \sum_{t=1}^{t=T}\frac{\mu t(t-1)}{4}\mathop{\mathbb{E}}\left({\left\Vert\theta_{t} - \theta^{*}\right\Vert}^2\right)\\
&- \sum_{t=1}^{t=T} \frac{\mu t(t+1)}{4}\mathop{\mathbb{E}}\left({\left\Vert\theta_{t+1} - \theta^{*}\right\Vert}^2\right)  + \sum_{t=1}^{t=T} D \mathop{\mathbb{E}}\left(t \left \Vert \nabla_{\theta}L_w(\theta_t) - \nabla_{\theta}L(\theta_t)\right\Vert\right)
\end{aligned}
\end{equation}

\begin{equation}
\begin{aligned}
\sum_{t=1}^{t=T}\mathop{\mathbb{E}}\left(t(L(\theta_t) - L(\theta^*))\right) &\leq \frac{{\sigma_T}^2 T}{\mu } + \frac{\mu}{4}(0 - T(T+1)\mathop{\mathbb{E}}\left({\left\Vert\theta_{T+1} - \theta^{*}\right\Vert}^2)\right)\\
&+ \sum_{t=1}^{t=T} D \mathop{\mathbb{E}}\left(t \left \Vert \nabla_{\theta}L_w(\theta_t) - \nabla_{\theta}L(\theta_t)\right\Vert\right)
\end{aligned}
\end{equation}

Since $\frac{\mu}{4}(0 - T(T+1)\mathop{\mathbb{E}}\left({\left\Vert\theta_{T+1} - \theta^{*}\right\Vert}^2\right)) \leq 0$, we have:

\begin{equation}
\begin{aligned}
\sum_{t=1}^{t=T}\mathop{\mathbb{E}}\left(t(L(\theta_t) - L(\theta^*))\right) &\leq \frac{{\sigma_T}^2 T}{\mu } + \sum_{t=1}^{t=T} D\mathop{\mathbb{E}}\left(t \left \Vert \nabla_{\theta}L_w(\theta_t) - \nabla_{\theta}L(\theta_t)\right\Vert\right)
\end{aligned}
\end{equation}

Since, $L(\theta_t) - L(\theta^*) \leq \min_{t = 1:T} L(\theta_t) - L(\theta^*)$ and multiplying the above equation by $\frac{2}{T(T+1)}$, we have: 

\begin{equation}
\begin{aligned}
\frac{2}{T(T+1)}\mathop{\mathbb{E}}\left(\sum_{t=1}^{t=T}t(\min_{t = 1:T} L(\theta_t) - L(\theta^*))\right) &\leq \frac{2}{T(T+1)} \frac{{\sigma_T}^2 T}{\mu } + \frac{2}{T(T+1)} \sum_{t=1}^{t=T} \mathop{\mathbb{E}}\left(Dt \left \Vert \nabla_{\theta}L_w(\theta_t) - \nabla_{\theta}L(\theta_t)\right\Vert\right)
\end{aligned}
\end{equation}

This in turn implies:
\begin{equation}
\begin{aligned}
\mathop{\mathbb{E}}\left(\min_{t = 1:T} L(\theta_t)\right) - L(\theta^*) &\leq \frac{{\sigma_T}^2 2}{\mu (T+1)} + \sum_{t=1}^{t=T} \frac{2D}{T(T+1)} \mathop{\mathbb{E}}\left(t\left \Vert \nabla_{\theta}L_w(\theta_t) - \nabla_{\theta}L(\theta_t)\right\Vert\right)
\end{aligned}
\end{equation}

\end{case}

\end{proof}

\subsection{Conditions for adaptive data selection algorithms to reduce the objective value at every iteration}
\label{app-reduce-loss-dss}
We provide conditions under which the adaptive subset selection strategy reduces the objective value of $L$ (which can either be the training loss $L_T$ or the validation loss $L_V$):
\begin{theorem}
\label{thm:reduceobjective}
If the Loss Function $L$ is Lipschitz smooth with parameter $\mathcal L$, and the gradient of the training loss is bounded by $\sigma_T$, the adaptive data selection algorithm will reduce the objective function at every iteration, i.e. $L(\theta_{t+1}) \leq L(\theta_t)$ as long as $(\sum_{i \in \Xcal} w_i \nabla_{\theta} L_T^i(\theta))^T \nabla_{\theta} L(\theta) \geq 0$ and the learning rate schedule satisfies $\alpha_t \leq \min_t 2\frac{\Vert \nabla_{\theta} L(\theta)\rVert \cos(\theta_t)}{\mathcal{L} \sigma_T}$, where $\theta_t$ is the angle between $\sum_{i \in \Xcal} w_i \nabla_{\theta} L_T^i(\theta)$ and $\nabla_{\theta} L(\theta)$.
\end{theorem}

Before proving this result, notice that any data selection approach that attempts to minimize the error term $\mbox{Err}(\wb^t, \Xcal^t, L, L_T, \theta_t) = {\Vert \sum_{i \in \Xcal^t} w^t_i \nabla_{\theta}L_T^i(\theta_t) -  \nabla_{\theta}L(\theta_t)\Vert}$, will essentially also maximize $\left(\sum_{i \in \Xcal^t} w^t_i \nabla_{\theta} L_T^i(\theta)\right)^T \nabla_{\theta} L(\theta)$. Hence we expect the condition above to be satisfied, as long as the learning rate can be selected appropriately.

\begin{proof}
Suppose we have a validation set $\mathcal{V}$ and the loss on the validation set or training set is denoted as $L(\theta)$  depending on the usage. Suppose the subset selected by the \model{}  is denoted by $S$ and the subset training loss is $L_T(\theta, \Xcal)$. Since validation or training loss $L$ is lipschitz smooth, we have,
\begin{equation}
    \begin{aligned}
    \label{lipschitz-smooth}
    L(\theta_{t+1}) \leq  L(\theta_t) + \frac{\mathcal{L}{\left\Vert\Delta \theta\right\Vert}^2}{2} + \nabla_{\theta}{L(\theta_t)}^{T} \Delta \theta, \text{\hspace{0.5cm}where,\hspace{0.2cm}} \Delta \theta = \theta_{t+1} - \theta_t \text{\hspace{3cm}}
    \end{aligned}
\end{equation}

Since, we are using SGD to optimize the subset training loss $L_T(\theta, S)$ model parameters our update equations will be as follows:
\begin{equation}
\label{sgd}
    \theta_{t+1} = \theta_t - \alpha \sum_{i \in \Xcal} w_i \nabla_{\theta} L_T^i(\theta_t)
\end{equation}

Plugging our updating rule (Equation~\ref{sgd}) in (Equation\ref{lipschitz-smooth}):
\begin{equation}
    \begin{aligned}
    \label{lipschitz-smooth1}
    L(\theta_{t+1})  \leq  L(\theta_t) - \alpha \nabla_{\theta}{L(\theta_t)}^T \sum_{i \in \Xcal} w_i \nabla_{\theta} L_T^i(\theta))  + \frac{\mathcal{L}}{2}{\left\Vert- \alpha \sum_{i \in \Xcal} w_i \nabla_{\theta} L_T^i(\theta)\right\Vert}^2
    \end{aligned}
\end{equation}

Which gives,
\begin{equation}
    \begin{aligned}
    \label{final-equation}
    L(\theta_{t+1}) - L(\theta_t) \leq - \alpha \nabla_{\theta}{L(\theta_t)}^T \sum_{i \in \Xcal} w_i \nabla_{\theta} L_T^i(\theta)  + \frac{\mathcal{L}\alpha^2}{2}{\left\Vert \sum_{i \in \Xcal} w_i \nabla_{\theta} L_T^i(\theta)\right\Vert}^2
    \end{aligned}
\end{equation}

From (Equation~\ref{final-equation}), note that:
\begin{equation}
\begin{aligned}
\label{thm2-cndtn}
L(\theta_{t+1}) \leq L(\theta_{l}) \text{\hspace{0.2cm}if\hspace{0.2cm}} \alpha \left({\bigg(\sum_{i \in \Xcal} w_i \nabla_{\theta} L_T^i(\theta)\bigg)}^T \nabla_{\theta}{L(\theta_t)} - \frac{\mathcal{L} \alpha}{2}{\left\Vert \sum_{i \in \Xcal} w_i \nabla_{\theta} L_T^i(\theta)\right\Vert}^2\right) \geq 0
\end{aligned}
\end{equation}

Since we know that ${\left\Vert \sum_{i \in \Xcal} w_i \nabla_{\theta} L_T^i(\theta)\right\Vert}^2 \geq 0$, we will have the necessary condition:
$${\bigg(\sum_{i \in \Xcal} w_i \nabla_{\theta} L_T^i(\theta)\bigg)}^T \nabla_{\theta}{L(\theta_t)} \geq 0$$

We can also re-write the condition in (Equation\ref{thm2-cndtn}) as follows:

\begin{equation}
\begin{aligned}
\label{learning_rate_cndtn}
    {\bigg(\sum_{i \in \Xcal} w_i \nabla_{\theta} L_T^i(\theta)\bigg)}^T \nabla_{\theta}{L(\theta_t)} \geq \frac{\alpha \mathcal{L}}{2} {\bigg(\sum_{i \in \Xcal} w_i \nabla_{\theta} L_T^i(\theta)\bigg)}^T{\bigg(\sum_{i \in \Xcal} w_i \nabla_{\theta} L_T^i(\theta)\bigg)}
\end{aligned}
\end{equation}

The Equation~\ref{learning_rate_cndtn} gives the necessary condition for learning rate i.e.,
\begin{equation}
\begin{aligned}
\label{eta_cndtn}
    \alpha \leq \frac{2}{\mathcal{L}} \frac{{\bigg(\sum_{i \in \Xcal} w_i \nabla_{\theta} L_T^i(\theta)\bigg)}^T \nabla_{\theta}{L(\theta_t)}}{{\bigg(\sum_{i \in \Xcal} w_i \nabla_{\theta} L_T^i(\theta)\bigg)}^T{\bigg(\sum_{i \in \Xcal} w_i \nabla_{\theta} L_T^i(\theta)\bigg)}}
\end{aligned}
\end{equation}

The above Equation~\ref{eta_cndtn} can be written as follows:
\begin{equation}
\begin{aligned}
    \alpha \leq \frac{2}{\mathcal{L}} \frac{\left\Vert \nabla_{\theta}L(\theta_t)\right\Vert \cos(\Theta_t)}{\left\Vert\sum_{i \in \Xcal} w_i \nabla_{\theta} L_T^i(\theta)\right\Vert}\\
    \text{where\hspace{0.2cm}} \cos{\Theta_t} = \frac{{\bigg(\sum_{i \in \Xcal} w_i \nabla_{\theta} L_T^i(\theta)\bigg)}^T\nabla_{\theta}{L(\theta_t)}}{\left\Vert\sum_{i \in \Xcal} w_i \nabla_{\theta} L_T^i(\theta)\right\Vert\left\Vert\nabla_{\theta}L(\theta_t)\right\Vert}
\end{aligned}
\end{equation}

Assuming we normalize the subset weights at every iteration i.e., $\underset{t}{\forall} \sum_{i \in [1, |\Xcal|} w_i^t = 1$, we know that the gradient norm $\left\Vert\sum_{i \in \Xcal} w_i \nabla_{\theta} L_T^i(\theta)\right\Vert \leq \sigma_T$, the condition for the learning rate can be written as follows,
\begin{equation}
\begin{aligned}
\label{step-size-cndtn}
    \alpha \leq \frac{2\left\Vert \nabla_{\theta}L(\theta_t)\right\Vert \cos(\Theta_t)}{\mathcal{L}\sigma_T} \text{\hspace{0.2cm}where\hspace{0.2cm}} \cos{\Theta_t} = \frac{{\bigg(\sum_{i \in \Xcal} w_i \nabla_{\theta} L_T^i(\theta)\bigg)}^T\nabla_{\theta}{L(\theta_t)}}{\left\Vert\sum_{i \in \Xcal} w_i \nabla_{\theta} L_T^i(\theta)\right\Vert\left\Vert\nabla_{\theta}L(\theta_t)\right\Vert}
\end{aligned}
\end{equation}

Since, the condition mentioned in Equation~\ref{step-size-cndtn} needs to be true for all values of $l$, we have the condition for learning rate as follows:
\begin{equation}
\begin{aligned}
\label{final-step-size-cndtn}
    \alpha \leq \min_t \frac{2\left\Vert \nabla_{\theta}L(\theta_t)\right\Vert \cos(\Theta_t)}{\mathcal{L}\sigma_T} \\\text{\hspace{0.2cm}where\hspace{0.2cm}} \cos{\Theta_t} = \frac{{\bigg(\sum_{i \in \Xcal} w_i \nabla_{\theta} L_T^i(\theta)\bigg)}^T\nabla_{\theta}{L(\theta_t)}}{\left\Vert\sum_{i \in \Xcal} w_i \nabla_{\theta} L_T^i(\theta)\right\Vert\left\Vert\nabla_{\theta}L(\theta_t)\right\Vert}
\end{aligned}
\end{equation}
\end{proof}

\subsection{Proof of Theorem~\ref{thm:approx-submod-result}}
\label{app-approx-submod-result}
We first restate Theorem~\ref{thm:approx-submod-result}
\begin{theorem-nono}
   If $|\Xcal|\leq k,\ \max_{i} ||\nabla_{\theta} L^i _T(\theta_t)||_{2} < \nabla_{\max}$, then $F_{\lambda}(\Xcal)$ is $\gamma$-weakly submodular, with
    $\gamma \geq \frac{\lambda}{\lambda+k\nabla_{\max} ^2  }$
\end{theorem-nono}
\begin{proof}
We first note that the minimum eigenvalue
of $F_{\lambda}(\Xcal)$ is atleast $\lambda$.
Next, we note that the maximum eigenvalue of
$F_{\lambda}(\Xcal)$ is atmost 
\begin{align}
    \lambda + \text{Trace}(F(\Xcal))\\
    &= \lambda + \text{Trace}\left(\left[ \begin{array}{c} \nabla L^{1\top} _T(\theta_t) \\
    \nabla L^{2\top} _T(\theta_t) \\
    \ldots\\
    \nabla L^{k\top} _T(\theta_t) 
    \end{array}\right]\left[ \begin{array}{c} \nabla L^{1\top} _T(\theta_t) \\
    \nabla L^{2\top} _T(\theta_t) \\
    \ldots\\
    \nabla L^{k\top} _T(\theta_t) 
    \end{array}\right] ^\top \right)\\
    & = \lambda + \sum_{i\in[k]} || \nabla L^{i} _T(\theta_t)  ||^2
\end{align}
which immediately proves the theorem following~\cite{elenberg2018restricted}.
\end{proof}

\subsection{Proof of Theorem~\ref{weak-submod-setcover}}
\label{app-weak-submod-setcover}
We start this subsection by first restating Theorem~\ref{weak-submod-setcover}.
\begin{theorem-nono}
If the function $F_{\lambda}(\Xcal)$ is $\gamma$-weakly submodular, $\Xcal^*$ is the optimal subset and $\max_{i} ||\nabla_{\theta} L^i _T(\theta_t)||_{2} < \nabla_{\max}$, (both) the greedy algorithm and OMP (Algorithm~\ref{alg:algo1}), run with stopping criteria  $E_{\lambda}(\Xcal) \leq \epsilon$ achieve sets $\Xcal$ such that $|\Xcal| \leq \frac{|\Xcal^*|}{\gamma} \log(\frac{L_{\max}}{\epsilon})$ where $L_{\max}$ is an upper bound of $F_{\lambda}$ .
\end{theorem-nono}
\begin{proof}
From Theorem~\ref{weak-submod-setcover}, we know that $F_{\lambda}(\Xcal)$ is weakly submodular with parameter $\gamma = \frac{\lambda}{\lambda+k\nabla_{\max} ^2}$.

We first prove the result using the greedy algorithm, or in particular the submodular set cover algorithm~\cite{wolsey1982analysis}. Note that an upper bound of $F_{\lambda}$ is $L_{\max}$, and consider the stopping criteria of the greedy algorithm to be achieving a subset $\Xcal$ such that $F_{\lambda}(\Xcal) \geq L_{\max} - \epsilon$. The goal is then to bound the $|\Xcal|$ of the subset $\Xcal$ achieving it compared to the optimal subset $\Xcal^*$. 

Given a set $X_i$ which is obtained at step $i$ of the greedy algorithm, and denote $e_i$ to be the best gain at step $i$. Note that:
\begin{align}
    \gamma (F_{\lambda}(\Xcal^* \cup X_i) - F_{\lambda}(X_i)) &\leq \sum_{j \in \Xcal^*} F_{\lambda}(j | X_i) \nonumber \\
    &\leq |\Xcal^*| F_{\lambda}(e_i | X_i)
\end{align}
where the last inequality holds because of the greedy algorithm. This then implies that:
\begin{align} \label{recursion-rel}
    F_{\lambda}(\Xcal^*) - F_{\lambda}(X_i) \leq \frac{|\Xcal^*|}{\gamma} (F_{\lambda}(X_{i+1}) - F_{\lambda}(X_i))
\end{align}
We modify the second term to be $(F_{\lambda}(X_{i+1}) - F_{\lambda}(X_i)) = (F_{\lambda}(\Xcal^*) - F_{\lambda}(X_i)) - (F_{\lambda}(\Xcal^*) - F_{\lambda}(X_{i+1}))$ and then obtain the following recursion:

\begin{align}
   F_{\lambda}(\Xcal^*) - F_{\lambda}(X_{i+1})  \leq (1 - \frac{\gamma}{|\Xcal^*|}) (F_{\lambda}(\Xcal^*) - F_{\lambda}(X_{i}))
\end{align}
We can then recursively multiply the right hand and the left hand sides, until we reach a set $\Xcal$ such that $F_{\lambda}(\Xcal) \geq L_{\max} - \epsilon$ (which is the stopping criteria). We then achieve:
\begin{align}
    F_{\lambda}(\Xcal^*) - F_{\lambda}(\Xcal) \leq (1 - \gamma/|\Xcal^*|)^{|\Xcal|} (F_{\lambda}(\Xcal^*) - F_{\lambda}(\emptyset)) \leq (1 - \gamma/|\Xcal^*|)^{|\Xcal|} F_{\lambda}(\Xcal^*) \leq (1 - \gamma/|\Xcal^*|)^{|\Xcal|} L_{\max}
\end{align}
where the second-last inequality holds since $F_{\lambda}(\emptyset) \geq 0$, and the last inequality holds because $F_{\lambda}(\Xcal^*) \leq L_{\max}$. 

This implies that we have the following inequality: $F_{\lambda}(\Xcal^*) - F_{\lambda}(\Xcal) \leq (1 - \gamma/|\Xcal^*|)^{|\Xcal|} L_{\max}$. Next, notice that since $F_{\lambda}(\Xcal^*) \leq L_{\max}$, which in turn implies that $F_{\lambda}(\Xcal^*) - F_{\lambda}(\Xcal) \leq \epsilon$. Hence, we can pick a $\Xcal$ such that $(1 - \gamma/|\Xcal^*|)^{|\Xcal|} L_{\max} \leq \epsilon$, which will then automatically imply that $F_{\lambda}(\Xcal^*) - F_{\lambda}(\Xcal) \leq \epsilon$. The above condition requires $(1 - \gamma/|\Xcal^*|)^{|\Xcal|} \geq \epsilon$, which in turn implies that $|\Xcal| \leq |\Xcal^*|/\gamma \log L_{\max}/\epsilon$. This shows the result for the standard greedy algorithm.

Finally, we prove it for the OMP case. In particular, from Lemma 4 and the proof of Theorem 5 in~\cite{elenberg2018restricted}, we can obtain a recursion very similar to Equation~\ref{recursion-rel}, except that we have the ratio of the $m$ and $M$ corresponding to strong concavity and smoothness respectively. From the proof of Theorem~\ref{thm:approx-submod-result}, this is exactly the bound used for weak submodularity of $F_{\lambda}$, and hence the bound follows for OMP as well.
\end{proof}

\subsection{Convergence result for \model\ using the OMP algorithm}
\label{app-omp-convergence}
The following result shows the convergence bound of \model\ using OMP as the optimization algorithm.
\begin{lemma}
Suppose the subsets $\Xcal^t$ satisfy the condition that $E_{\lambda}(\Xcal^t) \leq \epsilon$, for all $t = 1, \cdots, T$, then  OMP based data selection achieves the following convergence result: 
\begin{itemize}[leftmargin=*,noitemsep]
    \item if $L_T$ is lipschitz continuous with parameter $\sigma_T$ and $\alpha = \frac{D}{\sigma_T \sqrt{T}}$, then $\min_{t = 1:T} L(\theta_t) - L(\theta^*) \leq \frac{D\sigma_T}{\sqrt{T}} + D\epsilon$,
    \item if $L_T$ is lipschitz smooth with parameter $\mathcal{L}_T$, and $L_T^i$ satisfies $0 \leq L_T^i(\theta) \leq \beta_T, \forall i$. Then setting $\alpha = \frac{1}{\mathcal{L}_T}$, we have $\min_{t = 1:T} L(\theta_t) - L(\theta^*) \leq \frac{D^2 \mathcal{L}_T + 2\beta_T}{2T} + D\epsilon$,
    \item if $L_T$ is Lipschitz continuous (parameter $\sigma_T$) and $L$ is strongly convex with parameter $\mu$, then setting a learning rate $\alpha_t = \frac{2}{\mu(1+t)}$, achieves $\min_{t = 1:T} L(\theta_t) - L(\theta^*) \leq \frac{2{\sigma_T}^2}{\mu (T+1)} + D\epsilon$
\end{itemize}
\end{lemma}
\begin{proof}
We prove the first part and note that the other parts follow similarly. Notice that the stopping criteria of Algorithm 1 is $E_{\lambda}(\Xcal^t) \leq \epsilon$. Denote $w_t$ as the corresponding weight vector, and hence we have $E_{\lambda}(\Xcal^t) = E_{\lambda}(\Xcal^t, w_t) = E(\Xcal^t, w_t) + \lambda ||w_t||^2 \leq \epsilon$, where $E(\Xcal^t, w_t) = \mbox{Err}(w_t, \Xcal^t, L, L_T, \theta_t)$. Since $||w_t||^2 \geq 0$, this implies that $\mbox{Err}(w_t, \Xcal^t, L, L_T, \theta_t) \leq \epsilon$, which combining with Theorem~\ref{thm:convergence-result}, immediately provides the required convergence result for OMP. Finally, note that for the third part, $\sum_{t=1}^{t=T} \frac{2Dt}{T(T+1)}\epsilon = \epsilon$ and this proves all three parts. 
\end{proof}

\subsection{More details on \textsc{Craig}}
\label{app-craig-details}
\subsubsection{Connections between \textsc{Grad-Match} and \textsc{Craig}}
\begin{lemma}
The following inequality connects $\hat{E}(\Xcal)$ and $E(\Xcal)$
\begin{align}
E(\Xcal)& = \min_{\wb} \mbox{Err}( {\wb}, \Xcal, L, L_T, \theta_t) \leq \hat{E}(\Xcal)
\label{eq:min_upper}
\end{align}
Furthermore, given the set $\Xcal^t$ obtained by optimizing $\hat{E}$, the weights can be computed as: $\wb^t=\sum_{i\in \Wcal} \mathbb{I} \big[ j = {\arg\min_{s \in \Xcal^t}} \| \nabla_{\theta} L_T^i(\theta_t) - \nabla_{\theta} L^{s}(\theta_t) \|]$.
\end{lemma}
\begin{proof}
During iteration $t\in 1, \cdots, T$, we partition $W$  by assigning every element $i\in W$ to an  element $\pi_t^i \in \Xcal$ as follows: 
\begin{align}
 \pi_t^i \in {\arg\min}_{j \in \Xcal} \| \nabla_{\theta} L^i(\theta) - \nabla_{\theta} L_T^{j}(\theta) \|   
\end{align}
In other words, $\pi_t^i$ denotes the representative for a specific $i \in W$ in set $\Xcal$. Also recall that, $\hat{E}(\Xcal)$ is defined as follows:
\begin{align}
    \hat{E}(\Xcal) & =  \sum_{i\in W} \min_{j \in \Xcal} \| \nabla_{\theta} L^i(\theta_t) - \nabla_{\theta} L_T^j(\theta_t) \| \nonumber \\ & = \sum_{i\in W}  \| \nabla_{\theta} L^i(\theta_t) - \nabla_{\theta} L_T^{\pi_t^i}(\theta_t) \|
\end{align}

Then, for any $\theta_t$ we can write 
\begin{align}
\nabla_{\theta} L(\theta_t) 
&\!=\!  \sum_{i \in W}  \!\big( \nabla_{\theta} L^i_T(\theta_t) \!-\! \nabla_{\theta} L^{\pi_t^i}(\theta_t) \!+\!\nabla_{\theta} L^{\pi_t^i}(\theta_t) \big)\!\!\nonumber\\
= \sum_{i\in W}  \big( &\nabla_{\theta} L_T^i(\theta_t) - \nabla_{\theta} L^{\pi_t^i}(\theta_t) \big) + \sum_{j\in \Xcal} w_j \nabla_{\theta} L^{j}(\theta_t)\nonumber
\end{align}
where $w_j$ denotes the count of number of $i \in W$ that were assigned to an element $j \in \Xcal$. 
Subtracting the second term on the RHS, {\em viz.}, $\sum_{j\in \Xcal} w_j \nabla_{\theta} L^{j}(\theta_t)$ from the LHS and then taking the norm of the both sides, we get the following upper bound on the error of estimating the full gradient $\mbox{Err}({\mathbf w}, \Xcal, L, L_T, \theta_t)$: %for $\!j \!\in \!\Xcal$. I.e.,%, that is: 
\begin{align}
\mbox{Err}({\mathbf w}, \Xcal, L, L_T, &\theta_t) = \big\| \nabla L(\theta_t) - \sum_{i \in \Xcal} w_i \nabla L_T^i(\theta_t) \big\| \nonumber\\
&= \bigg\|\sum_{i\in W} \big(  \nabla_{\theta} L_T^i(\theta_t) - \nabla_{\theta} L^{\pi_t^i}(\theta_t) \big)  \bigg\|\nonumber\\
& \leq \sum_{i\in W}  \| \nabla_{\theta} L_T^i(\theta_t) - \nabla_{\theta} L^{\pi_t^i}(\theta_t) \|,  \nonumber
\end{align}
where the inequality follows from the triangle inequality.

With $\pi_t^i \in {\arg\min}_{j \in \Xcal} \| \nabla_{\theta} L^i(\theta) - \nabla_{\theta} L_T^{j}(\theta)\|$, %in which case 
the upper bound exactly equals the function $\hat{E}(\Xcal)$ defined below:
\begin{align}
    \hat{E}(\Xcal) & =  \sum_{i\in W} \min_{j \in \Xcal} \| \nabla_{\theta} L^i(\theta_t) - \nabla_{\theta} L_T^j(\theta_t) \| \nonumber \\ & = \sum_{i\in W}  \| \nabla_{\theta} L^i(\theta_t) - \nabla_{\theta} L_T^{\pi_t^i}(\theta_t) \|
\end{align}
Hence it follows that $\hat{E}(\Xcal)$ is an upper bound of $E(\Xcal)$.
\end{proof}

\subsubsection{Maximization Version of \textsc{Craig}}
We can similarly formulate the maximization version of this problem. Define:
\begin{align*}
    \hat{F}(\Xcal) &= \sum_{i\in \Wcal}  L_{\max} -  \min_{j \in \Xcal} \| \nabla_{\theta} L^i(\theta_t) - \nabla_{\theta} L_T^j(\theta_t)  \| \nonumber \\
    &= \sum_{i\in \Wcal} \max_{j \in \Xcal} \big(L_{\max} -   \| \nabla_{\theta} L^i(\theta_t) - \nabla_{\theta} L_T^j(\theta_t)\| \big) 
\end{align*}
Note that this function is exactly the Facility Location function considered in \textsc{Craig}~\cite{mirzasoleiman2020coresets}, and $\hat{F}(\Xcal)$ is a lower bound of $F(\Xcal)$. Maximizing the above expression under the constraint $|\Xcal| \leq k$ is an instance of cardinality constraint submodular maximization, and a simple greedy algorithm achieves a $1 - 1/e$ approximation~\cite{nemhauser1978analysis}.

Next, we look at the dual problem, {\em i.e.}, finding the minimum set size such that the error is bounded. Through the following minimization problem we obtain the smallest weighted subset $\Xcal$ that approximates the full gradient by an error of at most $\epsilon$ for the current parameters $\theta_t$:%$^\textrm{1}$
\begin{align}\label{eq:fl_min}
\Xcal^t =  {\min}_{\Xcal}|\Xcal|, \text{ so that, }\hat{E}(\Xcal) \leq \epsilon.
\end{align}
We can rewrite Equation~\ref{eq:fl_min} as an instance of submodular set cover:
\begin{align}\label{eq:fl_max}
\Xcal^t = &{\min}_{\Xcal}|\Xcal|, \text{s.t. } \hat{F}(\Xcal) \geq |\Wcal| L_{\max} - \epsilon.
\end{align}
This is an instance of submodular set cover, which can also be approximated up to a $\log$-factor~\cite{wolsey1982analysis,mirzasoleiman2015distributed}. In particular, denote $Q = \hat{F}(\Xcal^*)$ as  the optimal solution. Then  the greedy algorithm is guaranteed to obtain a set $\hat{\Xcal}$ such that $\hat{F}(\hat{\Xcal}) \geq |\Wcal| L_{\max} - \epsilon$ and $|\hat{\Xcal}| \leq |\Xcal^*|\log(Q/\epsilon)$. Next, note that obtaining a set $\hat{\Xcal}$ such that $\hat{F}(\hat{\Xcal}) \geq |\Wcal| L_{\max} - \epsilon$ is equivalent to $\hat{E}(\Xcal) \leq \epsilon$. Using this fact, and the convergence result of Theorem~\ref{thm:convergence-result}, we can derive convergence bounds for \textsc{Craig}. In particular, assume that using the submodular set cover, we achieve sets $\Xcal^t$ such that $\hat{E}(\Xcal^t) \leq \epsilon$. The following corollary provides a convergence result for the facility location based upper bound approach.

\subsubsection{Convergence Bound for \textsc{Craig}}
Next, we state and prove a convergence bound for \textsc{Craig}.
\begin{lemma}
Suppose the subsets $\Xcal^t$ satisfy $\hat{E}(\Xcal^t) \leq \epsilon, \forall t = 1, \cdots, T$, then using the facility location upper bound for data selection achieves the following convergence result:
\begin{itemize}[leftmargin=*,noitemsep]
    \item if $L_T$ is lipschitz continuous with parameter $\sigma_T$ and $\alpha = \frac{D}{\sigma_T \sqrt{T}}$, then $\min_{t = 1:T} L(\theta_t) - L(\theta^*) \leq \frac{D\sigma_T}{\sqrt{T}} + D\epsilon$,
    \item if $L_T$ is lipschitz smooth with parameter $\mathcal{L}_T$, and $L_T^i$ satisfies $0 \leq L_T^i(\theta) \leq \beta_T, \forall i$. Then setting $\alpha = \frac{1}{\mathcal{L}_T}$, we have $\min_{t = 1:T} L(\theta_t) - L(\theta^*) \leq \frac{D^2 \mathcal{L}_T + 2\beta_T}{2T} + D\epsilon$,
    \item if $L_T$ is Lipschitz continuous (parameter $\sigma_T$) and $L$ is strongly convex with parameter $\mu$, then setting a learning rate $\alpha_t = \frac{2}{\mu(1+t)}$, achieves $\min_{t = 1:T} L(\theta_t) - L(\theta^*) \leq \frac{2{\sigma_T}^2}{\mu (T+1)} + D\epsilon$
\end{itemize}
\end{lemma}
\begin{proof}
We prove the first part and note that the other parts follow similarly. Notice that the CRAIG algorithm tries to minimze the term $\hat{E}(\Xcal^t) =  \sum_{i\in W} \min_{j \in \Xcal^t} \| \nabla_{\theta} L^i(\theta_t) - \nabla_{\theta} L_T^j(\theta_t) \|$ which is an upper bound of $E(\Xcal^t) = \underset{\wb}{\min}\hspace{0.3cm}\mbox{Err}(\wb^t, \Xcal^t, L, L_T, \theta_t)$ from Lemma~\ref{app-craig-details} (i.e., $\mbox{Err}(\wb^t, \Xcal^t, L, L_T, \theta_t) \leq \hat{E}(\Xcal^t)$). From the assumption that $\hat{E}(\Xcal^t) \leq \epsilon$, we have  $\mbox{Err}(\wb^t, \Xcal^t, L, L_T, \theta_t) \leq \epsilon$, which combining with Theorem~\ref{thm:convergence-result}, immediately proves the required convergence result for CRAIG. Finally, note that for the third part, $\sum_{t=1}^{t=T} \frac{2Dt}{T(T+1)}\epsilon = \epsilon$ and this proves all three parts. 
\end{proof}

\section{More Experimental Details and Additional Results}
\label{app-exp}

\subsection{Datasets Description}

We used various standard datasets, namely, MNIST, CIFAR10, SVHN, CIFAR100, ImageNet, to demonstrate the effectiveness and stability of \model{}.

\begin{table}[H]
  \centering
  \begin{adjustbox}{width=1\textwidth}
      \begin{tabular}{|c|c|c|c|c|c|} 
     \hline 
     \textbf{Name} & \textbf{No. of classes} & \textbf{No. samples for} & \textbf{No. samples for} & \textbf{No. samples for} & \textbf{No. of features} \\  
     ~ & ~ & \textbf{training} & \textbf{validation} & \textbf{testing} & ~ \\ [0.5ex] 
     \hline
     CIFAR10 & 10  & 50,000 & - & 10,000 & 32x32x3\\ 
     \hline
     MNIST & 10  & 60,000 & - & 10,000 & 28x28\\ 
     \hline
     SVHN & 10 & 73,257 & - & 26,032 & 32x32x3\\
     \hline
     CIFAR100 & 100 & 50,000 & - & 10,000 & 32x32x3 \\
     \hline
     ImageNet & 1000 & 1,281,167 & 50,000 & 100,000 & 224x224x3 \\
     \hline
     \end{tabular}
 \end{adjustbox}
 \caption{Description of the datasets}
  \label{tab:d}
\end{table}

Table \ref{tab:d} gives a brief description about the datasets. Here not all datasets have an explicit validation and test set. For such datasets, 10\% and 20\% samples from the training set are used as validation and test set, respectively. The feature count given for the ImageNet dataset is after applying the \verb|RandomResizedCrop| transformation function from PyTorch ~\cite{paszke2017automatic}.

\begin{sidewaystable}[!htbp]
    \centering
    \scalebox{0.9}{
    \begin{tabular}{c c c|c c c c|c c c c} \hline \hline
\multicolumn{11}{c}{Data Selection Results}\\ \hline
\multicolumn{3}{c|}{} & \multicolumn{4}{c|}{Top-1 Test accuracy of the Model(\%)} & \multicolumn{4}{c}{Model Training time(in hrs)} \\ 
\multicolumn{1}{c}{} & \multicolumn{1}{c}{} & \multicolumn{1}{c|}{Budget(\%)} & \multicolumn{1}{c}{5\%} & \multicolumn{1}{c}{10\%} & \multicolumn{1}{c}{20\%} & \multicolumn{1}{c|}{30\%} & \multicolumn{1}{c}{5\%} & \multicolumn{1}{c}{10\%} & \multicolumn{1}{c}{20\%} & \multicolumn{1}{c}{30\%} \\ \hline
\multicolumn{1}{c}{Dataset} & \multicolumn{1}{c}{Model}  &\multicolumn{1}{c|}{Selection Strategy} & \multicolumn{4}{c|}{} & \multicolumn{4}{c}{} \\ \hline
CIFAR10 &ResNet18 &\textsc{Full} (skyline for test accuracy)  &95.09 &95.09  &95.09  &95.09  &4.34 & 4.34 &4.34 &4.34 \\ 
 & &\textsc{Random} (skyline for training time) &71.2 &80.8  &86.98  &87.6  &0.22 & 0.46 &0.92 &1.38\\
 & &\textsc{Random-Warm} (skyline for training time) &83.2 &87.8  &90.9  &92.6  &0.21 & 0.42 &0.915 &1.376\\\cline{3-11}
 & &\textsc{Glister} &85.5 &91.92 &92.78 &93.63 &0.43 &0.91 &1.13 &1.46 \\
 & &\textsc{Glister-Warm} &86.57 &91.56 &92.98 &94.09 &0.42 &0.88 &1.08 &1.40\\
 & &\textsc{Craig}  &82.74 &87.49 &90.79 &92.53 &0.81 &1.08 &1.45 &2.399\\
 & &\textsc{Craig-Warm}  &84.48 &89.28 &92.01 &92.82 &0.6636 &0.91 &1.31 &2.20\\
 & &\textsc{CraigPB}  &83.56 &88.77 &92.24 &93.58 &0.4466 &0.70 &1.13 &2.07\\
 & &\textsc{CraigPB-Warm}  &86.28 &90.07 &93.06 &93.8 &0.4143 &0.647 &1.07 &2.06\\
 & &\textsc{GradMatch}  &86.7 &90.9 &91.67 & 91.89 &0.40 &0.84 &1.42 &1.52\\
 & &\textsc{GradMatch-Warm}  &{\color{red} 87.2} &92.15 &92.11 &92.01 &0.38 &0.73 &1.24 &1.41\\
 & &\textsc{GradMatchPB}  &85.4 &90.01 &93.34 &93.75 &0.36 &0.69 &1.09 &1.38\\
 & &\textsc{GradMatchPB-Warm}  &86.37 &{\color{red}92.26} &{\color{red}93.59} &{\color{red}94.17} & {\color{red}0.32} &{\color{red}0.62} &{\color{red}1.05} &{\color{red}1.36}\\ \hline \hline
 CIFAR100 &ResNet18 &\textsc{Full} (skyline for test accuracy)  &75.37  &75.37 &75.37 &75.37 &4.871 &4.871 &4.871 &4.871\\ 
 & &\textsc{Random} (skyline for training time) &19.02 &31.56 &49.6 &58.56 &0.2475 &0.4699 &0.92 &1.453\\
 & &\textsc{Random-Warm} (skyline for training time) &58.2 &65.95 &70.3 &72.4 &0.242 &0.468 &0.921 &1.43\\\cline{3-11}
 & &\textsc{Glister} &29.94 &44.03 &61.56 &70.49 &0.3536 &0.6456 &1.11 &1.5255\\
 & &\textsc{Glister-Warm} &57.17 &64.95 &62.14 &72.43 &0.3185 &0.6059 &1.06 &{\color{red}1.452}\\
 & &\textsc{Craig}  &36.61 &55.19 &66.24 &70.01 &1.354 &1.785 &1.91 &2.654\\
 & &\textsc{Craig-Warm}  &57.44 &67.3 &69.76 &72.77 &1.09 &1.48 &1.81 &2.4112\\
 & &\textsc{CraigPB}  &38.95 &54.59 &67.12 &70.61 &0.4489 &0.6564 &1.15 &1.540\\
 & &\textsc{CraigPB-Warm}  &57.66 &67.8 &70.84 &73.79 &0.394 &0.6030 &1.10 &1.5567\\
 & &\textsc{GradMatch}  &41.01 &59.88 &68.25 &71.5 &0.5143 &0.8114 &1.40 &2.002\\
 & &\textsc{GradMatch-Warm}  &57.72 &68.23 &71.34 &74.06 &0.3788 &0.7165 &1.30 &1.985\\
 & &\textsc{GradMatchPB}  &40.53 &60.39 &70.88 &72.57 &0.3797 &0.6115 &1.09 &1.56\\
 & &\textsc{GradMatchPB-Warm}  &{\color{red}58.26} &{\color{red}69.58} &{\color{red}73.2} &{\color{red}74.62} &{\color{red}0.300} &{\color{red}0.5744} &{\color{red}1.01} &1.5683\\ \hline \hline
 SVHN &ResNet18 &\textsc{Full} (skyline for test accuracy)  &96.49 &96.49 &96.49 &96.49 &6.436 &6.436, &6.436 &6.436\\ 
 & &\textsc{Random} (skyline for training time) &89.33 &93.477 &94.7 &95.31 &0.342 &0.6383 &1.26 &1.90\\
 & &\textsc{Random-Warm} (skyline for training time) &94.1 &94.4 &95.87 &96.01 &0.34 &0.637 &1.26 &1.90\\\cline{3-11}
 & &\textsc{Glister} &94.78 &95.37 &95.5 &95.82 &0.5733 &0.9141 &1.62 &2.514\\
 & &\textsc{Glister-Warm} &{\color{red}94.99} &95.50 &95.8 &95.69 &0.5098 &0.8522 &1.58 &{\color{red}2.34}\\
 & &\textsc{Craig}  &94.003 &94.86 &95.83 &96.223 &1.3886 &1.7566 &2.39 &3.177\\
 & &\textsc{Craig-Warm}  &93.81 &95.27 &96.0 &96.15 &1.113 &1.4599 &2.15 &2.6617\\
 & &\textsc{CraigPB}  &94.26 &95.367 &95.92 &96.043 &0.5934 &1.009 &1.65 &2.413\\
 & &\textsc{CraigPB-Warm}  &94.339 &{\color{red}95.724} &96.06 &96.385 &0.5279 &0.93406 &1.58 &2.332\\
 & &\textsc{GradMatch}  &94.01 &94.45 &95.4 &95.73 &0.8153 &1.1541 &1.64 &2.981\\
 & &\textsc{GradMatch-Warm}  &94.94 &95.13 &96.03 &95.79 &0.695 &0.9313 &1.59 &2.417\\
 & &\textsc{GradMatchPB}  &94.37 &95.36 &96.12 &96.24 &0.5134 &0.8438 &1.6 &2.52\\
 & &\textsc{GradMatchPB-Warm}  &94.77 &95.64 &{\color{red}96.21} &{\color{red}96.425} &{\color{red}0.4618} &{\color{red}0.7889} &{\color{red}1.51} &2.398\\ \hline \hline
\end{tabular}}
    \caption{Data Selection Results for CIFAR10, CIFAR100 and SVHN datasets}
    \label{tab:data_sel_results}
\end{sidewaystable}

\begin{sidewaystable}[!htbp]
    \centering
    \scalebox{1}{
    \begin{tabular}{c c c|c c c c|c c c c} \hline \hline
\multicolumn{11}{c}{MNIST Data Selection Results}\\ \hline
\multicolumn{3}{c|}{} & \multicolumn{4}{c|}{Top-1 Test accuracy of the Model(\%)} & \multicolumn{4}{c}{Model Training time(in hrs)} \\ 
\multicolumn{1}{c}{} & \multicolumn{1}{c}{} & \multicolumn{1}{c|}{Budget(\%)} & \multicolumn{1}{c}{1\%} & \multicolumn{1}{c}{3\%} & \multicolumn{1}{c}{5\%} & \multicolumn{1}{c|}{10\%} & \multicolumn{1}{c}{1\%} & \multicolumn{1}{c}{3\%} & \multicolumn{1}{c}{5\%} & \multicolumn{1}{c}{10\%} \\ \hline
\multicolumn{1}{c}{Dataset} & \multicolumn{1}{c}{Model}  &\multicolumn{1}{c|}{Selection Strategy} & \multicolumn{4}{c|}{} & \multicolumn{4}{c}{} \\ \hline
MNIST &LeNet &\textsc{Full} (skyline for test accuracy)  &99.35 &99.35 &99.35 &99.35 &0.82 &0.82 &0.82 &0.82\\ 
 & &\textsc{Random} (skyline for training time) &94.55 &97.14 &97.7 &98.38 &0.0084 &0.03 &0.04 &0.084\\
 & &\textsc{Random-Warm} (skyline for training time) &98.8 &99.1 &99.1 &99.13 &0.0085 &0.03 &0.04 &0.085\\\cline{3-11}
 & &\textsc{Glister} &93.11 &98.062 &99.02 &99.134 &0.045 &0.0625 &0.082 &0.132\\
 & &\textsc{Glister-Warm} &97.63 &98.9 &99.1 &99.15 &0.04 &0.058 &0.078 &0.127\\
 & &\textsc{Craig} &96.18 &96.93 &97.81 &98.7 &0.3758 &0.4173 &0.434 &0.497\\
 & &\textsc{Craig-Warm}  &98.48 &98.96 &99.12 &99.14 &0.2239 &0.258 &0.2582 &0.3416\\
 & &\textsc{CraigPB}  &97.72 &98.47 &98.79 &99.05 &0.08352 &0.106 &0.1175 &0.185\\
 & &\textsc{CraigPB-Warm}  &98.47 &99.08 &99.01 &99.16 &0.055 &0.077 &0.0902 &0.1523\\
 & &\textsc{GradMatch}  &98.954 &99.174 &99.214 &99.24 &0.05 &0.0607 &0.097 &0.138\\
 & &\textsc{GradMatch-Warm}  &98.86 &99.22 &99.28 &99.29 &0.046 &0.057 &0.089 &0.132\\
 & &\textsc{GradMatchPB}  &98.7 &99.1 &99.25 &99.27 &0.04 &0.051 &0.07 &0.11\\
 & &\textsc{GradMatchPB-Warm}  &{\color{red}99.0} &{\color{red}99.23} &{\color{red}99.3} &{\color{red}99.31} &{\color{red}0.038} &{\color{red}0.05} &{\color{red}0.065} &{\color{red}0.10}\\ \hline \hline
\end{tabular}}
    \caption{Data Selection Results for MNIST dataset}
    \label{tab:mnist}

    \centering
    \scalebox{1}{
    \begin{tabular}{c c c|c c c|c c c} \hline \hline
\multicolumn{9}{c}{ImageNet Data Selection Results}\\ \hline
\multicolumn{3}{c|}{} & \multicolumn{3}{c|}{Top-1 Test accuracy(\%)} & \multicolumn{3}{c}{Model Training time(in hrs)} \\ 
\multicolumn{1}{c}{} & \multicolumn{1}{c}{} & \multicolumn{1}{c|}{Budget(\%)} & \multicolumn{1}{c}{5\%} & \multicolumn{1}{c}{10\%} & \multicolumn{1}{c|}{30\%} & \multicolumn{1}{c}{5\%} & \multicolumn{1}{c}{10\%} & \multicolumn{1}{c}{30\%}\\ \hline
\multicolumn{1}{c}{Dataset} & \multicolumn{1}{c}{Model}  &\multicolumn{1}{c|}{Selection Strategy} & \multicolumn{3}{c|}{} & \multicolumn{3}{c}{} \\ \hline
ImageNet &ResNet18 &\textsc{Full} (skyline for test accuracy)  &70.36 &70.36 &70.36 &276.28 &276.28 &276.28 \\ 
 & &\textsc{Random} (skyline for training time) &21.124 &33.512 &55.12 &14.12 &28.712 &81.7\\ \cline{3-9}
 & &\textsc{CraigPB}  &44.28 &55.36 &63.52 &22.24 &38.9512 &96.624\\
 & &\textsc{GradMatch}  &47.24 &56.81 &66.21 &18.24 &35.7042 &90.25\\
 & &\textsc{GradMatch-Warm}  &55.86 &58.21 &68.241 &16.48 &33.024 &88.248\\
 & &\textsc{GradMatchPB}  &45.15 &59.04 &68.12 &16.12 &30.472 &86.32\\
 & &\textsc{GradMatchPB-Warm}  &{\color{red}56.61} &{\color{red}61.16} &{\color{red}69.06} &{\color{red}15.28} &{\color{red}29.964} &{\color{red}86.05}\\ \hline \hline
\end{tabular}}
    \caption{Data Selection Results for ImageNet dataset}
    \label{tab:imagenet}
\end{sidewaystable}

\subsection{Experimental Settings}

We ran experiments using an SGD optimizer with an initial learning rate of 0.01, the momentum of 0.9, and a weight decay of 5e-4. We decay the learning rate using cosine annealing ~\cite{loshchilov2017sgdr} for each epoch. For MNIST, we use the LeNet model ~\cite{lecun1989backpropagation} and train the model for 200 epochs. For all other datasets, we use ResNet18 model ~\cite{he2016deep} and train the model for 300 epochs (except for ImageNet, where we train the model for 350 epochs).

To demonstrate our method's effectiveness in a robust learning setting, we artificially generate class-imbalance for the above datasets by removing almost 90\% of the instances from 30\% of total classes available. We ran all experiments on a single V100 GPU, except for ImageNet, where we used an RTX 2080 GPU. However, for a given dataset, all experiments were run on the same GPU so that the speedup and energy comparison across techniques is fair.

\subsection{Other specific settings}
Here we discuss various parameters' required by Algorithm~\ref{alg:algo1}, their significance, and the values used in the experiments.
\begin{itemize}
    \item \textbf{k} determines the subset size with which we train the model.
    \item \textbf{$\epsilon$} determines the extent of gradient approximation we want. We use a value of 1e-10 in our experiments. 
    \item \textbf{$\lambda$} determines how much regularization we want. We set $\lambda = 0.5$.
\end{itemize}

\subsection{Data Selection Results:}
\label{app:dss}
This section shows the results of training neural networks on subsets selected by different data selection strategies for various datasets. Table~\ref{tab:data_sel_results} shows the test accuracy and the training time of the ResNet18 model on CIFAR10, CIFAR100, and SVHN datasets for 300 epochs. Table~\ref{tab:mnist} shows the test accuracy and the training time of the LeNet model on the MNIST dataset for 200 epochs. Table~\ref{tab:imagenet} shows the test accuracy and the training time of the ResNet18 model on the ImageNet dataset for 350 epochs. From the results, it is evident that \textsc{Grad-MatchPB-Warm} not only outperforms other baselines in terms of accuracy but is also more efficient in model training times. Furthermore, \textsc{Glister} and \textsc{Craig} could not be run on ImageNet due to large memory requirements and running time. \textsc{Grad-Match}, \textsc{Grad-MatchPB}, and \textsc{CraigPB} were the only variants which could scale to ImageNet. Furthermore, \textsc{Glister} and \textsc{Craig} also perform poorly on CIFAR-100. Overall, we observe that \textsc{Grad-Match} and its variants consistently outperform all baselines by achieving higher test accuracy and lower training times.

\paragraph{Energy Consumption Results:}
Table~\ref{tab:energy_results} shows the energy consumption (in KWH) for different subset sizes of CIFAR10, CIFAR100 datasets. The results show that \textsc{Grad-MatchPB-Warm} strategy is the most efficient in energy consumption out of all other selection strategies. Similarly, we could also observe that the PerBatch variants, i.e., \textsc{CraigPB}, \textsc{Grad-MatchPB} have better energy efficiency compared to \textsc{Grad-Match} and \textsc{Craig}.

\begin{table}[!ht]
    \centering
    \scalebox{0.8}{
    \begin{tabular}{c c c|c c c c} \hline \hline
\multicolumn{7}{c}{Energy Consumption Results}\\ \hline
\multicolumn{3}{c|}{} & \multicolumn{4}{c}{Energy consumption for training the Model(in KWH)} \\ 
\multicolumn{1}{c}{} & \multicolumn{1}{c}{} & \multicolumn{1}{c|}{Budget(\%)} & \multicolumn{1}{c}{5\%} & \multicolumn{1}{c}{10\%} & \multicolumn{1}{c}{20\%} & \multicolumn{1}{c}{30\%} \\ \hline
\multicolumn{1}{c}{Dataset} & \multicolumn{1}{c}{Model}  &\multicolumn{1}{c|}{Selection Strategy} & \multicolumn{4}{c}{} \\ \hline
CIFAR10 &ResNet18 &\textsc{Full}  &0.5032	&0.5032 &0.5032 &0.5032\\ 
 & &\textsc{Random} (Skyline for Energy Consumption)  &0.0592	&0.0911	&0.1281 &0.18 \\
 & &\textsc{Random-Warm} (Skyline for Energy Consumption) &0.0581	&0.0901	&0.128 &0.176 \\\cline{3-7}
 & &\textsc{Glister} &0.0693 &0.1012 &0.1392 &0.1982\\
 & &\textsc{Glister-Warm} &0.0672 &0.0990	&0.1360 &0.1932\\
 & &\textsc{Craig}  &0.0832	&0.1195	&0.1499	&0.2063\\
 & &\textsc{Craig-Warm}  &0.0770	&0.1118 &0.1438	&0.2043\\
 & &\textsc{CraigPB}  &0.0709 &0.1031 &0.1384 &0.2005\\
 & &\textsc{CraigPB-Warm}  &0.0682	&0.1023	&0.1355	&0.2016\\
 & &\textsc{Grad-Match}  &0.0734	&0.1173 &0.1501	&0.2026\\
 & &\textsc{Grad-Match-Warm}  &0.0703	&0.1083	&0.1429	&0.2004\\
 & &\textsc{Grad-MatchPB}  &0.0670	&0.1006	&0.1378	&0.1927\\
 & &\textsc{Grad-MatchPB-Warm}  &{\color{red}0.0649}	&{\color{red}0.0978}	&{\color{red}0.1354}	&{\color{red}0.1912}\\ \hline \hline

 CIFAR100 &ResNet18 &\textsc{Full} &0.5051	&0.5051	&0.5051	&0.5051\\ 
 & &\textsc{Random} (Skyline for Energy Consumption) &0.0582	&0.0851	&0.1116 &0.1910\\
 & &\textsc{Random-Warm} (Skyline for Energy Consumption) &0.0581	&0.0850	&0.1115 &0.1910\\\cline{3-7}
 & &\textsc{Glister} &0.0674	&0.0991	&0.1454 &0.2084\\
 & &\textsc{Glister-Warm} &0.0650 &0.0940	 &0.1444 &0.2018\\
 & &\textsc{Craig}  &0.1146	&0.1294	&0.1795	&0.2378\\
 & &\textsc{Craig-Warm}  &0.0895	 &0.1209	&0.1651	&0.2306\\
 & &\textsc{CraigPB}  &0.0747	&0.0946	&0.1443	&0.2053\\
 & &\textsc{CraigPB-Warm}  &0.0710	&0.0916	&0.1447	&0.2039\\
 & &\textsc{Grad-Match}  &0.0721	&0.1129 &0.1577	&0.2297\\
 & &\textsc{Grad-Match-Warm}  &0.0688	&0.0980	&0.1531	&0.2125\\
 & &\textsc{Grad-MatchPB}  &0.0672	&0.0978	&0.1477	&0.2100\\
 & &\textsc{Grad-MatchPB-Warm}  &{\color{red}0.0649}	&{\color{red}0.0928}	&{\color{red}0.1411}	&{\color{red}0.2001}\\ \hline \hline
\end{tabular}}
    \caption{Energy consumptions results for training a ResNet18 model on CIFAR10, CIFAR100 datasets for 300 epochs}
    \label{tab:energy_results}
\end{table}

\begin{table}[!ht]
    \centering
    \scalebox{0.8}{
    \begin{tabular}{c c c|c c c c} \hline \hline
\multicolumn{7}{c}{Standard Deviation Results}\\ \hline
\multicolumn{3}{c|}{} & \multicolumn{4}{c}{Standard deviation of the Model(for 5 runs)} \\ 
\multicolumn{1}{c}{} & \multicolumn{1}{c}{} & \multicolumn{1}{c|}{Budget(\%)} & \multicolumn{1}{c}{5\%} & \multicolumn{1}{c}{10\%} & \multicolumn{1}{c}{20\%} & \multicolumn{1}{c}{30\%} \\ \hline
\multicolumn{1}{c}{Dataset} & \multicolumn{1}{c}{Model}  &\multicolumn{1}{c|}{Selection Strategy} & \multicolumn{4}{c}{} \\ \hline
CIFAR10 &ResNet18 &\textsc{Full}  &0.032	&0.032	&0.032	&0.032 \\ 
 & &\textsc{Random}  &0.483 &0.518 &0.524 &0.538 \\
 & &\textsc{Random-Warm}  &0.461 &0.348 &0.24 &0.1538 \\\cline{3-7}
 & &\textsc{Glister} &0.453	&0.107	&0.046	&0.345\\
 & &\textsc{Glister-Warm} &0.325 &0.086	&0.135 &0.129\\
 & &\textsc{Craig}  &0.289	&0.2657	&0.1894	&0.1647\\
 & &\textsc{Craig-Warm}  &0.123	&0.1185	&0.1058	&0.1051\\
 & &\textsc{CraigPB}  &0.152 &0.1021 &0.086	&0.064\\
 & &\textsc{CraigPB-Warm}  &0.0681	&0.061	&0.0623	&0.0676\\
 & &\textsc{Grad-Match}  &0.192	&0.123	&0.112	&0.1023\\
 & &\textsc{Grad-Match-Warm}  &0.1013 &0.1032	 &0.091	&0.1034\\
 & &\textsc{Grad-MatchPB}  &0.0581 &0.0571 &0.0542 &0.0584\\
 & &\textsc{Grad-MatchPB-Warm}  &0.0542	&0.0512	&0.0671	&0.0581\\ \hline \hline
 CIFAR100 &ResNet18 &\textsc{Full} &0.051	&0.051	&0.051	&0.051 \\ 
 & &\textsc{Random} &0.659	&0.584	&0.671	&0.635 \\
 & &\textsc{Random-Warm} &0.359	&0.242	&0.187	&0.175 \\\cline{3-7}
 & &\textsc{Glister} &0.463	&0.15	&0.061	&0.541\\
 & &\textsc{Glister-Warm} &0.375 &0.083	&0.121	&0.294\\
 & &\textsc{Craig}  &0.3214	&0.214	&0.195	&0.187\\
 & &\textsc{Craig-Warm}  &0.18	&0.132	&0.125	&0.115\\
 & &\textsc{CraigPB}  &0.12	&0.134	&0.123	&0.115\\
 & &\textsc{CraigPB-Warm}  &0.1176	&0.1152	&0.1128	 &0.111\\
 & &\textsc{Grad-Match}  &0.285	&0.176	&0.165	&0.156\\
 & &\textsc{Grad-Match-Warm}  &0.140	&0.134	&0.142	&0.156\\
 & &\textsc{Grad-MatchPB}  &0.104 &0.111	&0.105	&0.097\\
 & &\textsc{Grad-MatchPB-Warm}  &0.093	&0.101	&0.100	&0.098\\ \hline \hline
 \multicolumn{3}{c|}{} & \multicolumn{4}{c}{Standard deviation of the Model(for 5 runs)} \\ 
\multicolumn{1}{c}{} & \multicolumn{1}{c}{} & \multicolumn{1}{c|}{Budget(\%)} & \multicolumn{1}{c}{1\%} & \multicolumn{1}{c}{3\%} & \multicolumn{1}{c}{5\%} & \multicolumn{1}{c}{10\%} \\ \hline
 MNIST &LeNet &\textsc{Full} &0.012	&0.012	&0.012	&0.012\\ 
 & &\textsc{Random} &0.215 &0.265 &0.224 &0.213\\
 & &\textsc{Random-Warm} &0.15	&0.121	&0.110	&0.103\\\cline{3-7}
 & &\textsc{Glister} &0.256	&0.218	&0.145	&0.128\\
 & &\textsc{Glister-Warm} &0.128	&0.134	&0.119	&0.124\\
 & &\textsc{Craig}  &0.186	&0.178	&0.162	&0.125\\
 & &\textsc{Craig-Warm}  &0.0213 &0.0223 &0.0196	&0.0198\\
 & &\textsc{CraigPB}  &0.021 &0.0209 &0.0216 &0.0204\\
 & &\textsc{CraigPB-Warm}  &0.023 &0.0192 &0.0212	&0.0184\\
 & &\textsc{Grad-Match}  &0.156	&0.128	&0.135	&0.12\\
 & &\textsc{Grad-Match-Warm}  &0.087	&0.084	&0.0896	 &0.0815\\
 & &\textsc{Grad-MatchPB}  &0.0181	&0.0163	&0.0147	&0.0129\\
 & &\textsc{Grad-MatchPB-Warm}  &0.0098	&0.012	&0.0096	&0.0092\\ \hline \hline
\end{tabular}}
    \caption{Standard deviation results for CIFAR10, CIFAR100 and MNIST datasets for 5 runs}
    \label{tab:std_results}
\end{table}

\begin{table}[!htbp]
    \centering
    \scalebox{0.7125}{
    \begin{tabular}{c|ccccccccccc}
        %\cline{2-2}
        \textsc{Random} &  &   &   &   &   & &   &   &   &   \\ \cline{2-2}
        \textsc{Glister} & \multicolumn{1}{|c|}{0.0006} &  &   &   &   &&   &   &   &  \\ \cline{2-3}
        \textsc{Glister-Warm} & \multicolumn{1}{|c|}{0.0002} & \multicolumn{1}{c|}{0.0017} &  &  &   &  &   &   &   & \\ \cline{2-4}
        \textsc{Craig} & \multicolumn{1}{|c|}{0.0003} & \multicolumn{1}{c|}{0.048}& \multicolumn{1}{c|}{0.00866} &  &   &  &   &   &   & \\ \cline{2-5}
        \textsc{Craig-Warm} & \multicolumn{1}{|c|}{0.0002} & \multicolumn{1}{c|}{0.0375} & \multicolumn{1}{c|}{0.0492} & \multicolumn{1}{c|}{0.0004} & &  &   &   &   & \\ \cline{2-6}
        \textsc{CraigPB} & \multicolumn{1}{|c|}{0.0002} & \multicolumn{1}{c|}{0.0334} & \multicolumn{1}{c|}{0.0403} & \multicolumn{1}{c|}{0.0010} & \multicolumn{1}{c|}{0.0139} &  &   &   &   &\\ \cline{2-7}
        \textsc{CraigPB-Warm} & \multicolumn{1}{|c|}{0.0002} & \multicolumn{1}{c|}{0.003} & \multicolumn{1}{c|}{0.017} & \multicolumn{1}{c|}{0.0002} & \multicolumn{1}{c|}{0.0005} & \multicolumn{1}{c|}{0.0002} &   &   &   &\\ \cline{2-8}
        \textsc{Grad-Match} & \multicolumn{1}{|c|}{0.0002} & \multicolumn{1}{c|}{0.028} & \multicolumn{1}{c|}{0.031918} & \multicolumn{1}{c|}{0.0030} & \multicolumn{1}{c|}{0.0107} & \multicolumn{1}{c|}{0.0254} & \multicolumn{1}{c|}{0.015}   &    &   &\\ \cline{2-9}
        \textsc{Grad-Match-Warm} & \multicolumn{1}{|c|}{0.0002} & \multicolumn{1}{c|}{0.0008} & \multicolumn{1}{c|}{0.0018} & \multicolumn{1}{c|}{0.0008} & \multicolumn{1}{c|}{0.0057} & \multicolumn{1}{c|}{0.0065} & \multicolumn{1}{c|}{0.0117}  & \multicolumn{1}{c|}{0.0005}  &    & \\ \cline{2-10}
        \textsc{Grad-MatchPB} & \multicolumn{1}{|c|}{0.0002} & \multicolumn{1}{c|}{0.0048} & \multicolumn{1}{c|}{0.0067} & \multicolumn{1}{c|}{0.0002} & \multicolumn{1}{c|}{0.0305} & \multicolumn{1}{c|}{0.0002} & \multicolumn{1}{c|}{0.0075}  & \multicolumn{1}{c|}{0.0248}  & \multicolumn{1}{c|}{0.0305}  &\\ \cline{2-11}
        \textsc{Grad-MatchPB-Warm} & \multicolumn{1}{|c|}{0.0002} & \multicolumn{1}{c|}{0.00007} & \multicolumn{1}{c|}{0.0028} & \multicolumn{1}{c|}{0.0002} & \multicolumn{1}{c|}{0.0002} & \multicolumn{1}{c|}{0.0002} & \multicolumn{1}{c|}{0.0011}  &\multicolumn{1}{c|}{0.0005}   &\multicolumn{1}{c|}{0.0091}   &\multicolumn{1}{c|}{0.0002} &\\ \cline{2-12}
        \multicolumn{1}{c}{}& \rotatebox[origin=c]{90}{\textsc{Random}} & \rotatebox[origin=c]{90}{\textsc{Glister}} & \rotatebox[origin=c]{90}{\textsc{Glister-Warm}} & \rotatebox[origin=c]{90}{\textsc{Craig}} & \rotatebox[origin=c]{90}{\textsc{Craig-Warm}} & \rotatebox[origin=c]{90}{\textsc{CraigPB}} & \rotatebox[origin=c]{90}{\textsc{CraigPB-Warm}} & \rotatebox[origin=c]{90}{\textsc{Grad-Match}} & \rotatebox[origin=c]{90}{\textsc{Grad-Match-Warm}} & \rotatebox[origin=c]{90}{\textsc{Grad-MatchPB}} & \rotatebox[origin=c]{90}{\textsc{Grad-MatchPB-Warm}}\\ 
    \end{tabular}}
    \caption{Pairwise significance p-values using Wilcoxon signed rank test}
    \label{tab:significance}
\end{table}

\subsection{Standard deviation and statistical significance results:}
\label{app:std}
Table~\ref{tab:std_results} shows the standard deviation results over five training runs on CIFAR10, CIFAR100, and MNIST datasets. The results show that the \textsc{Grad-MatchPB-Warm} has the least standard deviation compared to other subset selection strategies. Note that the standard deviation of subset selection strategies is large for smaller subsets across different selection strategies. Furthermore, \textsc{Glister} has higher standard deviation values than random for smaller subsets, which partly explains the fact that it does not work as well for very small subsets (e.g. 1\%  - 5\%). We could also observe that the warm start variants of subset selection strategies have lower variance than non-warm-start ones from the standard deviation numbers, partly because of the better initialization they offer. Finally, the PerBatch variants \textsc{Grad-MatchPB} and \textsc{CraigPB} have lower standard deviation compared to \textsc{Grad-Match} and \textsc{Craig} which proves the effectiveness of Per-Batch approximation. 

In Table~\ref{tab:significance}, we show the p-values of one-tailed Wilcoxon signed-rank test \cite{wilcoxon1992individual} performed on every single possible pair of data selection strategies to determine whether there is a significant statistical difference between the strategies in each pair, across all datasets. Our null hypothesis is that there is no difference between the data selection strategies pair. From the results, it is evident that \textsc{Grad-MatchPB-Warm} variant significantly outperforms other baselines at $p<0.01$.

\subsection{Other Results:}
\paragraph{Gradient Errors:}
Table~\ref{tab:grad_difference} shows the average gradient error obtained by various subset selection algorithms for the MNIST dataset. We observe that the gradient error of \textsc{Grad-MatchPB} is the smallest, followed closely by \textsc{CraigPB}. From the results, it is also evident that the PerBatch variants i.e., \textsc{Grad-MatchPB} and \textsc{CraigPB} achieves lower gradient error compared to \textsc{Grad-Match} and \textsc{Craig}. Also note that \textsc{Grad-Match} has a lower gradient error compared to \textsc{Craig} and \textsc{Grad-Match-PB} has a lower gradient error compared to \textsc{Craig-PB}. This is expected since \textsc{Grad-Match} directly optimizes the gradient error while \textsc{Craig} minimizes an upper bound. Also note that \textsc{Glister} has a signifcantly larger gradient error at $1\%$ subset which partially explains the reason for bad performance of \textsc{Glister} for very small percentages.

\paragraph{Redundant Points:} 
Table~\ref{tab:redundancy} shows the redundant points, i.e., data points that were never used for training for various subset selection algorithms on the MNIST dataset. The results give us an idea of information redundancy in the MNIST dataset while simultaneously showing that we can achieve similar performances to full training using a much smaller informative subset of the MNIST dataset.
\begin{table}[!ht]
    \centering
    \scalebox{0.8}{
    \begin{tabular}{c c c|c c c c c} \hline \hline
\multicolumn{8}{c}{MNIST Gradient Error Results}\\ \hline
\multicolumn{3}{c|}{} & \multicolumn{5}{c}{Avg.Gradient error norm}\\ 
\multicolumn{1}{c}{} & \multicolumn{1}{c}{} & \multicolumn{1}{c|}{Budget(\%)} & \multicolumn{1}{c}{1\%} & \multicolumn{1}{c}{3\%} & \multicolumn{1}{c}{5\%} & \multicolumn{1}{c}{10\%} & \multicolumn{1}{c}{30\%}\\ \hline
\multicolumn{1}{c}{Dataset} & \multicolumn{1}{c}{Model}  &\multicolumn{1}{c|}{Selection Strategy} & \multicolumn{5}{c}{}\\ \hline
MNIST &LeNet &\textsc{Random}  &410.1258 &18.135 &10.515 &9.5214 &6.415\\ 
 & & \textsc{Craig} &68.3288 &19.2665 &10.9991 &6.5159 &0.3793\\
 & & \textsc{CraigPB} &17.6352 &2.9641 &1.3916 &0.4417 &0.0825\\
 & & \textsc{Glister} &545.2769	&7.9193	&1.8786	&2.8121	&0.3249\\
 & & \textsc{Grad-Match} &66.2003 &17.6965 &9.8202 &2.1122 &0.3797\\
 & & \textsc{Grad-MatchPB} &{\color{red}15.5273} &{\color{red}2.202} &{\color{red}1.1684} &{\color{red}0.3793}	&{\color{red}0.0587}\\\hline \hline
    \end{tabular}}
    \caption{Gradient approximation relative to full training gradient for various data selection strategies for different subset sizes of MNIST dataset}
    \label{tab:grad_difference}
    
    \centering
    \scalebox{0.8}{
    \begin{tabular}{c c c|c c c c c} \hline \hline
\multicolumn{8}{c}{MNIST Redundant Points Results}\\ \hline
\multicolumn{3}{c|}{} & \multicolumn{5}{c}{Percentage of Redundant Points in MNIST training data}\\ 
\multicolumn{1}{c}{} & \multicolumn{1}{c}{} & \multicolumn{1}{c|}{Budget(\%)} & \multicolumn{1}{c}{1\%} & \multicolumn{1}{c}{3\%} & \multicolumn{1}{c}{5\%} & \multicolumn{1}{c}{10\%} & \multicolumn{1}{c}{30\%}\\ \hline
\multicolumn{1}{c}{Dataset} & \multicolumn{1}{c}{Model}  &\multicolumn{1}{c|}{Selection Strategy} & \multicolumn{5}{c}{}\\ \hline
MNIST &LeNet &\textsc{Craig}  &90.381481 &74.057407 &60.492593 &36.788889 &14.425926\\ 
 & & \textsc{CraigPB} &90.405556 &73.653704 &60.327778 &35.301852 &2.875926\\
 & & \textsc{Glister} &90.712963 &77.540741 &67.544444 &45.940741 &7.774074\\
 & & \textsc{Grad-Match} &91.124074 &76.4 &62.109259 &36.114815 &2.942593\\
 & & \textsc{Grad-MatchPB} &90.187037 &73.468519 &59.757407 &36.164815 &6.751852\\\hline \hline
    \end{tabular}}
    \caption{Redunant points(i.e., points never used for training) for various data selection strategies for different subset sizes of MNIST dataset}
    \label{tab:redundancy}
\end{table}

\paragraph{Comparison between variants of \model{}:}
Table~\ref{tab:gradmatch_variants} shows the test accuracy and the training time for PerClass, PerClassPerGradient and PerBatch variants of \model{} using ResNet18 model on different subsets of CIFAR10 and CIFAR100 datasets. First, note that even though the PerClass variant achieves higher accuracy than the PerClassPerGradient variant, it is significantly slower, having a larger training time than full data training for the 30\% subset of CIFAR10 and CIFAR100. Since the PerClass variant of \model{} is not scalable, we use the PerClassPerGradient variant, which achieves comparable accuracies while being much faster. Finally, note that the PerBatch variants performed better than the other variants in test accuracy and training efficiency.
\begin{table}[!ht]
    \centering
        \scalebox{0.8}{
        \begin{tabular}{c c c|c c c c|c c c c} \hline \hline
            \multicolumn{11}{c}{Comparison between variants of \model{}}\\ \hline
            \multicolumn{3}{c|}{} & \multicolumn{4}{c|}{Top-1 Test accuracy(\%)} & \multicolumn{4}{c}{Model Training time(in hrs)} \\ 
            \multicolumn{2}{c}{} & \multicolumn{1}{c|}{Budget(\%)} & \multicolumn{1}{c}{5\%} & \multicolumn{1}{c}{10\%} & \multicolumn{1}{c}{20\%} & \multicolumn{1}{c|}{30\%} & \multicolumn{1}{c}{5\%} & \multicolumn{1}{c}{10\%} &  \multicolumn{1}{c}{20\%} & \multicolumn{1}{c}{30\%}\\ \hline
            \multicolumn{1}{c}{Dataset} & \multicolumn{1}{c}{Model}  &\multicolumn{1}{c|}{\model{} Variant} & \multicolumn{4}{c|}{} & \multicolumn{4}{c}{} \\ \hline
            CIFAR100 &ResNet18 &PerClassPerGradient &41.01 &59.88 &68.25 &71.5 &0.5143 &0.8114 &1.40 &2.002\\ 
             & & PerClass &{\color{red}41.57} &59.95 &70.87 &72.45 &0.5357 &1.225 &1.907 &3.796\\
             & & PerBatch &40.53 &{\color{red}60.39} &{\color{red}70.88} &{\color{red}72.57} &{\color{red}0.3797} &{\color{red}0.6115} &{\color{red}1.09} &{\color{red}1.56}\\
             \hline \hline
            CIFAR10 &ResNet18 &PerClassPerGradient &{\color{red}86.7} &90.9 &91.67 &91.89 &0.40 &0.84 &1.42 &1.52\\
             & &PerClass  &85.12 &{\color{red}91.04} &92.12 &93.69 &0.4225 &1.042 &1.92 &3.48\\
             & &PerBatch  &85.4 &90.01 &{\color{red}93.34} &{\color{red}93.75} &{\color{red}0.36} &{\color{red}0.69} &{\color{red}1.09} &{\color{red}1.38}\\
              \hline \hline
        \end{tabular}}
    \caption{Top-1 test accuracy(\%) and training times for variants of \model{} for different subset sizes of CIFAR10, CIFAR100 datasets}
    \label{tab:gradmatch_variants}
\end{table}
\begin{table}[!ht]
    \centering
        \scalebox{0.8}{
        \begin{tabular}{c c c|c} \hline \hline
            \multicolumn{4}{c}{Additional Data Selection Results}\\ \hline
            \multicolumn{3}{c|}{} & \multicolumn{1}{c}{Top-1 Test accuracy(\%)}\\ 
            \multicolumn{2}{c}{} & \multicolumn{1}{c|}{Budget(\%)} & \multicolumn{1}{c}{30\%}\\ \hline
            \multicolumn{1}{c}{Dataset} & \multicolumn{1}{c}{Model}  &\multicolumn{1}{c|}{Selection strategy} & \multicolumn{1}{c}{}\\ \hline
            CIFAR100 &ResNet164 &Facility Location &91.1\\
             & &Forgetting Events &92.3\\
             &&Entropy &90.4\\
             &ResNet18 &\textsc{Grad-MatchPB-Warm} &{\color{red}94.17}\\
             \hline \hline
            CIFAR10 &ResNet164 &Facility Location &64.8\\
            &&Forgetting Events &63.4\\
            &&Entropy &60.4\\
            &ResNet18 &\textsc{Grad-MatchPB-Warm} &{\color{red}74.62}\\
              \hline \hline
        \end{tabular}}
    \caption{Top-1 test accuracy(\%) and training times for additional data selection strategies on 30\% CIFAR10 and CIFAR100 subset}
    \label{tab:additional_results}
\end{table}

\paragraph{Comparison with additional subset selection methods:}
In addition to the baselines we considered so far, we compare \model{} with additional existing subset selection strategies like Facility Location~\cite{wolf2011facility}, Entropy~\cite{6813092} and Forgetting Events~\cite{toneva2018an} on CIFAR10 and CIFAR100 datasets. The results are in Table~\ref{tab:additional_results}. Note that we used the numbers reported in paper~\cite{coleman2020selection} for comparison. The authors in~\cite{coleman2020selection} used a ResNet-164 Model which is a higher complexity model compared to ResNet-18 which we use in our experiments. Even after using a lower complexity model (ResNet-18), we outperform these other baselines on both CIFAR-10 and CIFAR-100. Furthermore, we achieve this while being much faster (since we observed that the ResNet-164 model is roughly 4x slower compared to ResNet-18). Even though a much smaller model (ResNet-20) is used for data selection, the training is still done with the ResNet-164 model. Finally, note that the selection via proxy method is orthogonal to \model{} and can also be applied to \model{} to achieve further speedups. We expect that the accuracy of these baselines (Forgetting Events, Facility Location, and Entropy) to be even lower if they are used with a ResNet-18 model. The accuracy reported for these baselines (Table~\ref{tab:additional_results}) are the best among the different proxy models used in~\cite{coleman2020selection}.
%The results show that \textsc{Grad-MatchPB-Warm} outperformed the other subset selection methods even with a ResNet18 model, whereas the latter used a ResNet164 model. Even though the other subset selection methods were sped up using a selection via proxy method given in paper~\cite{coleman2020selection}, the speedups reported were significantly lower than the speedups achieved by \textsc{Grad-Match} and its variants.  

\end{document}